\documentclass[twoside,11pt]{article}

\usepackage{blindtext}

\usepackage[abbrvbib, preprint]{jmlr2e}

\usepackage{jmlr2e}
\usepackage{amsmath}
\usepackage{booktabs}
\usepackage{amsfonts}
\usepackage{algorithmic}
\usepackage{multirow}
\usepackage{algorithm}
\usepackage{subfig}
\usepackage{xcolor}
\usepackage{aliascnt}
\usepackage[nameinlink]{cleveref}

\hypersetup{
  pdftitle={Cheap Bootstrap for Fast Uncertainty Quantification of Stochastic Gradient Descent},
  pdfauthor={Henry Lam, Zitong Wang},
  pdfkeywords={stochastic gradient descent, bootstrap resampling, confidence intervals, Berry-Esseen bounds, statistical inference}
}

\newtheorem{assumption}{Assumption}
\newcommand{\indep}{\perp \!\!\! \perp}
\crefname{assumption}{Assumption}{Assumptions}

\newaliascnt{lemma}{theorem}
\newtheorem{lemma}[lemma]{Lemma}
\aliascntresetthe{lemma}
\crefname{lemma}{Lemma}{Lemmas}
\Crefname{lemma}{Lemma}{Lemmas}
\crefname{theorem}{Theorem}{Theorems}
\Crefname{theorem}{Theorem}{Theorems}
\crefname{algorithm}{Algorithm}{Algorithms}
\Crefname{algorithm}{Algorithm}{Algorithms}
\crefname{table}{Table}{Tables}
\Crefname{table}{Table}{Tables}
\crefname{figure}{Figure}{Figures}
\Crefname{figure}{Figure}{Figures}
\crefname{section}{Section}{Sections}
\Crefname{section}{Section}{Sections}
\crefname{equation}{}{}
\Crefname{equation}{}{}
\creflabelformat{equation}{#2(#1)#3}
\renewenvironment{proof}[1][\textbf{Proof}]{\par\noindent{#1\ }}{\hfill\BlackBox\\[2mm]}
\usepackage{lastpage}
\jmlrheading{27}{2026}{1-\pageref{LastPage}}{1/25}{3/26}{25-0008}{Henry Lam and Zitong Wang}

\ShortHeadings{
Cheap Bootstrap for Stochastic Gradient Descent
}{Lam and Wang}
\firstpageno{1}

\begin{document}

\title{Cheap Bootstrap for Fast Uncertainty Quantification of Stochastic Gradient Descent}
% Resampling Stochastic Gradient Descent Cheaply for Efficient Uncertainty Quantification}

\author{\name Henry Lam \email henry.lam@columbia.edu \\
       \addr Industrial Engineering and Operations Research\\
        Columbia University\\
        500 West 120th Street\\
        New York, NY 10027, USA\\
       \AND
       \name Zitong Wang \email zw2690@columbia.edu \\
       \addr Industrial Engineering and Operations Research\\
        Columbia University\\
        500 West 120th Street\\
        New York, NY 10027, USA\\}

\editor{Jianfeng Lu}

\maketitle

\begin{abstract}
Stochastic gradient descent (SGD) or stochastic approximation has been widely used in model training and stochastic optimization. While there is a huge literature on analyzing its convergence, inference on the obtained solutions from SGD has only been recently studied, yet it is important due to the growing need for uncertainty quantification. We investigate two computationally cheap resampling-based methods to construct confidence intervals for SGD solutions. One uses multiple, but few, SGDs in parallel via resampling with replacement from the data, and another operates this in an online fashion. Our methods can be regarded as enhancements of established bootstrap schemes to substantially reduce the computation effort in terms of resampling requirements, while bypassing the intricate mixing conditions in existing batching methods. We achieve these via a recent so-called cheap bootstrap idea and refinement of a Berry-Esseen-type bound for SGD. 
\end{abstract}


\section{Introduction}
Stochastic optimization commonly arises in many applications across machine learning, operations research, and scientific analysis. The problem can be formulated as:
\begin{equation}
\label{eqt: stochastic optimization problem}
    \min_{x\in\mathbb{R}^d} H(x) \triangleq \mathbb{E}_{\zeta\sim P}[h(x,\zeta)],
\end{equation}
in which $P$ is an underlying data distribution governing the randomness $\zeta\in\Omega$, and $h$ is a known real-valued function. Stochastic gradient descent (SGD) or stochastic approximation is a popular numerical approach to solve \eqref{eqt: stochastic optimization problem}. With an initial guess $x_0\in\mathbb{R}^d$, SGD iteratively updates the solution using 
\begin{equation}
\label{SA iteration}
x_{t+1} = x_t - \eta_t \nabla h(x_t, \zeta_{t+1}), \quad t = 0,\dots, n-1,
\end{equation}
where $\zeta_t$ is a sample drawn using a Monte Carlo model generator or real data. The Robbins-Monro procedure \citep{robbinsMonro} outputs $x_n$ after a large number of iterations \eqref{SA iteration}. Alternatively, one might take the average $\bar{x}_n \triangleq\frac{1}{n}\sum_{t=1}^nx_t$ as the output. This is known as the Polyak-Ruppert-Juditsky averaging \citep{doi:10.1137/0330046}, and for convenience in this paper, we call it averaged stochastic gradient descent (ASGD). Both approaches are prevalent, with ASGD known to be more robust with respect to the step size $\eta_t$ \citep{rakhlin2012making}.

We aim to conduct inference or quantify statistical uncertainty in SGD.  
% That is, we aim to construct, using the iterates \eqref{SA iteration}, a $1-\gamma$ confidence interval for (each component of) the true optimal solution $x^*$ of problem \eqref{eqt: stochastic optimization problem}. 
More specifically, our goal is to construct a $1-\gamma$ confidence interval for (each component of) the true optimal solution $x^*$ of problem \eqref{eqt: stochastic optimization problem} using the iterates \eqref{SA iteration}.
Despite the popularity of SGD, to our best knowledge, this problem has been systematically studied only recently, driven by applications in exploration \citep{lattimore2020bandit} and as stopping criteria \citep{su2018uncertainty,JMLR:v19:17-370,chen2020statistical}. In the following, we first review these recent approaches, discuss their main ideas as well as challenges, which then motivate our proposal in this paper.

% We discuss the ideas and challenges of 
\subsection{Existing Methods and Challenges}
One of the primary challenges in SGD inference arises from the serial dependence manifested by the sequence ${x_t}$. This dependence makes the construction of a consistent standard error estimator intricate. Several recent works aim to address this issue and, though with its own merits, each of these proposed approaches also encounters limitations. \cite{chen2020statistical} proposed two methods, one based on the delta method that directly approximates the asymptotic covariance of the gradient $\nabla h(x^*,\zeta)$ and the Hessian $\nabla^2 H(x^*)$ at the optimum. While this method is statistically valid, the required Hessian information is not always available in the context of SGD. For example, backpropagation can only provide first-order gradient information \citep{rumelhart1986learning}, and arguably, a major advantage of SGD lies in its Hessian-free nature. Moreover, storing a Hessian matrix requires an expensive $\mathcal O({d^2})$ space. These put aside the subtle regularity assumptions needed for consistency as noted by \cite{chen2020statistical} themselves. Along this vein, \cite{xie2024asymptotic} also proposed an inference tool for SA that gives a confidence sequence \citep{howard2021time,anytimeValidInference} based on the delta method with an asymptotic time-uniform coverage guarantee.

% to construct a $d$-dimensional confidence region for the optimal solution to problem \eqref{eqt: stochastic optimization problem}.
% Their method works 
Motivated by the previously mentioned challenges, \cite{chen2020statistical}'s second method borrows the batch mean idea in stochastic simulation output analysis \citep{glynn1990simulation,schmeiser1982batch,10.1287/opre.31.6.1090,glynn2018constructing} and Markov Chain Monte Carlo \citep{geyer1992practical,flegal2010batch,jones2006fixed}. This approach divides the iterations of SGD into $M$ batches of increasing sizes and aggregates the means of these batches to construct confidence intervals. Nonetheless, the batch mean method introduces the number of batches as a hyperparameter that needs to be tuned. Additionally, experiments show that this method is more sensitive to the quality of convergences of SGD and could underperform other methods. Relatedly, \cite{Li_Liu_Kyrillidis_Caramanis_2018} presented a batch mean method for inference in M-estimation by using an SGD trajectory with a constant step size. Instead of using batches with increasing lengths, they use batches with a fixed length but separated by gaps to overcome the dependence between iterations of SGD. \cite{9715437} studied a batch mean algorithm by elegantly canceling out the asymptotic covariance matrix of a rescaled SGD using an $F$-type statistic. The above batching methods require hyperparameter tuning like \cite{chen2020statistical} such as the batch sizes and gap between batches.

Compared to batch mean methods, \cite{Lee_Liao_Seo_Shin_2022} developed a method that directly accounts for the dependency along the SGD trajectory. By leveraging the random scaling technique, they constructed an asymptotically pivotal statistic to produce a confidence interval. This method nicely avoids the hyperparameter tuning faced in batching. However, it needs to update a $d\times d$ matrix on the fly to construct the random scaling matrix that induces some computational and memory overheads. Also using just one pass of data, \cite{pmlr-v206-chee23a} proposed a simple and scalable method that gives conservative confidence intervals based on the initial learning rate. The performance of this method relies on the estimation quality of the reciprocal of the smallest eigenvalue of the asymptotic covariance matrix.

% They also delved into the impacts of adjusting the number and sizes of batches on the overall performance of their algorithm.

Another approach is to use the bootstrap, which, advantageously, does not succumb to the computation load of variance estimation as well as the tuning and sensitivity challenges associated with batch sizes. \cite{JMLR:v19:17-370} developed an online bootstrap method that persistently maintains $B$ perturbed version of SGD estimates, updated upon each data arrival. 
% However, for their method to work properly, $B$ is required to be large.
However, as in other applications of the bootstrap, for their method to be effective, a large value of $B$ is necessary.
For linear regression problems of dimensions 10 or 20, they set $B=200$, which means 200 times more computational cost compared to running the SGD itself or using batch means.

Yet another method, HiGrad, was proposed by \cite{su2018uncertainty}. This approach is rooted in ``splitting'' an SGD trajectory. The process involves initially running SGD for a set number of steps. Once complete, the result of this iteration is used as a starting point for the next stage, where multiple SGD threads branch off, each utilizing different new data. This branching process continues for the outcome of each thread until all data is exhausted. Confidence intervals are then constructed using all the obtained split outcomes. HiGrad requires a substantial modification to the original SGD runs; in fact, there is no more ``original'' run of SGD in HiGrad.

Finally, we briefly mention a line of work on quantifying algorithmic randomness. This includes \cite{lunde2021bootstrapping}, which applied the bootstrap on streaming principal component analysis \citep{oja1982simplified}, and \cite{pmlr-v119-chen20o}, which investigated randomized Newton methods.  \cite{lopes2019estimating} and \cite{lopes2020measuring} utilized bootstrap methods to estimate the algorithmic variability for random ensembles such as bagging and random forests. Furthermore, \cite{Nesterov_CL} gave a complexity bound on the number of iterations of their method in relation to the confidence level on reaching the optimal value via SGD. However, all these works focus on assessing the uncertainty from algorithmic randomness and treat the data as fixed. As such, they are less relevant to our focus in this paper.

\subsection{Our Contributions}
Our discussion above reveals that existing approaches in SGD inference, while being carefully and elegantly designed, encounter either intricate algorithmic tuning that relates to mixing conditions (batching), substantial modification on the SGD itself (HiGrad), or computation and storage challenges (delta method, random scaling method, and online bootstrap). In this paper, we study a methodology designed to surmount these challenges concurrently. More precisely, we adopt the bootstrap approach, which does not require mixing-related tuning nor substantial modification to the original SGD. At the same time, we enhance the bootstrap to make it substantially lighter in terms of resampling cost. The latter is made possible by using a recent ``cheap bootstrap'' idea \citep{lam2022cheap,ll2023,lam2022conference} that we will describe in more detail momentarily.

Our methodology can be implemented in both offline and online fashions. The offline version, which we call the \emph{Cheap Offline Bootstrap (COfB)}, reruns the SGD using resampling with replacement from the data $B$ times and constructs confidence intervals from these resampled iterates via an approach similar to the standard error bootstrap. 
However, while this approach may appear to require heavy resampling effort, our key assertion is that the $B$ in our implementation can be very small (such as 3). In this way, our approach is computationally less demanding than the delta method \citep{chen2020statistical} and online bootstrap \citep{JMLR:v19:17-370}, does not require hyperparameter tuning in batch mean \citep{chen2020statistical,9715437}, and also does not substantially modify the SGD trajectory in HiGrad \citep{su2018uncertainty}.

A caveat of COfB is that we can only rerun SGD after all the data becomes available. Thus, it cannot be used in a single-pass streaming fashion. To address this, our online version, \emph{Cheap Online Bootstrap (COnB)}, runs multiple, namely $B+1$, SGDs in parallel on the fly as new data comes in. COnB borrows the idea of \cite{JMLR:v19:17-370} in perturbing the gradient estimate in the SGD iteration. However, like COfB, it is computationally much cheaper than \cite{JMLR:v19:17-370} as it only needs to maintain a very small number of SGD runs. In both our theory and experimentation, we illustrate that using $B=3$ already produces consistently better coverage than the existing approaches. 

Our methodology synthesizes two recent ideas. One, as mentioned earlier, is the recent cheap bootstrap idea. This approach integrates the analysis of the statistical error coming from the original data and the Monte Carlo error in approximating the resample distribution together, in contrast to separated treatment in conventional bootstraps. To explain, while conventional bootstraps rely on the approximation of the sampling distribution by the resample distribution, which is in turn approximated by running Monte Carlo to generate many realized resamples, the cheap bootstrap directly utilizes the joint distribution between the original estimate and each resample estimate to construct pivotal statistics. Subsequently, it allows the use of a minimal number of resample runs,  i.e., potentially as low as $B=1$, while maintaining large-sample exact coverages. However, it also results in longer intervals when $B$ is small. Nonetheless, as discussed in \cite{lam2022cheap} and \cite{ll2023}, the interval length advantageously shrinks quickly as $B$ increases away from 1. 

% . In particular, it capitalizes on their approximate independence so that, when combined with asymptotic normality, we can devise a
% statistical estimation error, as captured by resampling, and the Monte Carlo error In essence, rather than separating the treatment of  relying on the resemblance between the resample distribution and the sampling distribution—as is the norm with traditional bootstraps—the cheap bootstrap between the resample and original estimates. 
Our second main methodological element is to derive the asymptotic joint distribution, in particular independence, among SGD and resampled SGD's required in invoking the cheap bootstrap idea. More specifically, we prove a joint central limit theorem for both the original and resampled SGD runs when resampling with replacement. This subsequently guides us in suitably aggregating the outputs to construct asymptotically exact-coverage intervals. To this end, we generalize the recent non-asymptotic bounds for ASGD studied by \cite{shao2022berry} and \cite{anastasiou2019normal} to hold uniformly for both the original and resampled runs, under both SGD and ASGD settings. 

 % The requirement to modify SGD might limit the usage of HiGrad. 
\Cref{table: pro and con} summarizes the comparisons between our methods and benchmark techniques. 
HiGrad requires substantial changes to the SGD procedure, while other methods do not involve such changes.
The delta method, random scaling, and online bootstrap demand a relatively heavy computation or memory load. The first method requires memorizing a $d$ by $d$ Hessian approximation, the second method requires updating the $d$ by $d$ scaling matrix, and the third method requires maintaining a large number of perturbed trajectories $B$. Although our methods also introduce $B$, it can be kept very small, so we consider our methods light in terms of computational and memory load. 
As discussed in the previous section, the batch mean method, online bootstrap method, and HiGrad introduce hyperparameters that need to be tuned. For our methods, $B$ can be regarded as a hyperparameter as well, but this is typically selected to be the largest integer that fits in the computation budget, keeping in mind that $B$ as low as 1 or 2 already suffices to construct coverage-valid intervals while a larger $B$ would improve the interval width.
Lastly, the second derivative is only required by the delta method, which as discussed before can be a challenge since in some application scenarios of SGD, the second-order information may not be available. 

Finally, we conduct experiments that support our statements in several aspects. We compare our methods with established methods in the regression experiment regarding coverage probabilities and widths of confidence intervals. The results indicate that our methods generally deliver the most accurate coverage probabilities. Although our methods produce wider confidence intervals, the interval width decreases sharply when $B$ increases even slightly. In addition, our experiments also suggest that our method outperforms others in terms of robustness. Lastly, we analyze and apply our methods in high-dimensional sparse linear regression to enlarge the scope of applicability for our approach.
% to more general  versatility and possible application.
% (SAY BRIEFLY HOW OUR EXPERIMENTS SUPPORT THE STRENGTHS OF OUR METHOD?) 

\begin{table}
  \caption{Comparison among different methods. ``delta" denotes the delta method in \cite{chen2020statistical}, ``BM" denotes the batch mean method in \cite{chen2020statistical}, ``RS'' denotes the random scaling method in \cite{Lee_Liao_Seo_Shin_2022}, and ``OB" denotes the online bootstrap method in \cite{JMLR:v19:17-370}. 
  % The shorthand HiGrad, COfB, and COnB are as discussed in the main text.  
  }
  \label{table: pro and con}
\resizebox{\columnwidth}{!}{%
  \begin{tabular}{cccccccl}
    \toprule
    Property $\backslash$ Method & COfB & COnB & delta & BM & RS & OB & HiGrad\\
    \midrule
    Require substantial procedural modification & No & No & No & No & No & No & Yes\\
    Computational/memory Load & light & light & heavy & light & heavy & heavy & light\\
    Hyperparameter tuning & No & No & No & Yes & No & Yes & Yes\\
    Require second derivative & No & No & Yes & No & No & No & No\\
  \bottomrule
\end{tabular}
}
\end{table}

\section{Methodology}
Denote the underlying data distribution by $P$.
Let $x_t$ be the solution obtained in the $t$-th iteration of \eqref{SA iteration}. So the output of SGD is $x_n$ and the output of ASGD is $\bar{x}_n = \frac{1}{n}\sum_{t=1}^n x_t$.
% and consider $\{\zeta_t\}_{t=1}^n$ to be a series of independent and identically distributed ($i.i.d.$) samples drawn from this distribution. 
% Let $x_\text{out}$ be the output of (A)SGD, using step sizes $\eta_t = \eta t^{-\alpha}$ and $i.i.d.$ data $\{\zeta_t\}_{t=1}^n$ drawn from $P$. More precisely, in ASGD $x_\text{out} = \frac{1}{n}\sum_{t=1}^nx_t$, and in SGD $x_\text{out} = x_n$, where $x_t$ is the solution obtained in the $t$-th iteration of \eqref{SA iteration}. 
Let $\hat P_n$ denote the empirical distribution from data $\{\zeta_t\}_{t=1}^n$, i.e., $\hat{P}_n(\cdot) = \frac{1}{n}\sum_{t=1}^n I(\zeta_t\in\cdot)$,  where $I(\cdot)$ denotes the indicator function. We also use $(\cdot)_i$ to denote the $i$-th entry of a vector and $(\cdot)_{i,j}$ to denote the $(i,j)$-th entry of a matrix.

Our first method, COfB, works as follows. After obtaining $\bar{x}_n$ with data $\{\zeta_t\}_{t=1}^n$, we repeatedly resample with replacement from the data (i.e., draw $n$ observations from $\hat{P}_n$) and run ASGD on the resampled data for $B$ times. Denote the $b$-th resample output by $x_\text{COfB}^{*b},\ b=1,\dots,B$.
Then, the $1-\gamma$ confidence interval for the $i$-th entry of $x^*$, $i=1,\dots,d$, is given by
\begin{equation}
\label{eqt: CI}
    \mathcal{I}_{i,n}^\text{COfB} = \left[(\bar{x}_n)_{i} - t_{B,1-\frac{\gamma}{2}} s^\text{ASGD}_i, (\bar{x}_n)_{i} + t_{B, 1-\frac{\gamma}{2}}s_i^\text{ASGD}\right],
\end{equation}
where $s_i^\text{ASGD} \triangleq \sqrt{\frac{1}{B}\sum_{b=1}^B\left((x_\text{COfB}^{*b})_i - (\bar{x}_n)_i\right)^2}$, and the scalar $t_{B,1-\frac{\gamma}{2}}$ denotes the $1-\frac{\gamma}{2}$ quantile of the student-$t$ distribution with degree of freedom $B$. Importantly, in this construction, the number of reruns $B$ is not necessarily large and can be any integer at least 1. A pseudocode for COfB can be found in \Cref{Alg: COfB}.

\begin{algorithm}
\caption{Cheap Offline Bootstrap (COfB)}
\label{Alg: COfB}
\begin{algorithmic}[1]
    \STATE {\bfseries Input: } $i.i.d.$ data $\{\zeta_t\}_{t=1}^n$, number of bootstrap runs $B\geq 1$, step size sequence $\{\eta_t\}$, initial guess $x_0$, nominal coverage level $1-\gamma$.
    \STATE {\bfseries Output: } $\mathcal{I}_{i,n}^\text{COfB},\ i=1,\dots,d$.
    \STATE Run ASGD \eqref{SA iteration} to obtain $\bar{x}_n$.
   \FOR{$b \gets [1,2,\dots,B]$}
    \STATE Resample with replacement from $\{\zeta_t\}_{t=1}^n$ to obtain $\{\zeta_1^{*b},\dots, \zeta_n^{*b}\}$.
    \STATE Run ASGD for $n$ steps on $\{\zeta_t^{*b}\}_{t=1}^n$ with initialization $x_0$ to obtain $x_\text{COfB}^{*b}$.
   \ENDFOR
   \FOR{$i\gets [1,2,\dots,d]$}
    \STATE $s_i^\text{ASGD} \gets \sqrt{\frac{1}{B}\sum_{b=1}^B\left((x_\text{COfB}^{*b})_i - (\bar{x}_n)_i\right)^2}$
    \STATE$\mathcal{I}_{i,n}^\text{COfB} \gets\left[(\bar{x}_n)_{i} - t_{B,1-\frac{\gamma}{2}} s_i^\text{ASGD}, (\bar{x}_n)_{i} + t_{B,1-\frac{\gamma}{2}} s_i^\text{ASGD}\right]$
   \ENDFOR
\end{algorithmic}
\end{algorithm}

Moreover, we can replace the ASGD procedure in lines 3 and 6 of \Cref{Alg: COfB} by standard SGD. In this case, the radius of confidence intervals becomes $t_{B-1,1-\frac{\gamma}{2}} s_i^\text{SGD}$, with $s_i^\text{SGD} = \sqrt{\frac{1}{B-1}\sum_{b=1}^B\left((x_\text{COfB}^{*b})_i - (\overline{x}_\text{COfB}^{*})_i\right)^2}$, $ (\overline{x}_\text{COfB}^{*})_i \triangleq \frac{1}{B}\sum_{b=1}^B(x_\text{COfB}^{*b})_i$. Here we require $B$ to be at least 2. We will refer to this method by \textit{COfB running SGD} in the following discussions.

Note that COfB is an offline algorithm since resampling from $\{\zeta_i\}_{i=1}^n$ can only be accomplished when all the data points have been obtained. In contrast, our second method, COnB, works by maintaining $B+1$ parallel runs of ASGD starting from the same initialization. One of these trajectories is the original run following exactly \eqref{SA iteration}. The other $B$ trajectories update similarly, except that the gradient estimate $\nabla h(x_t, \zeta_{t+1})$ is perturbed by a factor $W_{t,b}$ following exponential distribution with rate $1$.
% , where $\{W_{t,b}\}_{(t,b) = (1,1)}^{(n,B)}$ are $i.i.d.$ random variables following exponential distribution with rate $1$. 
The confidence intervals $\mathcal{I}_{i,n}^\text{COnB}$ are constructed in the same way as COfB with $\bar{x}_n$ and $\{x_\text{COnB}^{*b}\}_{b=1}^B$. When new data $\zeta_{t}$ arrives, COnB uses only $B+1$ gradient calculations to update the original and resampled outputs. Moreover, like COfB, $B$ is not necessarily large, and in this method, it can be any positive integer. A pseudocode of COnB is in \Cref{alg: COnB}. 

\begin{algorithm}
\caption{Cheap Online Bootstrap (COnB)}
\label{alg: COnB}
\begin{algorithmic}[1]
    \STATE {\bfseries Input: } $i.i.d.$ data $\{\zeta_t\}_{t=1}^n$, number of bootstrap runs $B\geq 1$, step size sequence $\{\eta_t\}$, initial guess $x_0$, nominal coverage level $1-\gamma$. 
    % (DON'T WRITE PERTRUBATION FACTOR W AS INPUTS. INSTEAD, PLEASE WRITE THEM IN THE ALGO BELOW, AND SAY EXACTLY HOW THEY ARE GENERATED.)
    \STATE {\bfseries Output: }$\mathcal{I}_{i,n}^\text{COnB},\ i=1,\dots,d$.
\FOR{$t\gets [1,2,\dots,n]$}
    \STATE $x_t \gets x_{t-1} - \eta_t \nabla h(x_{t-1}, \zeta_t)$
    \FOR{$b\gets [1,2,\dots,B]$}
    \STATE Randomly generate $W_{t,b}$ from exponential distribution with rate $1$.
    \STATE $x_t^{*b} \gets x_{t-1}^{*b} - \eta_t W_{t,b} \nabla h(x_{t-1}^{*b}, \zeta_t)$
    \ENDFOR
\ENDFOR
% \STATE $x_\text{out} \gets \frac{1}{n} \sum_{t=1}^n x_t$
\FOR{$b\gets [1,2,\dots,B]$}
\STATE $x_\text{COnB}^{*b}\gets \frac{1}{n}\sum_{t=1}^n x_t^{*b}$
\ENDFOR
\FOR{$i\gets [1,2,\dots,d]$}
\STATE $s_i^\text{ASGD} \gets \sqrt{\frac{1}{B}\sum_{b=1}^B\left((x_\text{COnB}^{*b})_i - (\bar{x}_n)_i\right)^2}$
\STATE $\mathcal{I}_{i,n}^\text{COnB} \gets\left[(\bar{x}_n)_{i} - t_{B,1-\frac{\gamma}{2}} s_i^\text{ASGD}, (\bar{x}_n)_{i} + t_{B,1-\frac{\gamma}{2}} s_i^\text{ASGD}\right]$
\ENDFOR
\end{algorithmic}
\end{algorithm}
\section{Main Theoretical Guarantees}
Our main theoretical guarantees on COnB and COfB is on coverage exactness, asymptotically as $n$ increases, for $B$ fixed to be as low as either one or two. To explain and state this result, 
% let $x^*$ denote an optimal solution to \eqref{eqt: stochastic optimization problem}, 
let $H_n(\cdot)= \frac{1}{n}\sum_{i=1}^n h(x,\zeta_i)$ denote the sample average approximation (SAA) of \eqref{eqt: stochastic optimization problem} and $\hat x_n$ the minimizer of $H_n(\cdot)$. 
$\|x\|_p$ denotes $\left(\mathbb{E}[\|x\|^p]\right)^{\frac{1}{p}}$ for a random variable $x$ and $\|\cdot\|$ denotes the standard Euclidean 2-norm for vectors. 
% {\color{blue}[TO REMOVE]
% Let $\mathcal{X}_1$ be a bounded subset of $\mathbb{R}^d$ containing $x^*$ in its interior, and let $\mathcal{X} = \{x | \sup_{y\in\mathcal{X}_1} \|x - y\| \leq \epsilon_1\}$ for some $\epsilon_1>0$. 
% For each $i,j$, define the function classes $\mathcal{F}_{i,j} = \{\partial^2_{i,j} h(x,\zeta) | x\in\mathcal{X}\}$ and $\tilde{\mathcal{F}}_{i} = \{(\partial_i h(x_1,\zeta) - \sum_j \partial^2_{i,j}h(x_2,\zeta)(x_1 - x_2))/\|x_1 - x_2\| | x_1 \in \mathcal{X}, x_2 \in \mathcal{X}_1, x_1 \neq x_2\}$. 
% }
Let $\mathcal{X}$ be a closed and bounded neighborhood of $x^*$. For each $i,j$, define the function class $\mathcal{F}_{i,j} = \{\partial^2_{i,j} h(x,\zeta) | x\in\mathcal{X}\}$.
These function classes represent the scopes of the higher-order terms of the Taylor expansion of $H$ at $x^*$.
Let $G(x) = \nabla^2 H(x)$ and $S(x) = \mathbb E[\nabla h(x,\zeta) (\nabla h(x,\zeta))^\top]$ be the Hessian of $H$ and covariance matrix of $\nabla h(x, \zeta)$ respectively. 
Given $n$ data points, define $G_n (x) = \frac{1}{n}\sum_{i=1}^n \nabla^2 h(x,\zeta_i)$ and $S_n (x) = \frac{1}{n}\sum_{i=1}^n \nabla h(x,\zeta_i) (\nabla h(x,\zeta_i))^\top$.
% {\color{red} , which are needed developing the required asymptotic properties
% \begin{assumption}
% % [Strongly Convex Objective with Bounded Hessian]
% \label{assumption: strongly convex objective with bounded hessian}
%     $h$ and $H$ are twice continuously differentiable in $x$. Moreover, the eigenvalues of $\nabla^2h (x,\zeta)$ lies in $[l,L]$ for some positive real numbers $0<l<L$ for all $x,\zeta$.
% \end{assumption}
% \begin{assumption}
% % [Unbiased Gradient Estimation]
% \label{assumption: unbiased gradient estimation}
%     The noise of estimated gradient $\{\nabla h(x_{t-1}, \zeta_t)-\nabla H(x_{t-1})\}_{t=1}^n$ is $i.i.d.$ with mean 0.
% \end{assumption}
% }

\begin{assumption}
% [Strongly Convex Objective with Bounded Hessian]
\label{assumption: strongly convex objective with bounded hessian}
    $h$ and $H$ are twice continuously differentiable in $x$. The eigenvalues of $\nabla^2h (x,\zeta)$ lie in $[l,L]$ for some positive real numbers $0<l<L$ for all $x,\zeta$. Moreover, $\nabla^2h$ is Lipschitz continuous in $x$ with constant $l_1$ on $\mathcal{X}$ uniform in $\zeta$, i.e., for all $\zeta$ 
    \begin{equation}
    \label{eqt: hessian lipschitz condition}
    \|\nabla^2 h(x_1,\zeta) - \nabla^2 h(x_2, \zeta)\| \leq l_1 \|x_1- x_2\|,\quad \forall x_1, x_2 \in \mathcal{X}.
    \end{equation}
    % there exists $L_1 > 0$ such that 
    % \begin{equation}\label{eqt: G lipschitz}
    % \|\nabla H(x) - G(x^*)(x-x^*)\| \leq L_1 \| x-x^*\|^2
    % \end{equation}
    % for all $x$. And given data $\zeta_1, \dots, \zeta_n$,
    % \begin{equation}\label{eqt: Gn lipschitz}
    % \|\sum_{i=1}^n \nabla h(x, \zeta_i) - G_n(\hat{x}_n) (x - \hat{x}_n)\| \leq L_1 \|x-\hat{x}_n\|^2
    % \end{equation}
\end{assumption}
\begin{assumption}
% [Unbiased Gradient Estimation]
\label{assumption: unbiased gradient estimation}
    Consider the filtration $\{\mathcal{F}_t = \sigma(\zeta_k | k\leq t)\}_{t\geq 0}$. The noise of estimated gradient is a martingale-difference process, i.e.,
    \[
    \mathbb{E} [\nabla h(x_{t-1}, \zeta_t) - \nabla H(x_{t-1}) | \mathcal{F}_{t-1}] = 0 \quad a.s..
    \]
\end{assumption}

\begin{assumption} \label{assumption: gradient variability}
There are $\tau_0, \tau>0$ such that $\|x_0 - x^*\| \leq \tau_0$ and $\|\nabla h(x^*,\zeta)\|_4 \leq \tau$. The eigenvalues of $S(x^*)= \mathbb E[\nabla h(x^*,\zeta) (\nabla h(x^*,\zeta))^\top]$ lie in the interval $[\lambda_1, \lambda_2]$ for some positive constants $\lambda_1<\lambda_2$.
\end{assumption}

% Only differentiability of $H$ is required in \cite{shao2022berry}. The last requirement of \Cref{assumption: strongly convex objective with bounded hessian} is in fact a local property in the sense that it can be derived from Lipschitz property of hessian of $H$ near $x^*$. We put it in the current form to make the notation less dense.  However, establishing such bounds significantly complicate the presentation and distract from our main contribution. So we adopt this stronger sample-wise curvature assumption for clarity of exposition.

\Cref{assumption: strongly convex objective with bounded hessian} specifies that the objective function $h$ exhibits strong convexity along with a bounded Hessian, which implies the same property holds for $H$, in particular its strong convexity. 
Thus, it guarantees the existence and uniqueness of $x^*$ that satisfies the first-order optimality condition $\nabla H(x^*) = 0$. The role of the sample-wise strong convexity assumption is to guarantee that the empirical Hessians for both the original and bootstrap samples are uniformly well-conditioned, which ensures sufficient control on the aggregate errors exhibited in the (A)SGD trajectory. This assumption could potentially be replaced by high probability lower bounds on the smallest eigenvalues of $G_n$ and its bootstrap versions, although we do not pursue this relaxation for this work (nonetheless, see \Cref{app: additional discussion on assumptions} for some discussions).
% , and all our numerical examples satisfy our current assumptions (CORRECT?)

\eqref{eqt: hessian lipschitz condition} is introduced to prove that the estimation error of SGD induced by non-linearity of the gradient of objective function vanishes. In fact, $G$ and $G_n$, conditional on data, being Lipschitz at optimum is sufficient for this purpose.

\Cref{assumption: unbiased gradient estimation} stipulates that the evaluation noise in the first-order gradient oracle is unbiased, which is a standard assumption to ensure the convergence of (A)SGD. 
\Cref{assumption: gradient variability} limits the variability of $\nabla h(x,\zeta)$, which is required to establish asymptotic normality. A short discussion on how \Cref{assumption: strongly convex objective with bounded hessian,assumption: unbiased gradient estimation,assumption: gradient variability}
% , \Cref{assumption: unbiased gradient estimation,assumption: gradient variability} 
imply the assumptions in \cite{shao2022berry}, which we utilize later in this paper, can be found in \Cref{app: assumptions discussion}.

% We also need the estimation of the objective of the original stochastic optimization using data, namely,2 $H_n$, to be consistent in the sense that the absolute value of the difference between $H_n$ and $H$ converges to 0 in probability uniformly in $x$ 
We also make the following assumption regarding the uniform convergence of $H_n$ to $H$, which is mildly stronger than pointwise convergence given that $\mathcal{X}$ is compact. This assumption, together with \Cref{assumption: strongly convex objective with bounded hessian}, will guarantee that the SAA solution $\hat{x}_n$ is consistent in the sense that $\hat{x}_n$ converges to $x^*$ almost surely. 
% The following assumption together with \Cref{assumption: strongly convex objective with bounded hessian} is sufficient for this requirement. , and that $\hat{x}_n$ lies in $\mathcal{X}$ for large $n$
% The condition $\hat{x}_n \in \mathcal{X}$ is also mild.
% (EXPLAIN WHAT'S M-ESTIMATOR, OR OTHERWISE DON'T USE IT; ALSO, EXLPAIN CONSISTENCY IN THE SENSE OF THE FOLLOWING ASSUMPTION?).
\begin{assumption}\label{assumption: m estimator consistency}
$\hat{x}_n \in \mathcal{X}$ with probability $1$ for $n$ large enough, and
\[
\sup_{x\in\mathcal{X}} |H_n(x) - H(x)| \rightarrow 0\quad w.p. 1.
\]
\end{assumption}
 
% A further sufficient condition for \Cref{assumption: m estimator consistency} is that the function class $\{h(x,\zeta) | x\in\mathbb{R}^d\}$ is Glivenko-Cantelli, which can be implied by \Cref{assumption: strongly convex objective with bounded hessian} if the space of $x$ is a bounded subset of $\mathbb{R}^d$ \cite{van2000asymptotic}, though we do not assume the latter here. 

The following two assumptions are specialized for ASGD and SGD considered in this work respectively. The specific choice of step size guarantees the convergence of (A)SGD in distribution. 
% In the analysis, we break down the residual $x_\text{out}-x^*$ into three terms: 1) the term coming from the residual of the initial guess; 2) the linear term of the Taylor expansion of $H$; 3) and the second-order remainder the Taylor expansion.
% The Glivenko-Cantelli assumptions help us analyze the vanishing property of some terms in our analysis of the residual $x_\text{out}-x^*$.
The Glivenko-Cantelli assumption helps us bridge the gap between the asymptotic distributions of the residual of the original output and the resampled output.
% of the second-order remainder of Taylor expansion 
% of $H$ evaluated along the trajectory of (A)SGD 
% (WHY DO WE NEED THIS?) 
% (ALSO, AS WE HAVE DISCUSSED BEFORE, THERE IS NO FIRST OR SECOND ORDER OPTIMALITY CONDITIONS ASSUMED HERE. CAN YOU EXPLAIN? I TRUST THAT THE ASSUMPTIONS HERE ARE ENOUGH, BUT FOR A READER, INCLUDING MYSELF, IT'S NOT EASY TO SEE WHY YOU DON'T NEED THE OPTIMALITY CONDITIONS)
% (DO YOU CAPTIALIZE "BOOTSTRAP" OR NOT? PLEASE CHECK TO BE CONSISTENT THROUGHTOUT THE PAPER. WE DON'T CAPITALIZE "BOOTSTRAP" IN THE INTRO).
\begin{assumption}\label{assumption: ASGD step size}
The step size satisfies $\eta_t  = \eta t^{-\alpha}$ for some $\alpha\in (\frac{1}{2},1]$. For each $i,j$, function class $\mathcal{F}_{i,j}$ is $P$-Glivenko-Cantelli. 
% (HAVE WE INTRODUCED $P$?)
\end{assumption}
\begin{assumption}\label{assumption: SGD step size}
The step size is $\eta_t = \eta t^{-1}$, and the initial step size $\eta$ satisfies $\eta l > \frac{1}{2}$. For each $i,j$, function class $\mathcal{F}_{i,j}$ is $P$-Glivenko-Cantelli.
\end{assumption}
% For the perturbation sequence used in COnB, we need the following assumption:
% \begin{assumption}
% \label{assumption: online bootstrap pertubation}
%     $\{W_{t,b}\}_{(t,b) = (1,1)}^{(n,B)}$ are $i.i.d.$ non-negative random variables with mean and variance 1.
% \end{assumption}
% }

Essentially, a function class is Glivenko-Cantelli if the law of large numbers holds uniformly over the class. Given the continuity of $\nabla^2 h$ in $x$ and compactness of $\mathcal{X}$, $\mathcal{F}_{i,j}$ is Glivenko-Cantelli if it admits an integrable envelope function. 

With the above assumptions, we have the following theorem:
% \begin{theorem}
% \label{thm: main theorem}
% Under \Cref{assumption: strongly convex objective with bounded hessian,assumption: unbiased gradient estimation,assumption: gradient variability,assumption: m estimator consistency,assumption: ASGD step size} for COfB running ASGD, or \Cref{assumption: strongly convex objective with bounded hessian,assumption: unbiased gradient estimation,assumption: gradient variability,assumption: m estimator consistency,assumption: SGD step size} for COfB running SGD, we have, for any fixed $B\geq 2$, $i=1,\dots,d$, the COfB $1-\gamma$ confidence interval for the $i$-th entry is asymptotically exact in the sense
% \begin{equation}\label{eqt: main theorem}
% \lim_{n\rightarrow\infty}\mathbb{P}(x_i^*\in\mathcal{I}_{i,n}^\text{COfB}) = 1-\gamma.
% \end{equation}
% Moreover, under \Cref{assumption: strongly convex objective with bounded hessian,assumption: unbiased gradient estimation,assumption: gradient variability,assumption: m estimator consistency,assumption: ASGD step size} for COnB, we have, for any fixed $B\geq1$, $i=1,\dots, d$, the COnB $1-\gamma$ confidence interval for the $i$-th entry is asymptotically exact in the sense
% \begin{equation}\label{eqt: main theorem1}
% \lim_{n\rightarrow\infty}\mathbb{P}(x_i^*\in\mathcal{I}_{i,n}^\text{COnB}) = 1-\gamma.
% \end{equation}
% \end{theorem}

\begin{theorem}
\label{thm: main theorem}
Under \Cref{assumption: strongly convex objective with bounded hessian,assumption: unbiased gradient estimation,assumption: gradient variability,assumption: m estimator consistency,assumption: ASGD step size}, for any fixed $B\geq1$, $i=1,\dots, d$, and $\gamma\in(0,1)$, the confidence intervals for the $i$-th entry generated by \Cref{Alg: COfB,alg: COnB} are asymptotically exact in the sense
\begin{equation}\label{eqt: main theorem1}
\lim_{n\rightarrow\infty}\mathbb{P}(x_i^*\in\mathcal{I}_{i,n}^\text{COfB})  = 1-\gamma,\quad \lim_{n\rightarrow\infty}\mathbb{P}(x_i^*\in\mathcal{I}_{i,n}^\text{COnB}) = 1-\gamma.
\end{equation}
Moreover, under \Cref{assumption: strongly convex objective with bounded hessian,assumption: unbiased gradient estimation,assumption: gradient variability,assumption: m estimator consistency,assumption: SGD step size}, for any fixed $B\geq2$, $i=1,\dots, d$, and $\gamma\in(0,1)$, the confidence interval for the $i$-th entry generated by COfB running SGD is also asymptotically exact in the sense
\begin{equation}
    \label{eqt: main theorem2}
\lim_{n\rightarrow\infty}\mathbb{P}(x_i^*\in\mathcal{I}_{i,n}^\text{COfB}) = 1-\gamma.
\end{equation}
\end{theorem}

\Cref{thm: main theorem} states that COfB and COnB attain asymptotically exact coverage as the sample size $n\to\infty$, \emph{regardless of any fixed choice of $B$}. Note the subtlety that COfB running SGD requires $B\geq 2$, but our methods running ASGD are valid even for $B$ as small as $1$. This discrepancy comes from the slight difference in the joint asymptotic limits among the original and resample runs of SGD and ASGD, which will be discussed in \Cref{thm: asymptotic result} in the following section. Moreover, note that COnB works only for ASGD. Whether it will work for SGD is still open to us due to the delicacy of the asymptotic behavior for SGD in this case.

% quantile used in our method is $12.71$, as opposed to  $1.96$ for the normal quantile. $t$-quantile becomes $3.18$, $2.57$, and $2.28$, respectively. Consequently, 
We conclude this section by remarking on the expected width of the confidence intervals generated by our methods. As will be evidenced by our subsequent analyses, our intervals are based on $t$-statistic construction and thus follow the behavior of $t$-intervals. Specifically, our widths are larger than those of normality intervals, but shrink rapidly as $B$ increases. 
% For instance, when $B=1$, with a confidence level of $95\%$, the expected width can be calculated to be $417.3\%$ larger than the normality interval. However, when $B=3, 5$, and $10$, the corresponding relative inflations become $49.6\%$, $24.8\%$ and $10.9\%$ respectively. That is, the expected lengths of our confidence intervals decrease rapidly towards the normality-based level as $B$ moves away from 1. This behavior can also be observed in the experiments. 
In our experiments, it appears that using a small $B$, such as $B=3$ or $5$, already gives a good balance in execution time and expected interval width.
In addition, this moderate compensation in the interval width significantly improves coverage probability over the benchmark methods.
Our experimental discussions in \Cref{section: experiments discussions} provide more details on these observations.

\section{Ideas behind the Main Guarantees}

In this section, we delineate the development of \Cref{thm: main theorem} in three layers. First, we establish the conditional convergence for the error of the resample runs of our methods, which is widely utilized in classical bootstraps. For COnB, we borrow this result from \cite{JMLR:v19:17-370}. For COfB, we generalize a newly developed Berry-Esseen type bound from recent work by \cite{shao2022berry}. Second, we show a translation from conditional convergence to the asymptotic independence between the error of the original estimate and the resample estimates. Finally, we leverage the cheap bootstrap method by \cite{lam2022cheap} to convert the asymptotic independence above into large-sample coverage-exact interval construction.
% which is contingent upon the previously established asymptotic independence result.

\subsection{Conditional Convergence via a Uniform Non-Asymptotic Bound}
We start with the following asymptotic result, which describes the resemblance between the errors of the original and resample runs. 
\begin{theorem}
\label{thm:nonasymptotic bound}
Under the same assumptions as in \Cref{thm: main theorem}, we have
\begin{equation}\label{eqt: SA asymptotic convergence}
    \sqrt{n}(\bar{x}_n -x^*)\xrightarrow[n\rightarrow\infty]{d} Z^\text{ASGD},\quad 
    \sqrt{n}(x_n -x^*)\xrightarrow[n\rightarrow\infty]{d} Z^\text{SGD},
\end{equation}
where both $Z^\text{ASGD}$ and $Z^\text{SGD}$ are d-dimensional Gaussian random variable with mean $0$. 
 
The covariance matrix of $Z^\text{ASGD}$ is given by $G(x^*)^{-1}S(x^*)G(x^*)^{-1}$, where $G(x^*) = \nabla^2 H(x^*)$ and $S(x^*) = \mathbb{E}[\nabla h(x^*,\zeta) (\nabla h(x^*,\zeta))^\top]$. 
    
For the covariance matrix of $Z^\text{SGD}$, consider the singular value decomposition $G(x^*) = Q D Q^\top$ with $D=\text{diag}(d_1, \dots, d_d)$, where $d_1, \dots, d_d$ are eigenvalues of $G(x^*)$ in decreasing order and $Q$ the matrix consisting of eigenvectors.
    
Let $x^{*b}$ denote $x_\text{COfB}^{*b}$ or $x_\text{COnB}^{*b}$ in \Cref{Alg: COfB,alg: COnB} respectively. We have
 \begin{equation}\label{eqt: classical bootstrap theorem1}
\sqrt{n} (x^{*b}-\bar{x}_n) \xrightarrow[n\rightarrow\infty]{d}Z^\text{ASGD} \ \ \text{conditional on } \zeta_1, \zeta_2, \dots \ \text{in probability}.
\end{equation}
    
 For COfB running SGD, we have
 \begin{equation}\label{eqt: classical bootstrap theorem}
\sqrt{n} (x_\text{COfB}^{*b}-\hat{x}_n)\xrightarrow[n\rightarrow\infty]{d}Z^\text{SGD} \ \ \text{conditional on } \zeta_1, \zeta_2, \dots\ \text{in probability}.
\end{equation}
\end{theorem}
Partial results in \Cref{thm:nonasymptotic bound} have already been established in the literature. Note that \eqref{eqt: SA asymptotic convergence} is the classical asymptotic normality of (A)SGD guaranteed by our assumptions \citep{chung1954stochastic,sacks1958asymptotic}. 
On the other hand, the type of conditional convergence in \eqref{eqt: classical bootstrap theorem1} and \eqref{eqt: classical bootstrap theorem} is the main driver of classical bootstrap methods that allow the approximation of a sampling distribution using its resampled counterpart.
In the case of COnB, the desired conditional convergence result \eqref{eqt: classical bootstrap theorem1} is well established in Appendix A.2 of \cite{JMLR:v19:17-370}. 

% We thus focus on proving \eqref{eqt: classical bootstrap theorem,eqt: classical bootstrap theorem1} 
For COfB, nonetheless, it remains to prove \eqref{eqt: classical bootstrap theorem1} and \eqref{eqt: classical bootstrap theorem}. First, notice that the ASGD output $\bar{x}_n$ and the SAA solution $\hat{x}_n$ have an asymptotically negligible discrepancy (see \Cref{sup: ASGD equiv SAA}), so that it suffices to prove the modified version of \eqref{eqt: classical bootstrap theorem1} that replaces the center $\bar{x}_n$ by $\hat{x}_n$ for COfB running ASGD
to establish \eqref{eqt: classical bootstrap theorem1}. That is, it suffices to show that
\begin{equation}\label{eqt: classical bootstrap theorem 2}
\sqrt{n} (x^{*b}_\text{COfB}-\hat{x}_n) \xrightarrow[n\rightarrow\infty]{d}Z^\text{ASGD} \ \ \text{conditional on } \zeta_1, \zeta_2, \dots \ \text{in probability},
\end{equation}
for COfB running ASGD.
In the rest of this subsection, we outline the sketch of proof for \eqref{eqt: classical bootstrap theorem} and \eqref{eqt: classical bootstrap theorem 2}.

To elucidate our proof idea, we denote $\psi(P)$ as the minimizer for \eqref{eqt: stochastic optimization problem} with data $\zeta$ following distribution $P$, where $\psi$ is viewed as a mapping from the data distribution to $\mathbb{R}^d$. 
Correspondingly, define $\psi_n$ as the mapping from the data distribution to the outcome of (A)SGD.
Then $\psi_n(P)\in \mathbb{R}^d$ is the (random) outcome of (A)SGD after $n$ iterations, as a function of data distribution $P$ with $h$ and $\{\eta_t\}_{t=1}^n$ implicitly chosen. 
With the introduced notation, \eqref{eqt: SA asymptotic convergence} can be restated as the weak limit of $\sqrt{n}(\psi_n(P) - \psi(P))$ being equal to $Z_0$.
This $Z_0$, depending on the context, denotes the Gaussian variable $Z^\text{SGD}$ or $Z^\text{ASGD}$ as described in \Cref{thm:nonasymptotic bound}, whose variance depends on $P$. Correspondingly, let $\hat{Z}_m$ denote a normal variable that replaces $P$ in its variance with $\hat P_m$, conditional on the collected data.
% following the corresponding limiting distributions.
% Let $D$ be any Borel measurable set on $\mathbb{R}^d$, 
With these new notations,
\eqref{eqt: classical bootstrap theorem} and \eqref{eqt: classical bootstrap theorem 2} hold if for any Borel measurable set $D\subset \mathbb{R}^d$,
we have
\begin{equation}
    \lim_{n\rightarrow\infty}|\mathbb{P}^*( \sqrt{n}\left( \psi_n(\hat{P}_n) - \psi(\hat{P}_n) \right) \in D) - \mathbb{P}(Z_0\in D) | = 0 \quad \text{in probability},
\end{equation}
where $\mathbb{P}^*$ denotes the probability conditional on the data.  
% We will also use $\mathbb{E}^*$ to denote the corresponding conditional expectation (THIS NOTATION DOESN'T SEEM NEEDED IN THE MAIN BODY). 
By the triangle inequality, one can obtain
\begin{align*}
    &| \mathbb{P}^*(\sqrt{n}\left(\psi_n(\hat{P}_n)-\psi(\hat{P}_n)\right)\in D) - \mathbb{P}(Z_0\in D)|\\
    \leq & |\mathbb{P}^*(\sqrt{n}\left(\psi_n(\hat{P}_n)-\psi(\hat{P}_n)\right)\in D) - \mathbb{P}^*(\hat{Z}_n\in D)| + |\mathbb{P}^*(\hat{Z}_{n}\in D) - \mathbb{P}(Z_0\in D)|.
\end{align*}
It can be proved that the second term above vanishes in probability; see \Cref{lem: clt mn separate 1} in \Cref{subsection: convergence normal variables} for details. On the other hand, we have the following theorem for the first term: 
\begin{theorem}
\label{thm:nonasymptotic bound1}
Under the same assumptions as in \Cref{thm: main theorem}, for any Borel measurable set $D$, we have
\begin{align}\label{eqt: non-asymptotic bound}
\begin{split}
    \lim_{n\rightarrow\infty}|\mathbb{P}^*\left(\sqrt{n}(\psi_n(\hat{P}_n) - \psi(\hat{P}_n))\in D\right) - \mathbb{P}^*(\hat{Z}_n\in D)|
    = 0 \quad \text{in probability}.
\end{split}
\end{align}
\end{theorem}
% To be specific, if we establish an upper bound that vanishes as $n\rightarrow\infty$ for the following term
% \begin{equation}\label{eqt: inner term}
% |\mathbb{E}^*[g(\sqrt{n}(\psi_n(\hat{P}_n)-\hat{\psi}(\hat{P}_n))) - g(Z_n)]|
% \end{equation}
% then, we can prove the desired relation \eqref{eqt: classical bootstrap theorem}.
% Discussion on how fast the outcome of ASGD converges to its asymptotic distribution can also be found in \cite{anastasiou2019normal, shao2022berry}. Using the same tool (\Cref{lem: shao lemma 4}) giving a variational bound for the convergence rate of the term of our interest, we can also establish the desired bound for the SGD case.

% The following Lemma gives a non-asymptotic bound describing how fast the outcome of (A)SGD converges to its asymptotic distribution, which is discussed in \cite{anastasiou2019normal} and \cite{shao2022berry}. In particular, we state the result by \cite{shao2022berry}, which gives a variational bound for the convergence rate of the term of our interest

% With \Cref{lem: shao lemma 4}, we can fill the gap between \eqref{eqt: mn separate} and \eqref{eqt: classical bootstrap theorem} for both ASGD and SGD cases for COfB.
% Notice that \eqref{eqt: non-asymptotic bound} is equivalent to $|\mathbb{P}^*(\sqrt{n}(\psi_n(\hat{P}_n) - \hat{\psi}(\hat{P}_n))\in D) - \mathbb{P}(\hat{Z}_n\in D)|\rightarrow 0$. 
The proof invokes an expansive analysis on the behavior of the (A)SGD output. 
From the iterative scheme \eqref{SA iteration}, we obtain
\begin{align}\label{eqt: sgd closed form}
\begin{split}
    x_{n+1} = B_{0n}x_1 - \sum_{m=1}^n\eta_m B_{mn}\delta_m - \sum_{m=1}^n \eta_m B_{mn} E_m,
    \end{split}
\end{align}
where $\delta_k \triangleq \delta(x_k) = \nabla H(x_k) - G(x^*) (x_k-x^*)$ is the second-order residual of the Taylor expansion of $\nabla H$ at $x^*$, $E_{k-1} = \nabla h(x_{k-1}, \zeta_k) - \nabla H(x_{k-1})$, and $B_{mn} = \prod_{j=m+1}^n (I-\eta_j G(x^*)) \in \mathbb{R}^{d\times d}$.

In the ASGD case, from \eqref{eqt: sgd closed form} we show that there exist $\hat{\tau_0}, \hat{\tau}$, and $\hat{C}$ such that for any $\delta > 0$, there is integer $N$ satisfying
\begin{align}
\begin{split}
\label{eqt: ASGD COfB inner bound uniform}
    \sup_{n>N}|\mathbb{P}^*(\sqrt{n}\left(\psi_n(\hat{P}_n) - \psi(\hat{P}_n)\right)\in D) - 
    % \mathbb{P}(\sigma(\hat{P}_n)Z\in D)|
    \mathbb{P}^*(\hat{Z}_n \in D)|
    \\ \leq  \hat C(d^{3/2} + \hat \tau^3 + \hat \tau_0^3) d^{\frac{1}{2}} n^{-\alpha + \frac{1}{2} + \epsilon},
    \end{split}
\end{align}
with probability at least $1-\delta$, for any measurable $D$ and $\epsilon > 0$. To achieve this, we generalize the result in
\cite{shao2022berry}, who gave an inequality similar to \eqref{eqt: ASGD COfB inner bound uniform} but with a fixed distribution instead of a varying distribution, to establish uniform rates across all data distributions including the empirical distribution $\hat{P}_n$. 
Detailed proof for \eqref{eqt: ASGD COfB inner bound uniform} can be found in \Cref{sup: ASGD}.

% and using results from \cite{shao2022berry},
 % $\hat{P}_n$
% For the SGD case, the first term $B_{0n}x_1$ of \eqref{eqt: sgd closed form} consists of the deviation caused by the initial guess and will converge to 0 since $B_{0n}$ vanishes as $n$ grows. $\sum_{m=1}^n\eta_m B_{mn}\delta_m $ consists of the higher-order term after the second-order derivative of $H$ in the Taylor series expansion, which will also be small under proper smoothness assumption on $H$. 
In the SGD case, the first two terms in \eqref{eqt: sgd closed form} correspond to the interaction of the error of the initial solution and the second-order residual in the Taylor expansion of $\nabla H$. One can show the following vanishing property
\[
\lim_{n\rightarrow\infty}\mathbb{E} [\|B_{0n}x_1 - \sum_{m=1}^n\eta_m B_{mn}\delta_m\|] = 0.
\]
On the other hand, the last term $\sum_{m=1}^n \eta_m B_{mn} E_m$ consists of the difference between sample gradient $\nabla h$ and true gradient $\nabla H$, which converges to a normal distribution. 
% In \cite{sacks1958asymptotic}, a martingale Central Limit Theorem is applied to establish the asymptotic normality for this term. However, the result is not desirable in our work since it lacks information on the speed of convergence. 
We use the Berry-Esseen-type result from Corollary 2.3 in \cite{shao2022berry} to give a non-asymptotic convergence result for this term and thus establish a bound similar to \eqref{eqt: ASGD COfB inner bound uniform} that is uniform across all data distributions including the empirical distribution. Details can be found in \Cref{sup: SGD}.
% Details are omitted here.

% For COnB, the desired result \eqref{eqt: classical bootstrap theorem1} is well established in \cite{JMLR:v19:17-370}, which states as follows
%     \begin{equation}
%     \sup_{v\in\mathbb{R}^d} |\mathbb{P}^*(\sqrt{n}(x_\text{out}^* - x_\text{out}) \leq v) - \Phi(v)| \rightarrow 0, \text{   in probability}
% \end{equation}
% \begin{equation}
%     \sup_{v\in\mathbb{R}^d} |\mathbb{P}(\sqrt{n}(x_\text{out} - x^*) \leq v) - \Phi(v)| \rightarrow 0, \text{   in probability}
% \end{equation}
% where $\Phi$ denotes the distribution function of $Z\sim \mathcal{N}(0, \sigma^2)$.

\subsection{From Conditional Convergence to Asymptotic Independence}
The results in \Cref{thm:nonasymptotic bound} imply that the error of the original estimate is asymptotically independent of the errors of the resample estimates. This implication, which follows a similar idea in the cheap bootstrap \citep{lam2022cheap}, can be stated as follows:
\begin{theorem}
\label{thm: asymptotic result}
Supposing \eqref{eqt: SA asymptotic convergence} and \eqref{eqt: classical bootstrap theorem1} hold, we have
\begin{equation}\label{eqt: Thm2 COnB}
    \sqrt{n}\left(\begin{array}{l}
        \bar{x}_n -x^* \\
        x^{*1}-\bar{x}_n \\
        \vdots \\
        x^{*B}-\bar{x}_n
    \end{array}\right)
    \xrightarrow[n\rightarrow\infty]{d}
    \left(\begin{array}{l}
          Z^\text{ASGD}_0 \\
          Z^\text{ASGD}_1 \\
         \vdots \\
          Z^\text{ASGD}_B
    \end{array}
    \right).
\end{equation}
For COfB running SGD, supposing \eqref{eqt: SA asymptotic convergence} and \eqref{eqt: classical bootstrap theorem} hold, we have
\begin{equation}\label{eqt: Thm2 COfB}
    \sqrt{n}\left(\begin{array}{l}
        x_n -x^* \\
        x^{*1}_\text{COfB}-\hat{x}_n \\
        \vdots \\
        x^{*B}_\text{COfB}-\hat{x}_n
    \end{array}\right)
    \xrightarrow[n\rightarrow\infty]{d}
    \left(\begin{array}{l}
          Z^\text{SGD}_0 \\
          Z^\text{SGD}_1 \\
         \vdots \\
          Z^\text{SGD}_B
    \end{array}
    \right).
\end{equation}
$\{Z^\text{ASGD}_b\}_{b=0}^B$ and $\{Z^\text{SGD}_b\}_{b=0}^B$ are $i.i.d.$ copies of $Z^\text{ASGD}$ and $Z^\text{SGD}$ described in \Cref{thm:nonasymptotic bound}, respectively.
\end{theorem}

\begin{proof}  
Let $\Delta$ and $\Delta^*$ denote $\sqrt{n}(\bar{x}_n - x^*)$ and $\sqrt{n}(x^{*1}-\bar{x}_n)$ respectively. Define $\Phi(z) = P\left((Z^\text{ASGD})_1 \leq z_1, \dots, (Z^\text{ASGD})_d \leq z_d\right)$ where $z=(z_1,\ldots,z_d)\in\mathbb R^d$. To simplify notation, $\Delta \leq x$ is defined componentwise, i.e., $\Delta_i \leq x_i,\  \forall i=1,\dots,d$.
For any $x, y \in \mathbb{R}^d$,
% \begin{align*}
%     & |\mathbb{P}(\Delta \leq x, \Delta^* \leq y) -  \Phi (x) \Phi (y)| \\
%    = & |\mathbb{E}\left[\mathbb{E}[I(\Delta \leq x, \Delta^* \leq y)|\{\zeta_i\}_{i=1}^n]\right] -  \Phi (x) \Phi (y)| \\
%    = & |\mathbb{E}\left[I(\Delta \leq x)\mathbb{E}[I(\Delta^* \leq y)|\{\zeta_i\}_{i=1}^n]\right] -  \Phi (x) \Phi (y)| \\
%    \leq & |\mathbb{E}\left[I(\Delta \leq x)\mathbb{E}[I(\Delta^* \leq y)|\{\zeta_i\}_{i=1}^n] - \Phi(y)\right]| + |\mathbb{P}(\Delta \leq x) - \Phi(x)|\Phi(y)\\
%    \leq &  |\mathbb{E}\left[\mathbb{E}[I(\Delta^* \leq y)|\{\zeta_i\}_{i=1}^n] - \Phi(y)\right]| + |\mathbb{P}(\Delta \leq x) - \Phi(x)|.
% \end{align*}

\begin{align*}
    & \big|\mathbb{P}(\Delta \leq x, \Delta^* \leq y) -  \Phi (x) \Phi (y)\big| \\
   = & \big|\mathbb{E}\big[\mathbb{E}[I(\Delta \leq x, \Delta^* \leq y)|\{\zeta_i\}_{i=1}^n]\big] -  \Phi (x) \Phi (y)\big| \\
   = & \big|\mathbb{E}\big[I(\Delta \leq x)\mathbb{E}[I(\Delta^* \leq y)|\{\zeta_i\}_{i=1}^n]\big] -  \Phi (x) \Phi (y)\big| \\
   \leq & \big|\mathbb{E}\big[I(\Delta \leq x)\big[\mathbb{E}[I(\Delta^* \leq y)|\{\zeta_i\}_{i=1}^n] - \Phi(y)\big]\big]\big| + \big|\mathbb{P}(\Delta \leq x) - \Phi(x)\big|\Phi(y)\\
   \leq & \mathbb{E}\big[ \big|\mathbb{E}[I(\Delta^* \leq y)|\{\zeta_i\}_{i=1}^n] - \Phi(y)\big|\big] + \big|\mathbb{P}(\Delta \leq x) - \Phi(x)\big|.
\end{align*}
% \\
%    \to&0
The last line in the above relationship vanishes since we assume \eqref{eqt: SA asymptotic convergence} and \eqref{eqt: classical bootstrap theorem1}. Thus $\mathbb{P}(\Delta \leq x, \Delta^* \leq y)\to  \Phi (x) \Phi (y)$ for any $x,y\in\mathbb R^d$. Notice that conditional on $\{\zeta_i\}_{i=1}^n$, the bootstrap replications are independent. So we can generalize the above arguments for $B>1$ by replacing $\mathbb{E}[I(\Delta^* \leq y)|\{\zeta_i\}_{i=1}^n]$ by a product of $B$ conditional expectations, each of them on a bootstrap replication of $I(\Delta^* \leq y)$ for a different $y$.
Hence we obtain \eqref{eqt: Thm2 COnB}.
Lastly, the second case in the theorem follows the same argument as above.
\end{proof}

The reason why we introduce the implications \eqref{eqt: Thm2 COnB} and \eqref{eqt: Thm2 COfB} from \Cref{thm:nonasymptotic bound} is to utilize them in a different manner from the classical bootstrap. In particular, instead of using \Cref{thm:nonasymptotic bound} to reveal the closeness between the sampling and resample distributions followed by Monte Carlo resample approximation, we will use the sample-resample joint distribution in \eqref{eqt: Thm2 COnB} and \eqref{eqt: Thm2 COfB} to construct pivotal statistics directly for inference. This latter idea, which comes from the so-called cheap bootstrap, can substantially reduce the requirement on $B$ as we describe in the next subsection. 
% The proof of Theorem \ref{thm: asymptotic result} is similar to but also generalizes the proof of Proposition 1 in \cite{lam2022cheap}. presents a bootstrap methodological route that

\subsection{From Asymptotic Independence to the Cheap Bootstrap}
By putting together \Cref{thm:nonasymptotic bound,thm: asymptotic result}, we immediately conclude that, under the assumptions in \Cref{thm: main theorem}, the error of a resample run (compared against $\bar{x}_n$ or $\hat x_n$) and the error of the original (A)SGD run (compared against $x^*$) are asymptotically independent and follow the same Gaussian distribution. This subsequently allows us to construct asymptotic pivotal $t$-statistics that can be converted into coverage-exact confidence intervals. We present this argument in the proof of \Cref{thm: main theorem} below.
\\

% have the following result:
% \begin{corollary}
   
% \end{corollary}

% summarizes the results in the previous two subsections, which involve two aspects.
% First, the asymptotic distribution of the error of the resample run compared with the SAA solution in COfB, namely, $x^{*b}_\text{COfB}-\hat{x}_n$ (or compared with the original ASGD run in COnB, $x^{*b}_\text{COnB}-x_\text{out}$) is the same as that of the original run compared with the true minimizer, $x_\text{out} - x^*$. Second, more importantly, is the asymptotic independence among all these errors. 
 % With this result, one can construct the confidence intervals with center $x_\text{out}$ and length obtained from $B$ resample runs.
 
\begin{proof}[\textbf{Proof of \Cref{thm: main theorem}}]
Under the assumptions in \Cref{thm: main theorem}, we have from \Cref{thm:nonasymptotic bound,thm: asymptotic result} that \eqref{eqt: Thm2 COnB} and \eqref{eqt: Thm2 COfB} hold. Now, let $x^{*b}$ denote $x_\text{COfB}^{*b}$ and $x_\text{COnB}^{*b}$ in \Cref{Alg: COfB,alg: COnB} respectively. In these cases \eqref{eqt: Thm2 COnB} applies. Observe that
\begin{align*}
    \frac{(\bar{x}_n)_i - x^*_i}{s_i^{\text{ASGD}}} 
& =\frac{\sqrt{n}\left((\bar{x}_n)_i - x^*_i\right)}{\sqrt{\frac{\sum_{b=1}^B\left(\sqrt{n}\left(x_i^{*b} - (\bar{x}_n)_i\right)\right)^2}{B}
}}.
\end{align*}
Taking $n\rightarrow \infty$, we have
\begin{align*}
    \frac{\sqrt{n}\left((\bar{x}_n)_i - x^*_i\right)}{\sqrt{\frac{\sum_{b=1}^B\left(\sqrt{n}\left(x_i^{*b} - (\bar{x}_n)_i\right)\right)^2}{B}
}}
\stackrel{d}{\to}
\frac{(Z^\text{ASGD}_0)_i}{\sqrt{\frac{\sum_{b=1}^B (Z^\text{ASGD}_b)_i^2}{B}}}
\stackrel{d}{=}
\frac{N}{\sqrt{\frac{\chi_B^2}{B}}}
\stackrel{d}{=}t_B,
\end{align*}
where $N$ stands for a standard normal variable,  $\chi_{B}^2$ a $\chi^2$-variable with $B$ degree of freedom, $t_{B}$ a student $t$-variable with $B$ degree of freedom, and ``$\stackrel{d}{=}$'' equality in distribution. The convergence in distribution above comes from the continuous mapping theorem. The first equality in distribution comes from the \emph{i.i.d.} normality limit in \Cref{thm: asymptotic result} and the elementary relation between $\chi^2$ and normal. The second equality in distribution comes from the elementary construction of a $t$-variable. Thus, by a pivotal argument, we obtain the asymptotic exact coverage of the confidence intervals generated from \Cref{Alg: COfB,alg: COnB}.
% $\bar Z=(1/B)\sum_{b=1}^BZ_b$, and 

A similar argument works for COfB using SGD. In this case \eqref{eqt: Thm2 COfB} applies, and we use the average of $x_\text{COfB}^{*b}$ across $b$, denoted $\overline{x}_\text{COfB}^{*}$, in our pivotal construction. This would result in a student $t$-distribution with degree of freedom $B-1$. More precisely, we have 
\begin{align*}
\frac{(x_n)_i - x^*_i}{s_i^{\text{SGD}}} 
% & = \frac{\sqrt{n}\left((x_\text{out})_i - x^*_i\right)}{\sqrt{n\times (s_i^{\text{COfB}})^2}}\\
 &=\frac{\sqrt{n}\left((x_n)_i - x^*_i\right)}{\sqrt{n\times\frac{\sum_{b=1}^B\left((x_\text{COfB}^{*b})_i - (\overline{x}_\text{COfB}^{*})_i\right)^2}{B-1}
}}\\
 &=\frac{\sqrt{n}\left((x_n)_i - x^*_i\right)}{\sqrt{\frac{\sum_{b=1}^B\left(\sqrt n\left((x_\text{COfB}^{*b})_i-(\hat x_n)_i\right) - \sqrt n\left((\overline{x}_\text{COfB}^{*})_i-(\hat x_n)_i\right)\right)^2}{B-1}
}}.
\end{align*}
As $n\to\infty$, we have
\begin{eqnarray*}
\frac{\sqrt{n}\left((x_n)_i - x^*_i\right)}{\sqrt{\frac{\sum_{b=1}^B\left(\sqrt n\left((x_\text{COfB}^{*b})_i-(\hat x_n)_i\right) - \sqrt n\left((\overline{x}_\text{COfB}^{*})_i-(\hat x_n)_i\right)\right)^2}{B-1}
}}
\stackrel{d}{\to} \frac{(Z^\text{SGD}_0)_i}{\sqrt{\frac{\sum_{b=1}^B\left((Z^\text{SGD}_b)_i-(\bar Z)_i\right)^2}{B-1}}}\stackrel{d}{=}\frac{N}{\sqrt{\frac{\chi_{B-1}^2}{B-1}}}\stackrel{d}{=}t_{B-1},
\end{eqnarray*}
where we note that the $\chi^2$ and $t$-distributions now have degrees of freedom $B-1$, and $\bar Z$ denotes $(1/B)\sum_{b=1}^BZ_b^\text{SGD}$. A pivotal argument gives rise to the asymptotic exactness of the confidence interval generated from COfB using SGD. 
\end{proof}

We note that the distinction between using $t_B$ and $t_{B-1}$, in the intervals constructed in \Cref{Alg: COfB,alg: COnB}, and in COfB using SGD respectively, stems from the subtle difference in the large-sample asymptotics described in \Cref{thm: asymptotic result}. For \Cref{Alg: COfB,alg: COnB}, the center in the sample variance is $\bar{x}_n$, which is the known outcome from the original SGD run. In contrast, for COfB using SGD, $\hat{x}_n$ is the optimizer of SAA, which is unknown. Thus, when calculating the sample variance for confidence intervals, COfB using SGD needs to set the sample mean as its center and consumes one degree of freedom. Consequently, \Cref{Alg: COfB,alg: COnB} can use $B$ at least 1, while COfB using SGD requires $B$ to be at least 2.

Our COnB procedure leverages the online bootstrap method of \cite{JMLR:v19:17-370}, notably in maintaining multiple ASGD trajectories in parallel, and that each bootstrap run is generated by perturbing the stochastic gradient at each iteration $t$ by a random multiplicative factor $W_{t,b}$. However, there is a fundamental difference in the way we construct the confidence intervals that crucially allows us to substantially reduce computation effort. \cite{JMLR:v19:17-370} approximate the sampling distribution of the ASGD estimator through the empirical distribution of the $B$ bootstrap outputs, and construct confidence intervals using quantiles or variance estimates from this empirical distribution. This way of constructing intervals follows the conventional bootstrap route. Consequently, the coverage accuracy of their online bootstrap method depends not only on the asymptotic validity of the bootstrap approximation (when $n$ is large), but also on the Monte Carlo accuracy of bootstrapping, which in turn requires a large $B$. In contrast, our COnB leverages the cheap bootstrap principle, using the asymptotic independence among original and bootstrap runs to construct a pivotal $t$-statistic, thus allowing a fixed (and small) $B$ while still achieving asymptotic exact coverage. In this case, $B$ only serves to reduce the width of the interval, rather than being required for asymptotic validity.

\section{Application in Sparse Linear Regression}
\label{sec: sparse linear regression}

In the preceding sections, we discussed the case when the dimension of the stochastic optimization problem, $d$, is fixed as the sample size $n\rightarrow \infty$. In many modern applications, the ambient dimension $d$ of the parameter space can be very large relative to the available data. In this section, we discuss a way to extend our methods to high-dimensional but sparse settings.

Throughout this section, we distinguish between the ambient dimension $d$ and the effective dimension $p$. The ambient dimension $d=d(n)$ can be much larger than $n$, and may grow with $n$. The true underlying parameter is assumed to be sparse, with only a fixed number $p$ of non-zero components. Under this setting, our methods proceed by first reducing the problem to a lower-dimensional subspace of size $p$ via model selection, then applying our theory to this reduced space.

To be more concrete, we consider a sequence of regression problems indexed by $n$, where the ambient dimension $d=d(n)$ scales in the number of observations, whereas the effective dimension stays constant at $p$:
\[
b^{(n)} = A^{(n)} x^{(n)} + \epsilon^{(n)}.
\]
Here, superscript $(n)$ indicates dependence on the sample size $n$.
$\epsilon^{(n)} \in\mathbb{R}^n$ consists of $n$ $i.i.d.$ entries, $b^{(n)}\in\mathbb{R}^n$ represents the vector of $n$ responses, $A^{(n)}\in\mathbb{R}^{n\times d}$ encodes the matrix of $n$ feature vectors each of length $d$, and $x^{(n)} = [x_1^{(n)}, \dots, x_d^{(n)}]^\top \in \mathbb{R}^{d}$ has $p$ nonzero entries. 
Given an index set $T$ and a vector $v$, $(v)_T$ denotes the $|T|$-dimensional subvector of $v$ consisting of entries of $v$ with indices in $T$. Let
$x^* := (x^{(n)})_{T^*}= [x^*_1, \dots, x^*_p]^\top$ be the non-zero entries of the true model coefficients, and we assume $x^*$ is not dependent on $n$.
For matrix $A$ and index sets $T_1$ and $T_2$, $A_{T_1, T_2}$ denotes the $|T_1| \times |T_2|$ submatrix of $A$ with row indices inside $T_1$ and column indices inside $T_2$.
Then, $A^{(n)}_1\colon = A_{\mathbb{N}, T^*} \in\mathbb{R}^{n\times p}$ is the submatrix of $A^{(n)}$ consisting of columns with indices in $T^*$ and $A^{(n)}_2\colon = A_{\mathbb{N}, T^{*C}}\in\mathbb{R}^{n\times(d-p)}$ consists of columns with indices outside $T^*$.

% Under this setting, 
% THE REVIEWER ASKED IF WE HAVE INSIGHTS ON WHEN OUR METHOD WORKS AND NOT WORK.. HERE, IT SEEMS WE ONLY EXPLAIN WHEN OUR METHOD WOULD NOT WORK, IN THEORY. CAN WE DISCUSS WHEN OUR METHOD WOULD WORK IN THEORY, AND ALSO WHEN OUR METHOD WORKS OR NOT WORK FROM EMPIRICAL FINDINGS? IS IT IN OUR NUMERICAL SECTION? WE DID NOT RESPOND TO THAT POINT IN OUR CURRENT RESPONSE LETTER. NOTE: IF OUR NUMERICAL RESULTS IN THE PREVIOUS VERSION ALREADY HAVE THE DISCUSSION ON THE PERFORMANCE IN HIGH-DIMENSIONAL PROBLEMS THAT THE REVIEWER ASKED, THEN WE DON'T NEED EXTRA NUMERICS, BUT NEED TO TELL THE REVIEWER WHERE WE PUT AND WHY IT'S ALREADY SUFFICIENT) 

Our methods described in the previous sections under a fixed-dimensional asymptotic regime are not directly applicable in this setting, as the dimension $d$ of this problem is no longer fixed as $n \rightarrow \infty$. 
In particular, the non-asymptotic bound in \eqref{eqt: ASGD COfB inner bound uniform} deteriorates with increasing $d$. Inspecting \eqref{eqt: ASGD COfB inner bound uniform}, the leading dimension dependent term in the bound scales as $d^2 n^{-\alpha + \frac{1}{2} + \epsilon}$.
% (THIS COMES FROM NOWHERE. CITE PREVIOUS WORK, OR DISCUSS HOW WE GOT THIS?)
Consequently, the right hand side of \eqref{eqt: ASGD COfB inner bound uniform} vanishes only if $d(n)$ grows sufficiently slowly relative to $n$, i.e.,
$\lim_{n\rightarrow\infty}d(n)^2 n^{-\alpha + \frac{1}{2} + \epsilon} = 0$. If this latter condition fails, the bound in \eqref{eqt: ASGD COfB inner bound uniform} no longer guarantees small resampling error, and in this regime we should not expect theoretical control from our previous arguments. This in particular rules out validity when $d$ grows faster than $n^{\alpha/2 - 1/4}$.

% Conversely, when $d(n)$ grows slowly enough to satisfy the above rate, it is sill open whether our presented results are valid, because the rest of our theoretical development is tailored to fixed $d$. Proving guarantees for diverging dimension would require a future (and interesting) high-dimensional analysis.
 
% . That is, the bound requires
% % (AGAIN, PLEASE CITE OR DISCUSS)
% , which rules out regimes where $d$ grows faster than $n^{\alpha/2 - 1/4}$.

A feasible approach to handle this challenge is to first transform the problem into a lower dimension by Lasso model selection techniques \citep{zhao2006model, 10.3150/11-BEJ410}, which produces a sparse solution by solving the following problem:
% \begin{equation}
% \label{eqt: lasso}
% \hat{x}^{(n)}(\lambda) = \arg\min_{x} \|b^{(n)} - A^{(n)\top} x\|^2 + \lambda \sum_{i=1}^d |x_i|,
% \end{equation}
\begin{equation}
\label{eqt: lasso}
\hat{x}^{(n)}(\lambda) = \arg\min_{x} \|b^{(n)} - A^{(n)} x\|^2 + \lambda \sum_{i=1}^d |x_i|,
\end{equation}
where $\lambda\geq 0$ is the regularization parameter that controls the sparsity. Let $\mathcal T$ be the support function of a $d$-dimensional vectors, i.e., for a vector $x\in\mathbb{R}^d$, $\mathcal T(x) \colon = \{i: x_i \neq 0\}$. The estimated support corresponding to parameter $\lambda$ is $\mathcal T(\hat{x}^{(n)}(\lambda))$, and the true support $T^* = \mathcal T(x^{(n)})$.
% {\color{blue}The support of $\hat{x}^{(n)}$ is given by $(\hat{x}^{(n)})_T = \{i: (\hat{x}^{(n)})_i \neq 0\}$ (THE NOTATIONS ARE CONFUSING, WHAT'S $T$? IS THAT THE TRUE SUPPORT OR ESTIMATED SUPPORT? IN FACT, YOU WROTE $T(\hat{x}^{(n)})$ WHICH I GUESS WAS A TYPO. WHAT'S $\hat x^{(n)}$? YOU'VE ONLY DEFINED $\hat x^{(n)}(\lambda)$. SIMILAR ISSUES IN THE ALGORITHM. YOU SOMETIMES USE $T_n$ AND SOMETIMES $T$. ALSO, YOU SHOULD CALL $T_n$ THE ESTIMATED SUPPORT INSTEAD OF "SUPPORT OF TRUE MODEL" IF I UNDERSTAND CORRECTLY?).} 
Under assumptions to be specified, the difference between estimated support and true support vanishes \citep{zhao2006model}. 
After the model selection step, our methods can be adapted and applied to the problem confined to the support of $\hat{x}^{(n)}$. Specifically, apply COnB or COfB to the problem given by \eqref{eqt: stochastic optimization problem} and \eqref{SA iteration}. In this context, the $i$-th data is the vector corresponding to $b_i^{(n)}$ and entries in $\mathcal T(x^{(n)})$ for the $i$-th row of $A^{(n)}$. 
% Given an index set $T \subset \mathbb{Z}_{\geq 0}$, define the linear regression problem \textbf{confined on $T$} by
% \begin{equation}
%     \label{eqt: linear regression confined on T}
%     \min_{x \in \mathbb{R}^{|T|}} \mathbb{E} [(a_{T}^\top x- b)^2]
% \end{equation}
A pseudocode can be found in \Cref{alg: two-stage}.

\begin{algorithm}
\caption{Two-stage Method for Sparse Linear Regression}
\label{alg: two-stage}
\begin{algorithmic}
    \STATE {\bfseries Input: } $i.i.d.$ data $\{\zeta_t\}_{t=1}^n$, number of bootstrap runs $B\geq 1$, step size sequence $\{\eta_t\}$, initial guess $x_0$, nominal coverage level $1-\gamma$, Lasso regularization parameter $\lambda_n$. 
    % (DON'T WRITE PERTRUBATION FACTOR W AS INPUTS. INSTEAD, PLEASE WRITE THEM IN THE ALGO BELOW, AND SAY EXACTLY HOW THEY ARE GENERATED.)
    \STATE {\bfseries Output: }confidence intervals $\mathcal{I}_{i,n},\ i=1,\dots,d$.
    \STATE Solve \eqref{eqt: lasso} to obtain $\hat{x}^{(n)}(\lambda_n)$.
    \STATE $T_n \leftarrow \{i: (\hat{x}^{(n)}(\lambda_n))_i \neq 0\}$
    \STATE $\mathcal I_{i,n} \leftarrow \{0\}$ for all $i\notin T_n$
    \STATE Obtain confidence intervals on $T_n$ by running \Cref{Alg: COfB,alg: COnB} with input: $i.i.d.$ data $\{(\zeta_t)_{T_n}\}_{t=1}^n$, number of bootstrap runs $B$, step size sequence $\{\eta_t\}$, initial guess $x_0$, nominal coverage level $1-\gamma$.
\end{algorithmic}
\end{algorithm}
% , estimated support of the model $T_n$

We put forth the following assumptions to support the efficacy of \Cref{alg: two-stage}. 

\begin{assumption}
    \label{assumption: lasso correctness}
    Entries of $\epsilon^{(n)}$ are $i.i.d.$ and bounded. 
    Diagonal entries of the sample covariance matrix $\frac{1}{n}A^{(n)^\top}A^{(n)}$ are upper bounded by constant $M_1$ not depending on $n$. The submatrix $\frac{1}{n}A^{(n)^\top}_1A^{(n)}_1$ is positive definite. And as $n\rightarrow\infty$, the regularization parameter satisfies $\lambda_n/n \rightarrow 0$ and $\lambda_n / \sqrt{n} \rightarrow \infty$.
\end{assumption}

These assumptions are mild in the context of linear regression. 
The boundedness of $\epsilon$ and the diagonal entries of the covariance matrix are typical results of the common data normalization procedure.
Positive definiteness of $\frac{1}{n}A^{(n)^\top}_1A^{(n)}_1$ ensures the identifiability of the model. 
In addition to \Cref{assumption: lasso correctness}, we need one further assumption to guarantee the correctness of the selected model by Lasso.

\begin{assumption}
\label{assumption: strong irrepresentable condition}
    There exists a positive constant $\eta$ such that for all $i=1,...,d-p$, we have
    \[
    |(A^{(n)\top}_{2} A^{(n)}_1 (A^{(n)^\top}_1A^{(n)}_1)^{-1}\text{sign}(x^*))_i| \leq 1-\eta,
    \]
    where $\text{sign}(x^*)\in\mathbb{R}^p$ denotes the entrywise sign function for $x^*$.
\end{assumption}

\Cref{assumption: strong irrepresentable condition} is referred to as the irrepresentable condition in \cite{zhao2006model}, which turns out to relate closely to the consistency of Lasso.
Under \Cref{assumption: lasso correctness,assumption: strong irrepresentable condition}, the probability of Lasso selecting the correct model converges to 1 as data size $n \rightarrow \infty$ \citep{zhao2006model}. 

\begin{assumption}
\label{assumption: confined problem assumption}
Rows of $A^{(n)}_1$ are bounded and $i.i.d.$ following distribution $P_a$ such that $\mathbb{E}_{a\sim P_a} [a a^\top]\in\mathbb{R}^{p\times p}$ is positive definite with eigenvalues greater than $l>0$. Rows of $A^{(n)}_2$ are bounded and $i.i.d.$.
% Denoting $P_b$ as the distribution of entries of $b^{(n)}$, we have
% \[
% \sup_{x\in\mathbb{R}^p} |\frac{1}{n} \sum_{i=1}^n \frac{1}{2} (A^{(n)\top}_{i, T^*} x - b^{(n)}_i)^2 - \mathbb{E}_{a\sim P_a, b\sim P_b} [\frac{1}{2}(a^\top x - b)^2]| \xrightarrow{P} 0
% \]
\end{assumption}

The above assumptions are sufficient to guarantee \Cref{assumption: strongly convex objective with bounded hessian,assumption: unbiased gradient estimation,assumption: gradient variability,assumption: m estimator consistency} for the linear regression problem confined to $T^*$ which, together with one of the following two assumptions, guarantees the correctness of coverage probability of confidence intervals generated after the model selection step.

\begin{assumption}
    \label{assumption: confined problem assumption ASGD}
For COfB running ASGD or COnB, the step size $\eta_t = \eta t^{-\alpha}$ for some $\alpha \in (\frac{1}{2}, 1]$. 
\end{assumption}

% {\color{blue}
% [THE FOLLOWING \Cref{assumption: confined problem assumption SGD ver1} IS ONE POSSIBLE ASSUMPTION CORRESPONDING TO \Cref{assumption: SGD step size} FOR SGD CASE. [VERSION 1]]
\begin{assumption}
\label{assumption: confined problem assumption SGD}
For COfB running SGD, the step size $\eta_t = \frac{\eta}{t}$ such that $\eta l > \frac{1}{2}$. 
% Let $\mathcal{X}_1$ be a bounded subset of $\mathbb{R}^p$ containing $x^*$ in its interior and $\mathcal{X} = \{x | \sup_{y\in\mathcal{X}_1} \|x-y\|\leq \epsilon_1\}$ for some $\epsilon_1>0$. Define $l(x_1, x_2, a, b) = a (a^\top x_1 - b)  - \sum_{j=1}^p a a_j (x_1 - x_2)_j$. There is $\epsilon > 0$ such that
% \[
% || \leq \|x_1- x_2\|^{1+\epsilon}
% \]
\end{assumption}

% \begin{assumption}
% \label{assumption: confined problem assumption SGD}
% Consider COfB running SGD, the step size $\eta_t = \frac{\eta}{t}$ such that $\eta l > \frac{1}{2}$. Let $\mathcal{X}_1$ be a bounded subset of $\mathbb{R}^p$ containing $x^*$ in its interior and $\mathcal{X} = \{x | \sup_{y\in\mathcal{X}_1} \|x-y\|\leq \epsilon_1\}$ for some $\epsilon_1>0$. For $i=1,\dots,p$, $\tilde{F}_i = \{\left(\left((a^\top x_1 - b) a\right)_i - \sum_{j=1}^p a_i a_j (x_1 - x_2)\right) / \|x_1 - x_2\| | x_1 \in \mathcal{X}, x_2 \in \mathcal{X}_1, x_1 \neq x_2\}$ are $(P_a, P_b)$-Glivenko-Cantelli, and $\mathbb{P}(x_t = x^*) = 0, \ \forall t$.
% \end{assumption}
% } 
With these assumptions in place, we are ready to state the theoretical guarantee for \Cref{alg: two-stage}.

\begin{theorem}
\label{thm: two-stage guarantee}
    In the sparse linear regression setting, suppose \Cref{assumption: lasso correctness,assumption: strong irrepresentable condition,assumption: confined problem assumption}, and one of \Cref{assumption: confined problem assumption ASGD,assumption: confined problem assumption SGD} hold.
    % {\color{blue}Further suppose the assumptions for \Cref{thm: main theorem} hold for the problem confined on $T^*$. Here, confining on $T^*$ means the problem given by \eqref{eqt: stochastic optimization problem,SA iteration} with $P$ being the distribution generating the $n$ $i.i.d.$ data $\{\zeta_j = (A^{n}_{j, T^*}, b^{(n)}_j)\}_{j=1}^n$. }
    % (THIS NEEDS TO BE MORE EXPLICIT. DOES IT MEAN YOU IMPOSE ASSUMPITONS ON $A^{(n)}$ ETC? CAN ONE KNOW READILY IF YOUR ASSUMPTIONS HOLD?).
    For $i=1,\dots,d$, the $1-\gamma$ confidence interval produced by \Cref{alg: two-stage} satisfies
    \[
    \lim_{n\rightarrow\infty} \mathbb{P}(x^{(n)}_i \in \mathcal{I}_{i,n}) = \begin{cases}
        1 & \text{if }\ i\notin T^*\\
        1-\gamma & \text{if }\ i\in T^*
    \end{cases}.
    \]
\end{theorem}

\Cref{thm: two-stage guarantee} states that, as $n\rightarrow\infty$, \Cref{alg: two-stage} correctly identifies the support of the true model parameter $T^*$ and provides confidence intervals with exact coverage for non-zero entries of $x^{(n)}$. It is worth noticing that, although COnB could work as the second stage of \Cref{alg: two-stage}, the whole procedure is no longer suitable for the online setting as the model selection part requires all the $n$ data. If new data comes, one needs to solve the Lasso again. The proof for \Cref{thm: two-stage guarantee} can be found in \Cref{subsection: two-stage proof}.

In \cite{chen2020statistical}, the authors also discussed inference for high-dimensional linear regression setting. In their work, they developed a plug-in method that estimates the true coefficient and the inverse of the covariance matrix at the same time, using the Regularization Annaled epoch Dual AveRaging (RADAR) algorithm \citep{agarwal2012stochastic}, a variant of SGD. Confidence intervals with a nominal level $1-\gamma $ for each entry of the true coefficient are the outputs of their algorithm, and they have a theoretical guarantee similar to that of \Cref{thm: main theorem}. In comparison, our approach for the sparse linear regression provides more information since it also gives asymptotically correct support of the relevant coefficients, as described in \Cref{thm: two-stage guarantee}. On the other hand, the method in \cite{chen2020statistical} could handle new data with small additional computations, while ours, as discussed earlier, only works under the offline setting.

\section{Experiments}\label{section: experiments}

In this section, we illustrate the numerical performance of our approaches and compare them with the other methods in the regression setting. The code to reproduce all experiments in this paper is publicly available at \href{https://github.com/BeetrootWang/Cheap-Bootstrap-for-Fast-Uncertainty-Quantification-of-Stochastic-Gradient-Descent}{https://github.com/BeetrootWang/Cheap-Bootstrap-for-Fast-Uncertainty-Quantification-of-Stochastic-Gradient-Descent}.
% {GitHub}.
% through numerical experiments. 

\subsection{Regression Problems with Fixed Dimensionality}
We consider two sets of problems:

\paragraph{Linear Regression}
We consider the linear regression problem of dimension $d$. The data $\zeta=(a,b)$ with distribution $P$ consists of the independent variable $a\in\mathbb{R}^d$ and dependent variable $b\in\mathbb{R}$. In this case, 
$h(x,\zeta) = \frac{1}{2}(a^\top x - b)^2.$ $a$ follows a multivariate normal distribution $\mathcal{N}(0, \Sigma)$. Let $x^*$ be the true regression coefficient.
Then, $b$ satisfies the model $b=a^\top x^* + \epsilon$ for some error term $\epsilon$ that is assumed to have normal distribution $\mathcal{N}(0,\sigma^2)$.
In this experiment, $\epsilon$ and $a$ are independent. And we have
$
\psi(P) = x^*$, $ S = \sigma^2 \Sigma$, and $G = \Sigma$.

\paragraph{Logistic Regression} 
Similar to the linear regression setup, the data $\zeta=(a,b)$ coming from distribution $P$ consists of the independent variable $a\in\mathbb{R}^d$ and dependent variable $b\in\{-1,1\}$. $h(x,\zeta) = \log{(1+e^{-b\times a^\top x})},
$ where $a$ follows a multivariate normal distribution $\mathcal{N}(0,\Sigma)$, and $b=1$ with probability $\frac{1}{1+e^{-a^\top x^*}}$, where $x^*$ denotes the true regression coefficient. In this case, we have
$
\nabla h(x,\zeta) = \frac{-b\times a}{1+e^{b\times a^\top x}}$ and $ \nabla^2 h(x,\zeta) = \frac{aa^\top}{(1+e^{a^\top x})(1+e^{-a^\top x})}
$. 
The Hessian information $\nabla^2 h(x,\zeta)$ above will only be used in the delta method.

\begin{figure*}[ht]
        \subfloat[]{
        \includegraphics[width=0.48\textwidth]{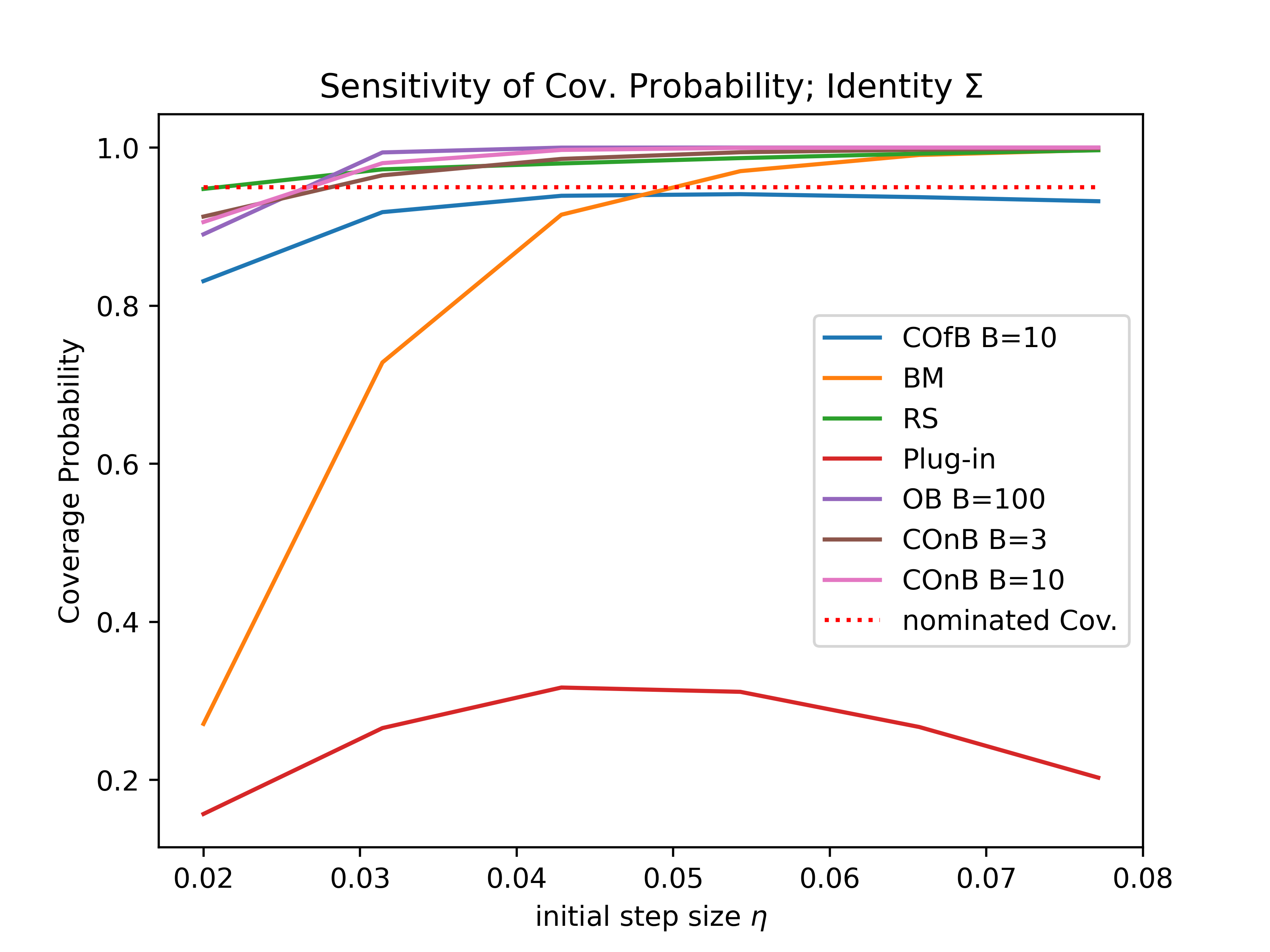}
        }
        \subfloat[]{
        \includegraphics[width=0.48\textwidth]{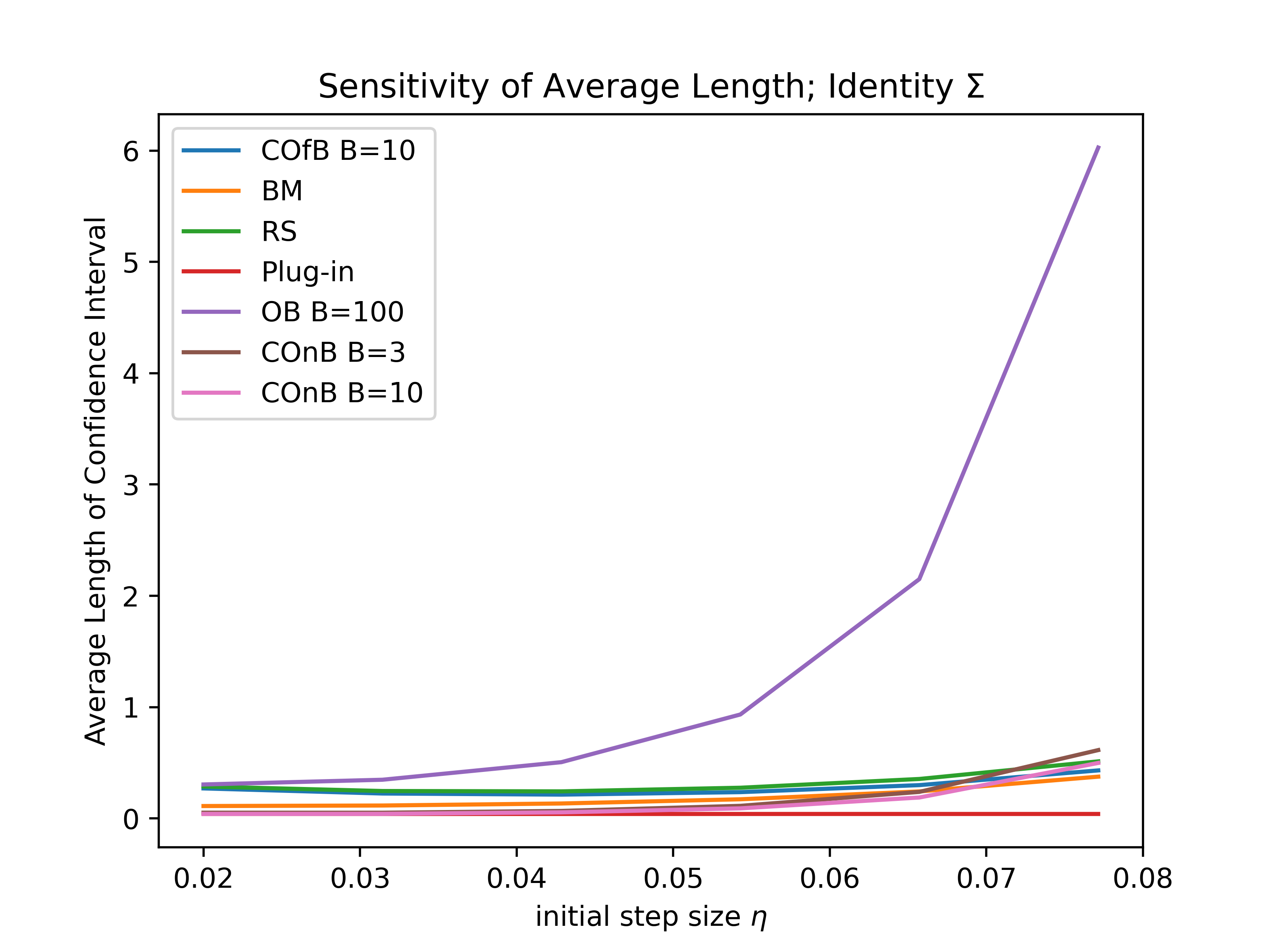}
        }
    \caption{Performance of methods concerning different choices of initial step size $\eta$, with sample size $n=10^4$. We compare the delta method, batch mean method with $M=n^{0.25}$, COfB method with $B=10$, COnB with $B=3, 10$, and the online bootstrap method with $B=100$, when $n=10^4$. The left figure shows the sensitivity of the average coverage probability against $\eta$. The right figure shows the sensitivity of the average length of confidence interval against $\eta$. We report the results for identity $\Sigma$.}
    \label{fig: sensitivity experiments}
\end{figure*}

\subsubsection{Baselines} 
In both experiments, we compare with the batch mean method (BM) and delta method (delta) in \cite{chen2020statistical}, the random scaling method (RS) in \cite{Lee_Liao_Seo_Shin_2022}, the online bootstrap (OB) in \cite{JMLR:v19:17-370}, and the HiGrad method (HiGrad) in \cite{su2018uncertainty}. 

The batch mean method splits $\{x_i\}_{i=0}^n$ into $M$ batches, where $M$ is an extra hyperparameter, with $e_i$ and $s_i$ denoting the ending index and starting index of the $i$-th batch respectively. $n_k$ denotes the number of iterates in $k$-th batch, and the estimator is defined to be $\frac{1}{M}\sum_{k=1}^M n_k(\bar{x}_{n_k} - \bar{x}_M)(\bar{x}_{n_k} - \bar{x}_M)^\top$,
where $\bar{x}_{n_k} = \frac{1}{n_k}\sum_{i=s_k}^{e_k}x_i$ and $\bar{x}_M = \frac{1}{e_{M} - e_{0}}\sum_{i=s_1}^{e_M}x_i$. Let $N = \frac{n^{1-\alpha}}{M+1}$, and $e_k$ to be the closest integer to $((k+1)N)^{\frac{1}{1-\alpha}}$ for each $k=0,\dots,M$ as suggested in \cite{chen2020statistical}. The confidence interval for each entry of $x^*$ is constructed using diagonal entries of the batch mean estimator and a normal quantile.

The delta method \citep{chen2020statistical} generates confidence intervals using normal quantiles and $\tilde \Sigma_n^2 = \tilde G_n^{-1}\tilde S_n\tilde G_n^{-1}$, where $\tilde G_n = \frac{1}{n}\sum_{i=1}^n \nabla^2 h(x_{i-1}, \zeta_i)$, and $\tilde S_n = \frac{1}{n}\sum_{i=1}^n \nabla h(x_{i-1}, \zeta_i)\\(\nabla h(x_{i-1}, \zeta_i))^\top$ are computed on the fly.

The random scaling method \citep{Lee_Liao_Seo_Shin_2022} updates two quantities $A_t\in\mathbb{R}^{d\times d}$ and $b_t\in\mathbb{R}^d$ upon arrival of the $t$-th data with the update rule $A_t = A_{t-1} + t^2 \bar{x}_t\bar{x}_t^\top$ and $b_t = b_{t-1} + t^2 \bar{x}_t$, respectively. Then, construct the random scaling matrix $\hat{V}_n = n^{-2} \left(A_n - \bar{x}_n b_n^\top - b_n \bar{x}_n^\top + \bar{x}_n \bar{x}_n^\top \sum_{s=1}^n s^2 \right)$. The center of confidence interval for $i$-th entry of $x^*$ is $(\bar{x}_n)_i$ and the half-width is $cv_{(1-\alpha/2)} \sqrt{\frac{(\hat{V}_{n})_{i,i}}{n}}$, where $cv_{(1-\alpha/2)}$ denotes the critical value of the studentized statistic. We use $cv_{0.975} = 6.747$ in our experiment, as suggested by \cite{Lee_Liao_Seo_Shin_2022}.

The online bootstrap method \citep{JMLR:v19:17-370} runs $B+1$ ASGD threads in parallel. For each data $\zeta_t$, the update step for $b$-th thread is $x^{(b)}_{t} = x^{(b)}_{t-1} - \eta_t W^{(b)}_t \nabla h(x^{(b)}_{t-1}, \zeta_t),\ b=0,1,\dots,B$, and $W^{(0)}_t = 1,\ \forall t$. Then obtain $\{x^{(b)}\}_{b=0}^B$ by taking average along each thread of SGD. The sample variance for each coordinate $\sigma_{i},\ i=1,\dots,d$ is then calculated for $\{x^{(b)}\}_{b=1}^B$ with a known mean $x^{(0)}$. Then, using normal quantile and $\sigma_i$, one can construct confidence intervals for each entry of $x^*$ centered at $x^{(0)}$.

The HiGrad method \citep{su2018uncertainty} takes two tuples $(B_1, B_2, \dots, B_K)$ and $(n_0, n_1, \\ \dots, n_K)$ as hyperparameters describing when to break the SGD thread into how many branches. $B_i$ describes the number of branches a single branch divides into at $i$-th breaking. $n_i$ describes the number of data each thread uses between $i$ and $i+1$-th breaking. After all the breaking, there will be $T=\prod_{i=1}^K B_i$ threads and one obtains $\{x^{(j)}\}_{j=1}^T$ by averaging each thread. The confidence interval for $i$-th entry of $x^*$ is calculated by aggregating $\{x^{(j)}_i\}_{j=1}^T$. 
% In both experiments, we consider $(B_1, B_2) = (2,2)$ and $(n_0, n_1, n_2) = (n/7,n/7,n/7)$. This set of parameters corresponds to the strategy that first run SGD for $n/7$ steps, then using the current $x_{n/7}$ as initialization, run two SGD for another $n/7$ steps with different data, and then breaks each thread again into two threads and run for another $n/7$ steps. In total, this strategy will give $4$ outcomes. 
It is worth pointing out that the total number of data used in HiGrad should be no more than $n$. i.e. $n_0 + \sum_{i=1}^K B_i n_i \leq n$. Thus, the length of a single thread in HiGrad is typically shorter than that of other methods discussed in this paper. 
% In this example, the length of a single thread is $\frac{3}{7}n$. As mentioned in \cite{su2018uncertainty}, the choice $(B_1, B_2) = (2,2)$ and $(n_0, n_1, n_2) = (n/7,n/7,n/7)$ is desirable as it balances Accuracy, Coverage, and Informativeness. So we compare with this configuration in our numerics and denote it by $\text{HiGrad}_{(2,2)}$ in the table.

\begin{table*}
\centering
\resizebox{\linewidth}{!}
{\begin{tabular}{cccccccc}
     Dimension ($d$)  &  delta & BM & RS & OB ($B=100$) & $\text{HiGrad}_{(2,2)}$ &COfB ASGD ($B=3$) & COnB ($B=3$) \\\hline
     $10$ & $0.32$ & $0.34$ & $0.80$ & $45.90$ & $0.32$ & $0.88$ & $1.75$\\\hline
     $100$ & $0.59$ & $0.56$ & $2.92$ & $46.74$ & $0.49$ & $1.16$ & $1.94$\\\hline
     % $500$ & $2.85$ & $1.79$ & $34.25$ & $59.95$ & $1.46$ & $2.52$ & $3.32$\\\hline
     $1000$ & $8.92$ & $3.49$ & $171.41$ & $76.57$ & $2.78$ & $4.76$ & $5.15$\\\hline
     $2000$ & $44.79$ & $11.35$ & $799.56$ & $116.08$ & $9.72$ & $13.81$ & $13.53$\\\hline
\end{tabular}}
\caption{Average runtimes (seconds) for different methods for linear regression with $n=10^5$, identity $\Sigma$, and dimension $d\in\{10, 100, 1000, 2000\}$.}
\label{table: linear time}
\end{table*}

\subsubsection{Hyperparameters} 

The choices of hyperparameters in both experiments are listed here. The nominal coverage probability we consider is $95\%$. Dimension of the problem $d\in \{5, 20, 200\}$ and we report the result for three choices of covariance matrix $\Sigma$ of $a$. Namely, identity $\Sigma = I_d$, Toeplitz $\Sigma_{i,j} = 0.5^{|i-j|}$ and equicorrelation case $\Sigma_{i,j} = 0.2$ if $i\neq j$ and $\Sigma_{i,i} = 1$. The decay rate for learning rate $\alpha = 0.501$. The optimal solution $x^* = [0,\frac{1}{d-1},\frac{2}{d-1},\dots,1]^\top$ and we set the initial choice $x_0 = [0,0,\dots,0]^\top$. 

For each set of hyperparameters, we run 500 independent trials and report the mean and standard deviation of the coverage probabilities and the average length of the intervals across $d$ dimensions. We tune the initial step size $\eta$ within the range $[0.2, 0.7]$ and report the result with the most accurate average coverage probability. For the batch mean method, $M$ is selected to be the nearest integer to $n^{0.25}$ as suggested in \cite{chen2020statistical}. For HiGrad, the architecture we experimented with is $((2,2), (n/7, n/7, n/7))$. As mentioned in \cite{su2018uncertainty}, this choice is desirable as it balances accuracy, coverage, and informativeness. We report the performance of COfB and COnB with $B = 3, 5, 10$. For the online bootstrap method, $B=200$ is the suggested choice in \cite{JMLR:v19:17-370}. We also consider $B=10$ and $100$ based on the observation that online bootstrap with $B=100$ and $200$ perform similarly which prompts the question of whether smaller $B$'s would work.

\subsubsection{Results}\label{section: experiments discussions}

% Results from linear and logistic regression experiments are presented in \Cref{table: table 1,table: logistic regression table 3}, respectively.  
Result for Toeplitz $\Sigma$ are presented in \Cref{table: table 1 v4}. For full numeric, please refer to \Cref{app: additional numerical results}.
In the table, \textbf{bold} numbers highlight favorable results, where the coverage probability lies between $92\%$ and $98\%$. Conversely, \textit{italic} numbers indicate poor results, specifically those with coverage probabilities below $80\%$.

\paragraph{Coverage Probability}

% In terms of actual coverage probability for $d=5$ or $20$, the delta method, HiGrad method, and our approaches—irrespective of $B$—surpass both the batch mean method and the online bootstrap method with $B=10$. Our methodologies achieve an actual coverage rate of approximately $94-95\%$. In contrast, the batch mean method and online bootstrap with $B=10$ peak at about $92\%$.
Our method gives accurate coverage probability across all values of $d$ and $\Sigma$, typically between $94\%-96\%$, regardless of the choice of $B$. The delta method and the batch mean method consistently fall behind other techniques.
% Generally speaking, across all values of $d$ and $\Sigma$ considered in both experiments, the delta method and batch-mean method consistently fall behind other techniques. 
The random scaling method achieves accurate coverage probability in the linear regression experiment, comparable to our methods. However, it fails in some of the logistic regression experiments when $d=200$.
The HiGrad method displays a stark decline in performance at $d=200$, which is potentially attributed to its abbreviated SGD trajectory. Specifically, when $d=200$, the available data or iterations do not suffice for proper SGD convergence. Similarly, the performance of the delta method drops significantly as $d$ increases. This is largely due to its dependence on a Monte Carlo estimation of a full Hessian matrix, the precision of which diminishes as dimensionality grows. In contrast, the merits of data reuse in bootstrap-type approaches become increasingly evident in higher dimensions. To this end, note that while the online bootstrap method yields comparable coverage probabilities, our methodologies are significantly faster.

\paragraph{Interval Width}

% In terms of the interval width, our approach does result in a longer average confidence interval relative to both the batch-mean method and the delta method. This outcome stems from the $t$-quantile with a degree of freedom defined by either $B-1$ for our COfB running SGD or $B$ for our other methods. However, given that entries of $x^*$ linearly span $[0,1]$ and the confidence intervals are of magnitude $10^{-2}$, the computational advantages seem to more than compensate for the modest increase in interval length.  seem to more than compensate for the modest increase in interval length  when the interval width is within a reasonable range

Our approach results in a longer average confidence interval relative to the delta method, batch mean method, random scaling method, and HiGrad. This outcome stems from the $t$-quantile with a degree of freedom defined by either $B-1$ for our COfB running SGD or $B$ for our other methods. However, given that entries of $x^*$ span $[0,1]$ and the confidence intervals are of magnitude $10^{-2}$, the inflation in interval widths produced by our methods seems secondary compared to our computational gain.
In addition, our average width shrinks rapidly as $B$ increases, which follows the behavior of $t$-interval. With $95\%$ confidence level, the $t$-interval is $49.6\%$ wider than the normal interval when $B=3$, and only $10.9\%$ wider when $B=10$. On the other hand, the execution time grows approximately linearly in $B$.
Furthermore, we note that even though other methods result in narrower intervals, they can fall short significantly in attaining enough coverage probabilities, a performance measure often considered more important than the interval widths. Indeed, we can see from \Cref{table: table 1 v4} that although the delta method, batch mean method, random scaling method, and HiGrad constantly give narrower intervals, their coverage probabilities can lie in the range of $80\%$, $70\%$ to as low as $30\%$ under certain configurations. Nonetheless, despite the above desirable conclusions on our approach, we caution that in some problems the inflation in interval width could be a significant price relative to the magnitude of the solution, in which case one needs to consider a larger $B$ in our approach to rightly balance interval width with computation effort.

\paragraph{Execution Time}

Running times for the linear regression experiments across varied methodologies are available in \Cref{table: linear time}. These experiments were conducted on a single thread of an Apple M2 Pro processor implemented with Python. We report the average execution time of $10$ independent repetitions for each setting.
Both HiGrad and the batch mean method emerged as the swiftest, attributed to their avoidance of additional gradient steps or Hessian calculations. Our COfB and COnB are slightly slower compared with HiGrad and the batch mean method. However, our methods are only marginally slower, especially when dimension $d$ gets larger. In contrast, the online bootstrap method is much slower, with around 100 times longer runtimes than the batch mean method under all settings.
The execution time for the delta method and the random scaling method escalates significantly as dimensions increase. As discussed in the introduction, this is due to the manipulation of $d\times d$ matrices. Specifically, the random scaling method updates matrix $A_i$ at each iteration, while the delta method updates $\tilde{G}_i$ and $\tilde{S}_i $ during gradient steps and requires the $d\times d$ matrix inversion when constructing the confidence interval. Roughly speaking, the random scaling method has time complexity quadratic in $d$, and the delta method is cubic in $d$. The apparent faster performance of the delta method, compared to the random scaling method,  might be attributed to the way they are implemented. 
% For instance, some computationally intensive steps in the delta method are implemented using NumPy, which is known to be very efficient.
% Broadly, computation times align proportionately with the number of gradient steps except for the delta method. Both our methods take roughly three times longer than HiGrad, while the online bootstrap with $B=100$ takes around 100 times as long as HiGrad. One might also notice that COnB has a slightly longer runtime compared to COfB, which is due to the additional computational effort required to generate $\{W_{b,t}\}$.

\paragraph{Sensitivity Analysis}
% (WOULD IT BE MORE NATURAL TO ABSORB THIS INTO THE RESULT SUBSECTION ABOVE? OR YOU CAN HAVE "COVERAGE", "INTERVAL WIDTH" AND "SENSITIVITY" AS THREE SUBSECTIONS OF "RESULTS")
In \Cref{fig: sensitivity experiments}, we compare the performance of our methods, the delta method, the batch mean method, and the online bootstrap method with the number of samples $n=10^4$ for the linear regression problem. 
Observe that the coverage probability of COfB methods remains stable at around $95\%$ regardless of changes in the initial step size. The random scaling method also performs well in this experiment, although it has a slight tendency of over-coverage. On the other hand, the batch mean method requires a careful choice of the initial step size to give a comparable coverage rate, and the optimal choice is not the same across different problems. The delta method suffers from a huge under-coverage and fails to give a valid confidence interval. The delta method and the batch mean method have smaller average lengths, which can be associated with their under-coverages. It can be observed that the coverage probability and the average length of the batch mean estimator both increase as $\eta$ increases. Our COnB method has a similar sensitivity as the online bootstrap method. Nonetheless, as mentioned earlier, COnB is substantially faster. Additionally, the average length of our COnB becomes almost the same as that of the online bootstrap method when increasing $B$ to $10$.

\subsection{Sparse Linear Regression with Increasing Dimensionality}

\subsubsection{Problem Setting and Hyperparameters} 
We evaluate the performance of \Cref{alg: two-stage} under the setting discussed in \Cref{sec: sparse linear regression}. 
Specifically, we consider $n=100$ and $d=500$. 
The columns of the $n\times d$ design matrix are $i.i.d.$, generated from a multivariate normal distribution $\mathcal{N}(0,\Sigma)$, with choices of $\Sigma$ the same as in the fixed dimensionality experiment. 
We set the sparsity parameter $p$ to be $3$ or $15$, with the first $p$ entries of the true coefficients non-zero and each uniformly drawn over the interval $[0,2]$. Consequently, the true model coefficient $x^{(100)} = [x^*_1, \dots, x^*_p, 0,\dots,0]\in\mathbb{R}^{500}$. 
The nominal coverage probability for the entries in $T^*$ is still $95\%$. 

For the model selection stage, we set the penalty parameter $\lambda$ to $0.001 \times \frac{\log d}{n}$.
We tested both COfB and COnB as the second stage method. Choices of the number of resampling runs are $B \in \{3, 10\}$. In all the experiments, the decay rate of learning rate is fixed at $\alpha=0.501$, and the initial guess for SGD is $x_0 = [0, \dots, 0]^\top \in \mathbb{R}^{500}$. The initial learning rate $\eta$, ranging within $[0.02, 0.5]$, is tuned to ensure proper convergence of SGD.

\subsubsection{Results}

We conducted 500 independent trials for each hyperparameter configuration, reporting the mean and standard deviation of the coverage probability and average interval length for coefficients within and outside $T^*$, respectively. 
The results for Toeplitz $\Sigma$ can be found in \Cref{table: Sparse Linear Regression v4}. Please refer to \Cref{app: additional numerical results} for the full table. In the table, \textbf{bold} numbers represent good results, with coverage probability lying between $92\%$ and $98\%$ on $T^*$ or over $99\%$ outside $T^*$.
As indicated by \Cref{thm: two-stage guarantee}, the ideal coverage probability for coefficients in $T^*$ would be close to $95\%$. For coefficients not in $T^*$ (rows with $\notin T^*$ in the table), a coverage close to 1 is desirable. 

We conclude from the table that our methods, regardless of the configuration, achieve accurate coverage probability both within and outside $T^*$. 
The average length of confidence intervals is generally reasonable compared with the scale of the problem. 
Notably, confidence intervals for coefficients outside $T^*$ are significantly shorter than those inside $T^*$. This is due to the accurate model selection in the first stage, as our method will simply output singleton $\{0\}$ for entries not selected in the first stage. 
We also observe that the average length shrinks when $B$ increases while maintaining high accuracy in coverage probabilities across all $B$ values.

\begin{table*}
\centering
\resizebox{\linewidth}{!}{
\begin{tabular}{cc c c c cc} 
 \hline
 & \multicolumn{2}{c}{$d=5$} & \multicolumn{2}{c}{$d=20$} & \multicolumn{2}{c}{$d=200$}\\ 
 & Cov (\%) & Len ($\times 10^{-2}$)& Cov (\%) & Len ($\times 10^{-2}$) & Cov (\%) & Len ($\times 10^{-2}$) \\ \hline
\multicolumn{7}{c}{Linear Regression} \\ \hline
delta & $\textbf{94.20}$ ($0.07$) & $1.53$ ($0.00$)& $\textbf{93.23}$ ($0.08$) & $1.58$ ($0.00$)& $\textit{37.16}$ ($0.15$) & $1.60$ ($0.00$)\\ \hline
BM & $91.80$ ($0.09$) & $1.53$ ($0.00$)& $88.51$ ($0.10$) & $1.51$ ($0.00$)& $\textbf{97.37}$ ($0.05$) & $6.54$ ($0.01$)\\ \hline
RS & $\textbf{92.83}$ ($0.08$) & $1.94$ ($0.01$)& $\textbf{93.38}$ ($0.08$) & $2.15$ ($0.01$)& $\textbf{97.02}$ ($0.05$) & $8.60$ ($0.06$)\\ \hline
OB $B=10$ & $\textbf{92.48}$ ($0.08$) & $1.66$ ($0.00$)& $\textbf{93.68}$ ($0.08$) & $1.81$ ($0.00$) & $\textbf{93.33}$ ($0.25$) & $13.73$ ($0.01$)\\ \hline
OB $B=100$ & $\textbf{95.48}$ ($0.07$) & $1.72$ ($0.00$)& $\textbf{95.84}$ ($0.06$) & $1.97$ ($0.00$) & $\textbf{96.58}$ ($0.18$) & $14.26$ ($0.01$) \\ \hline
$\text{HiGrad}_{(2,2)}$ & $\textbf{94.33}$ ($0.07$) & $2.69$ ($0.01$)& $\textbf{94.88}$ ($0.07$) & $2.82$ ($0.01$)& $\textbf{94.09}$ ($0.07$) & $2.83$ ($0.01$)\\ \hline
COfB ASGD $B=3$  & $\textbf{94.72}$ ($0.07$) & $3.06$ ($0.00$)& $\textbf{94.95}$ ($0.07$) & $3.30$ ($0.00$)& $\textbf{94.41}$ ($0.07$) & $14.44$ ($0.02$)\\ \hline
COfB ASGD $B=5$  & $\textbf{95.24}$ ($0.07$) & $2.21$ ($0.00$)& $\textbf{94.89}$ ($0.07$) & $2.26$ ($0.00$)& $\textbf{93.91}$ ($0.08$) & $9.96$ ($0.01$)\\ \hline
COfB ASGD $B=10$ & ${\textbf{95.28}}$ ($0.07$) & ${1.74}$ ($0.00$)& ${\textbf{94.95}}$ ($0.07$) & ${2.12}$ ($0.00$)& $\textbf{94.01}$ ($0.08$) & $8.44$ ($0.01$)\\ \hline
COfB SGD $B=3$  & $\textbf{95.16}$ ($0.07$) & $6.93$ ($0.01$)& $\textbf{94.76}$ ($0.07$) & $4.34$ ($0.00$)& $\textbf{95.16}$ ($0.07$) & $6.90$ ($0.01$)\\ \hline
COfB SGD $B=5$  & $\textbf{95.84}$ ($0.06$) & $4.70$ ($0.00$)& $\textbf{94.20}$ ($0.07$) & $2.97$ ($0.00$)& $\textbf{95.04}$ ($0.07$) & $4.72$ ($0.00$)\\ \hline
COfB SGD $B=10$ & $\textbf{95.36}$ ($0.07$) & $3.95$ ($0.00$)& $\textbf{94.20}$ ($0.07$) & $2.50$ ($0.00$)& ${\textbf{95.00}}$ ($0.07$) & ${3.98}$ ($0.00$)\\ \hline
COnB $B=3$ & $\textbf{95.00}$ ($0.07$) & $2.49$ ($0.00$)& $\textbf{95.36}$ ($0.07$) & $2.94$ ($0.00$)& $\textbf{95.41}$ ($0.07$) & $21.04$ ($0.02$) \\ \hline 
COnB $B=5$ & $\textbf{94.60}$ ($0.07$) & $1.96$ ($0.00$)& $\textbf{95.43}$ ($0.07$) & $2.44$ ($0.00$)& $\textbf{95.48}$ ($0.07$) & $17.42$ ($0.02$) \\ \hline 
COnB $B=10$ & $\textbf{94.92}$ ($0.07$) & $1.77$ ($0.00$)& $\textbf{95.42}$ ($0.07$) & $2.18$ ($0.00$)& $\textbf{95.78}$ ($0.06$) & $15.61$ ($0.01$) \\ \hline
\multicolumn{7}{c}{Logistic Regression} \\
\hline
delta & $\textbf{94.83}$ ($0.07$) & $4.05$ ($0.00$)& $\textbf{93.29}$ ($0.08$) & $5.59$ ($0.00$)& $\textit{53.69}$ ($0.16$) & $9.56$ ($0.00$)\\ \hline
BM & $84.00$ ($0.12$) & $3.16$ ($0.01$)& $\textit{75.25}$ ($0.14$) & $3.75$ ($0.01$)& $\textit{34.93}$ ($0.15$) & $7.30$ ($0.03$)\\ \hline
RS & $\textbf{92.67}$ ($0.08$) & $5.14$ ($0.02$)& $90.88$ ($0.09$) & $7.26$ ($0.03$)& $76.38$ ($0.13$) & $17.45$ ($0.10$)\\ \hline
OB ($B=100$) & $\textbf{95.00}$ ($0.07$) & $4.22$ ($0.00$)& $\textbf{94.04}$ ($0.07$) & $6.65$ ($0.01$)& $99.78$ ($0.01$) & $69.55$ ($0.26$) \\ \hline
OB ($B=200$) & $\textbf{95.00}$ ($0.07$) & $4.24$ ($0.00$)& $\textbf{94.67}$ ($0.07$) & $6.70$ ($0.01$)& $99.78$ ($0.01$) & $69.28$ ($0.26$) \\ \hline
$\text{HiGrad}_{(2,2)}$ & $\textbf{95.33}$ ($0.07$) & $7.18$ ($0.03$)& $\textbf{93.38}$ ($0.08$) & $8.92$ ($0.03$)& $\textit{57.02}$ ($0.16$) & $10.27$ ($0.04$)\\ \hline
COfB ASGD $B=3$  & $\textbf{94.12}$ ($0.07$) & $5.70$ ($0.00$)& $\textbf{94.77}$ ($0.07$) & $11.49$ ($0.01$)& $\textbf{93.79}$ ($0.08$) & $42.27$ ($0.05$)\\ \hline
COfB ASGD $B=5$  & $\textbf{94.32}$ ($0.07$) & $4.82$ ($0.00$)& $\textbf{94.81}$ ($0.07$) & $7.97$ ($0.01$)& $\textbf{93.23}$ ($0.08$) & $28.81$ ($0.02$)\\ \hline
COfB ASGD $B=10$ & $\textbf{94.88}$ ($0.07$) & $4.61$ ($0.00$)& $\textbf{94.65}$ ($0.07$) & $6.69$ ($0.00$)& $\textbf{93.09}$ ($0.08$) & $24.43$ ($0.01$)\\ \hline
COfB SGD $B=3$  & $\textbf{95.36}$ ($0.07$) & $9.48$ ($0.01$)& $\textbf{94.80}$ ($0.07$) & $9.65$ ($0.01$)& $\textbf{94.41}$ ($0.07$) & $30.53$ ($0.03$)\\ \hline
COfB SGD $B=5$  & $\textbf{95.40}$ ($0.07$) & $7.98$ ($0.00$)& $\textbf{94.37}$ ($0.07$) & $8.16$ ($0.00$)& $\textbf{93.79}$ ($0.08$) & $20.76$ ($0.02$)\\ \hline
COfB SGD $B=10$ & $\textbf{95.40}$ ($0.07$) & $7.98$ ($0.00$)& $\textbf{94.37}$ ($0.07$) & $8.16$ ($0.00$)& $\textbf{93.62}$ ($0.08$) & $17.56$ ($0.01$)\\ \hline
COnB ($B=3$) & $\textbf{94.00}$ ($0.08$) & $6.22$ ($0.02$)& $\textbf{94.71}$ ($0.07$) & $10.27$ ($0.05$)& $\textbf{97.82}$ ($0.05$) & $99.61$ ($0.60$) \\ \hline
COnB ($B=5$) & $\textbf{95.00}$ ($0.07$) & $5.25$ ($0.02$)& $\textbf{94.71}$ ($0.07$) & $8.20$ ($0.03$)& $98.75$ ($0.04$) & $85.15$ ($0.45$) \\ \hline
COnB ($B=10$) & $\textbf{94.83}$ ($0.07$) & $4.84$ ($0.01$)& $\textbf{94.29}$ ($0.07$) & $7.25$ ($0.02$)& $99.31$ ($0.03$) & $78.02$ ($0.37$) \\ \hline
    \end{tabular}}
 \caption{Results for the linear and logistic regression with Toeplitz $\Sigma$, $n=10^5$.}
\label{table: table 1 v4}
\end{table*}

\begin{table*}
\centering
\resizebox{.9\linewidth}{!}
{\begin{tabular}{cc c c c c} 
\hline
& &\multicolumn{2}{c}{$p=3$}  & \multicolumn{2}{c}{$p=15$}\\ 
& & Cov (\%) & Len ($\times 10^{-2}$)& Cov (\%) & Len($\times 10^{-2}$) \\
\hline
\multirow{2}{*}{COfB ASGD $B=3$}
& $\in T^*$ & $\textbf{95.20}$ ($21.38$) &  $2.75 $ ($0.01$) & $\textbf{95.12}$ ($21.54$) & $34.87 $ ($0.11$)\\
& $\notin T^*$ & $\textbf{99.88}$ ($3.42$)& $0.05$ ($0.00$) &  $\textbf{99.09}$ ($9.52$)& $8.62$ ($0.07$) \\
\hline
\multirow{2}{*}{COfB ASGD $B=10$}
& $\in T^*$ & $\textbf{95.07}$ ($21.66$) &  $0.86 $ ($0.00$) & $\textbf{93.13}$ ($25.29$) & $4.33 $ ($0.01$)\\
& $\notin T^*$ & $\textbf{99.86}$ ($3.74$)& $0.02$ ($0.00$) &  $\textbf{99.12}$ ($9.32$)& $1.00$ ($0.01$) \\
\hline
\multirow{2}{*}{COnB ASGD $B=3$}
& $\in T^*$ & $\textbf{94.73}$ ($22.34$) &  $1.28 $ ($0.01$) & $\textbf{97.19}$ ($16.54$) & $10.68 $ ($0.03$)\\
& $\notin T^*$ & $\textbf{99.93}$ ($2.73$)& $0.03$ ($0.00$) &  $\textbf{99.80}$ ($4.47$)& $2.47$ ($0.02$) \\
\hline
\multirow{2}{*}{COnB ASGD $B=10$}
& $\in T^*$ & $\textbf{95.40}$ ($20.95$) &  $1.07 $ ($0.00$) & $98.85$ ($10.65$) & $10.27 $ ($0.04$)\\
& $\notin T^*$ & $\textbf{99.96}$ ($2.00$)& $0.02$ ($0.00$) &  $\textbf{99.98}$ ($1.51$)& $2.41$ ($0.02$) \\
\hline
\end{tabular}}
 \caption{Results for sparse linear regression with Toeplitz $\Sigma$, $n=100$. }
\label{table: Sparse Linear Regression v4}
\end{table*}

 Finally, we provide further sensitivity analyses of our approach with respect to ill-conditionedness and learning rate scheme in \Cref{app: additional discussion on assumptions}.

\section*{Acknowledgements}
% We gratefully acknowledge support from the InnoHK initiative, the Government of the HKSAR, and Laboratory for AI-Powered Financial Technologies. 
We gratefully acknowledge support from the InnoHK initiative, the Government of the HKSAR, Laboratory for AI-Powered Financial Technologies, and the Columbia Innovation Hub Award. We thank Sokbae (Simon) Lee for the helpful suggestions that have greatly improved our manuscript.

% with respect to our assumptions in \Cref{app: additional discussion on assumptions}.

% on We discuss and empirically evaluate the sensitivities 
% We caution that in practice, some of the above assumptions may not be strictly satisfied. For instance, the underlying optimization problem may be ill-conditioned, in which case the regularity conditions imposed in \Cref{assumption: strongly convex objective with bounded hessian} may not hold. In addition, implementations may adopt learning rate schemes different from the form required in Assumptions \ref{assumption: ASGD step size} and \ref{assumption: SGD step size}. We discuss and empirically evaluate the sensitivities of our approach with respect to our assumptions in \Cref{app: additional discussion on assumptions}.}
% % for discussions of such settings.} 

\appendix
\crefalias{section}{appendix}
\crefalias{subsection}{appendix}
\crefname{appendix}{Appendix}{Appendices}
\Crefname{appendix}{Appendix}{Appendices}

\section{Useful Lemmas and Missing Proofs}
\subsection{Convergence of Normal Variables with Empirical Variances}\label{subsection: convergence normal variables}
\begin{lemma}
    \label{lem: clt mn separate 1}
    For a fixed $n$ and given $\hat{P}_n$, let $\hat{Z}_n$ be the weak limit of $\sqrt{m}(\psi_m(\hat{P}_n) - \psi(\hat{P}_n))$ as $m\to\infty$. Let $Z_0$ be the weak limit of $\sqrt{m}(\psi_m(P) - \psi(P))$.
    Under the same assumptions as \Cref{thm: main theorem}, for any Borel set $D$, 
    \begin{equation*}
        |\mathbb{P}^*(\hat{Z}_{n}\in D) - \mathbb{P}(Z_0\in D)| \rightarrow 0 \quad \text{in probability},
    \end{equation*}
    as $n\to\infty$.
\end{lemma}
\begin{proof}
Consider the ASGD case now, and the SGD case can be treated similarly. 
By the classical convergence result for ASGD (see \cite{doi:10.1137/0330046}), we have the following:
\begin{align*}
    \hat{Z}_n\sim N(0, \sigma_n^2), &\quad \sigma_n^2 = G_n(\hat{x}_n)^{-1}S_n(\hat{x}_n)G_n(\hat{x}_n)^{-1},\\
    Z_0\sim N(0, \sigma^2), &\quad \sigma^2 = G(x^*)^{-1} S(x^*) G(x^*)^{-1}.
\end{align*}
As a result, $|\mathbb{P}^*(\hat{Z}_{n}\in D) - \mathbb{P}(Z_0\in D)|$ is a continuous function of entries of $G_n(\hat{x}_n)$ and $S_n(\hat{x}_n)$. Therefore, it suffices to prove that $(G_n(\hat{x}_n))_{i,j} \rightarrow (G(x^*))_{i,j}$ and $(S_n(\hat{x}_n))_{i,j} \rightarrow (S(x^*))_{i,j}$ in probability, for each $i,j \in \{1,\dots,d\}$. 

To simplify the notation, let us define $Q f(x) \triangleq \int f(x,\zeta) Q(d\zeta)$ for a function $f$ of $\zeta$ parameterized by $x$ and a probability measure $Q$. Then, $(G_n(\hat{x}_n))_{i,j} = \hat{P}_n \partial^2_{i,j} h(\hat{x}_n)$ and $(G(x^*))_{i,j} = P\partial^2_{i,j} h(x^*)$. 
Consider the following decomposition:
\begin{align}
\label{eqt: G_n decomposition}
\begin{split}
&(G_n(\hat{x}_n))_{i,j} - (G(x^*))_{i,j}\\
= & \left(\hat{P}_n \partial^2_{i,j} h(\hat{x}_n) - P \partial^2_{i,j} h(\hat{x}_n)\right) 
+ \left(P\partial^2_{i,j}h(\hat{x}_n) - P\partial^2_{i,j}h(x^*)\right).
\end{split}
\end{align}
By \Cref{assumption: strongly convex objective with bounded hessian}, $P\partial^2_{i,j}h(x)$ is continuous in $x$. Combined with the fact that $\hat{x}_n$ converges to $x^*$ in probability, the second term in \eqref{eqt: G_n decomposition} converges to 0 in probability.

By \Cref{assumption: ASGD step size},  $\mathcal{F}_{i,j}$ is $P$-Glivenko-Cantelli, and the first term in \eqref{eqt: G_n decomposition} vanishes
\begin{equation}\label{eqt: p glivenko cantelli 1st}
 \left|\hat{P}_n \partial^2_{i,j} h(\hat{x}_n) - P \partial^2_{i,j} h(\hat{x}_n)\right|  \leq \sup_{f\in\mathcal{F}_{i,j}} |\hat{P}_n f - P f| \xrightarrow{as*} 0.
\end{equation}
So we conclude that $(G_n(\hat{x}_n))_{i,j} - (G(x^*))_{i,j}$ converges to $0$ in probability.

By definition, $(S_n(\hat{x}_n))_{i,j} = \hat{P}_n \partial_i h(\hat{x}_n)\partial_j h(\hat{x}_n)$ and $(S(x^*))_{i,j} = P \partial_i h(x^*) \partial_j h(x^*)$. Define $\tilde{\mathcal{F}}_{i,j} = \{\partial_i h(x,\zeta)\partial_j h(x,\zeta): x\in\mathcal{X}\}$.
By \Cref{assumption: strongly convex objective with bounded hessian}, there is a constant $m$ depending on $L$ such that $|\partial_i h(x_1,\zeta)\partial_j h(x_1,\zeta) - \partial_i h(x_2,\zeta)\partial_j h(x_2,\zeta)| < m \|x_1 - x_2\|$, for all $x_1$, $x_2$, and $\zeta$. Therefore, $\tilde{\mathcal{F}}_{i,j}$ is $P$-Glivenko-Cantelli.
The remaining proof follows the same arguments as in the proof of $(G_n(\hat{x}_n))_{i,j} - (G(x^*))_{i,j}\xrightarrow{P}0$. Hence, we conclude that $|\mathbb{P}^*(\hat{Z}_{n}\in D) - \mathbb{P}(Z_0\in D)| \rightarrow 0$ in probability.
\end{proof}

\subsection{ASGD and SAA Have Asymptotically Negligible Discrepancy}\label{sup: ASGD equiv SAA}

\begin{theorem}\label{thm: SA and SAA equivalent}
Under the same assumptions as \Cref{thm: main theorem}, the outputs of ASGD and SAA have an asymptotically negligible discrepancy in the sense that
\[
\sqrt{n} (\bar{x}_n - \hat{x}_n) \rightarrow 0 \quad \text{in probability.}
\]
\end{theorem}

\begin{proof}
Let $\bar{\Delta}_n = \bar{x}_n - x^*$, the residual of ASGD. Define $\bar{\Delta}^1_n$ by 
\begin{align*}
& {\Delta}^1_0  = x_0 - x^*,\\
& {\Delta}^1_t = {\Delta}^1_{t-1} - \eta_t G(x^*) {\Delta}^1_{t-1} + \eta_t \xi_t,\ \xi_t = \nabla h(x_{t-1}, \zeta_t) - \nabla H(x_{t-1}),\ \forall t=1,\dots,n,\\
&\bar{\Delta}^1_n = \frac{1}{n} \sum_{t=1}^{n}{\Delta}^1_t.
\end{align*}
We use superscript $^1$ to indicate that $\bar{\Delta}^1_n$ is an approximated version of $\bar\Delta_n$. Moreover, $\bar{\Delta}^1_n$ is also the residual of the ASGD solution for minimizing $\frac{1}{2} G(x^*) (x - x^*)^2$ with gradient noise sequence $\{\xi_t\}_t$.
Consider the following decomposition:
\begin{align*}
    & \sqrt{n} (\bar{x}_n - \hat{x}_n)\\
    =& \sqrt{n} (\bar{\Delta}_n - \bar{\Delta}^1_n) + \sqrt{n}(\bar{\Delta}^1_n - \frac{1}{n} \sum_{t=1}^{n}G(x^*)^{-1}\xi_t) - \sqrt{n} (\hat{x}_n - x^* - \frac{1}{n} \sum_{t=1}^{n}G(x^*)^{-1}\xi_t).
\end{align*}
It suffices to show that all three terms in the above decomposition vanish.
In the decomposition, the first term describes the closeness of $\bar\Delta_n$ and its approximation with the same gradient noises but a surrogate objective. By the proof of Theorem 2, part 4 in \cite{doi:10.1137/0330046}, $
\sqrt{n} (\bar{\Delta}_n - \bar{\Delta}^1_n) \xrightarrow{P} 0
$.
The second term follows from the classical stochastic approximation result for linear problem.
In particular, by the proof of Theorem 1, part 1 in \cite{doi:10.1137/0330046}, 
$\sqrt{n} \left(\bar{\Delta}^1_n - \frac{1}{n} \sum_{t=1}^{n}G(x^*)^{-1}\xi_t\right) \xrightarrow{P} 0$.
The last term converges to $0$ in probability by standard M-estimator theory; see, for example, Theorem 5.21 in \cite{van2000asymptotic}.
\end{proof}

\subsection{Discussion on Assumptions}
\label{app: assumptions discussion}
The following set of assumptions is listed in \cite{shao2022berry} to guarantee a Berry-Esseen type bound for the residual of ASGD.

\begin{assumption}
\label{assumption: shao 1}
    There exists a constant $\tau_0 > 0$ such that $\|x_0 - x^*\|_4 \leq \tau_0$.
\end{assumption}

\begin{assumption}
\label{assumption: shao 2}
The sequence $\{\nabla h(x_{t-1}, \zeta_{t}) - \nabla H(x_{t-1})\}_t$ is independent of $x_0$, and for each $t\geq 1$, $\nabla h(x_{t-1}, \zeta_{t}) - \nabla H(x_{t-1})$ admits the following decomposition:
\[
\nabla h(x_{t-1}, \zeta_{t}) - \nabla H(x_{t-1}) = \xi_t + \gamma_t,
\]
where:
\begin{enumerate}
    \item $\{\xi_t\}$ is a sequence of independent random variables and $\mathbb{E}[\xi_i] = 0$ and $\mathbb{E}[\xi_i \xi_i^\top ] = \Sigma_i$; there exist positive numbers $\lambda_1$ and $\lambda_2$ such that for any $i\geq 1$, $\lambda_1 \leq \lambda_{min}(\Sigma_i) \leq \lambda_{max}(\Sigma_i)\leq \lambda_2$; moreover, there exists a positive number $\tau $ such that $\max_i \|\xi_i\|_4\leq \tau$.
    \item Let $\mathcal{F}_0 = \sigma(x_0)$, and for each $t\geq 0$, $\mathcal{F}_t=\sigma(x_0, \zeta_k| k\leq t)$; let $g(\cdot, \cdot):\mathbb{R}^d\times \mathbb{R}^d\rightarrow\mathbb{R}^d$, and let the random variable $\gamma_t = g(x_{t-1}, \zeta_t)$ satisfy $\mathbb{E}[\gamma_t | \mathcal{F}_{t-1}] = 0$. For any $x$ and $x'$, there exists a non-negative constant $c_1 \geq 0$ such that 
    \[
    g(x, \zeta) - g(x', \zeta) \leq c_1 \|x-x'\| \quad \text{and}\quad g(x^*,\zeta)=0,
    \]
    for all $\zeta\in\mathbb{R}^d$.
\end{enumerate}
\end{assumption}

\begin{assumption}
    \label{assumption: shao 3}
    The function $H$ is $L$-smooth and strongly convex with convexity constant $\mu>0$.
\end{assumption}

\begin{assumption}
    \label{assumption: shao 4}
    There exist positive constants $c_2$ and $\beta$ such that for all $x$ with $\|x - x^*\|\leq \beta$:
    \[
    \|G(x) - G(x^*)\| \leq c_2 \|x - x^*\|.
    \]
\end{assumption}

\begin{lemma}
    \label{lem:shaotheorem3-4}
    Consider the ASGD procedure \eqref{SA iteration}. Under \Cref{assumption: shao 1,assumption: shao 2,assumption: shao 3,assumption: shao 4}, for any Borel set $D$, we have:
    \begin{enumerate}
        \item if $\alpha \in (\frac{1}{2}, 1)$, 
            \begin{equation*}
                |\mathbb{P}(\sqrt{n} \Sigma_n^{-\frac{1}{2}} (\bar{x}_n - x^*) \in D ) - \mathbb{P}(Z^{ASGD} \in D)| \leq C (d^{3/2} + \tau^3 + \tau_0^3) (d^{1/2} n^{-1/2} + n^{-\alpha + 1/2});
            \end{equation*}
        \item if $\alpha = 1$, for any $\epsilon > 0$,
            \begin{align*}
                |\mathbb{P}(\sqrt{n} \Sigma_n^{-\frac{1}{2}} (\bar{x}_n - x^*) \in D ) - \mathbb{P}(Z^{ASGD} \in D)|
                \leq  C (d^{3/2} + \tau^3 + \tau_0^3)n^{-1/2+\epsilon}d^{1/2}.
            \end{align*}
    \end{enumerate}
\end{lemma}
Our \Cref{assumption: strongly convex objective with bounded hessian,assumption: unbiased gradient estimation,assumption: gradient variability} imply the above set of assumptions. Specifically, \Cref{assumption: strongly convex objective with bounded hessian} implies \Cref{assumption: shao 4}. \Cref{assumption: shao 1,assumption: shao 3} are implied by \Cref{assumption: gradient variability,assumption: strongly convex objective with bounded hessian}. 
To see that our assumptions imply \Cref{assumption: shao 2},
let $g(x,\zeta) = \nabla h(x, \zeta) - \nabla h(x^*, \zeta) - \nabla H(x)$, and $\xi_t = \nabla h(x^*, \zeta_t)$. 
Given \Cref{assumption: gradient variability}, and with $\xi_t$ and $\gamma_t = g(x_{t-1}, \zeta_t)$, it follows that $\mathbb{E}[\xi_t \xi_t^\top] = \mathbb{E}[\nabla h(x^*, \zeta_t) \nabla h(x^*, \zeta_t)^\top] \equiv S(x^*)$.
Consequently, the first condition of \Cref{assumption: shao 2} holds. The martingale-difference structure of $g$, which is part of the second condition of \Cref{assumption: shao 2}, stems from \Cref{assumption: unbiased gradient estimation}. One can also verify that $g(x^*, \zeta) = 0$ by direct calculation. For the Lipschitz property of $g$, notice that
\[
\|g(x,\zeta) - g(x',\zeta)\| \leq \|\nabla h(x,\zeta) - \nabla h(x', \zeta)\| + \|\nabla H(x) - \nabla H(x')\|,
\]
where the right-hand-side is bounded by a constant multiple of $\|x-x'\|$ since the Hessian of both $h$ and $H$ have bounded spectrums, as stated in \Cref{assumption: strongly convex objective with bounded hessian}.

To conclude, the two bounds in \Cref{lem:shaotheorem3-4} hold under our assumptions. 

% \subsection{A Useful Berry-Esseen-Type Bound}
% \begin{lemma}\label{lem: shao lemma 4}
% \citep{shao2022berry} Let $T$ be a d-dimensional statistic. Suppose that $T = T(\zeta_1, \dots, \zeta_n)$ admits the decomposition $T=W + D$, where $W = \sum_{i=1}^n f_i(\zeta_i)$ satisfies:
% \[
% \mathbb{E}[f_i(\zeta_i)] = 0, \forall i,\quad \text{and } \sum_{i=1}^n \mathbb{E}[f_i(\zeta_i)(f_i(\zeta_i))^\top] = \Sigma_n,
% \]
% where $0<\sigma_n = \lambda_{min} (\Sigma_n)$, and $f$ is some mapping to $\mathbb{R}^d$. Define $\gamma_n = \sum_{i=1}^n \mathbb{E}\|f_i(\zeta_i)\|^3$. Let $\Delta$ and $(\Delta^{(i)}){1\leq i \leq n}$ be random variables such that $\Delta \geq \|D\|$ and $\Delta^{(i)}$ is independent of $\zeta_i$. Let $Z_0\sim \mathcal{N}(0, I_d)$.
% Then, for any Borel set $D$,
% \begin{align*}
% & | \mathbb{P}(\Sigma_n^{-1/2} T\in D) - \mathbb{P}(Z_0\in D) | \\
% \leq & \ 259 \sigma_n^{-3/2} d^{1/2} \gamma_n + 2\sigma_n^{-1} \mathbb{E}[\|W\|\Delta]  + 2\sigma_n^{-1} \sum_{i=1}^n \mathbb{E} [\|f_i(\zeta_i)\| |\Delta - \Delta^{(i)}|].
% \end{align*}
% \end{lemma}

\subsection{Proof of Theorem \ref{thm:nonasymptotic bound} in the ASGD Case}\label{sup: ASGD}

\begin{proof}
Let $D$ be a Borel set. Consider $\alpha \in (\frac{1}{2}, 1)$ first.
As discussed in the main body. It is sufficient to establish the relationship \eqref{eqt: ASGD COfB inner bound uniform}. Specifically, we will show that there exist
$(\hat\tau_0, \hat\tau, \hat C)$ such that for any $\delta > 0$, there is an integer $N$ (depending on $\delta$), satisfying
\begin{align}
    \label{pf: thm1 claim2}
    \begin{split}
        &\sup_{n>N}|\mathbb{P}^*(\sqrt{n}(\psi_n(\hat{P}_n) - \psi(\hat{P}_n))\in D) - \mathbb{P}^*(\hat{Z}_n\in D)|\\
    & \quad \leq \hat C(d^{3/2} + \hat \tau^3 + \hat \tau_0^3) (d^{1/2}n^{-1/2} + n^{-\alpha+1/2}),
    \end{split}
\end{align}
with probability at least $1-\delta$.

By \Cref{lem:shaotheorem3-4}, we have the following Berry-Esseen bound for fixed data distribution $P$:
\begin{align}\label{eqt: A3 shao}
\begin{split}
    &|\mathbb{P}(\sqrt{n}(\psi_n(P) - \psi(P))\in D) - \mathbb{P}(Z_0\in D)|\\
    \leq& C(d^{3/2} + \tau^3 + \tau_0^3) (d^{1/2}n^{-1/2}+ n^{-\alpha+1/2}),
\end{split}
\end{align}
where $C = C(\eta, \lambda_1, \lambda_2, \alpha, l, L)>0$ is a constant independent of $D$, $d$, $\tau$, and $\tau_0$. It remains to specify $\hat C$, $\hat \tau$, and $\hat \tau_0$ in \eqref{pf: thm1 claim2}.

Consider $\hat\tau_0$ first. Let $\epsilon_0,\delta_0 > 0$, and set $\hat{\tau}_0 = (1+\epsilon_0)\tau_0$. By \Cref{assumption: strongly convex objective with bounded hessian,assumption: m estimator consistency} and M-estimator theory (see \cite{van2000asymptotic} for example), we have $\psi(\hat{P}_n)\xrightarrow{P}\psi(P)$.
As a result, there is $N_0=N_0(\epsilon_0, \delta_0)$ such that for any $n>N_0$, we have $\|\psi(P) - \psi(\hat{P}_n)\| < \epsilon_0 \tau_0$ with probability at least $1-\delta_0$. Consider the decomposition $\|x_0 - \psi(\hat{P}_n)\| \leq \|x_0 - \psi(P)\| + \|\psi(P) - \psi(\hat{P}_n)\|$. 
Therefore, for any $n>N_0$, $\mathbb{P}(\|x_0 - \psi(\hat{P}_n)\| \leq \hat\tau_0) > 1-\delta_0$.

Next, consider $\hat{\tau}$. For any $n>N_0$,
\begin{align*}
\|\nabla h(\psi(\hat{P}_n), \hat{\zeta})\|_4 & \leq \|\nabla h(\psi(P), \hat{\zeta})\|_4 + \|\nabla h(\psi(P), \hat{\zeta}) - \nabla h(\psi(\hat{P}_n), \hat{\zeta})\|_4 \\
& \leq \|\nabla h(\psi(P), \hat{\zeta})\|_4 + L \tau_0,
\end{align*}
where $\hat{\zeta}$ follows distribution $\hat{P}_n$. By the law of large numbers, for the first term, we have 
\begin{equation*}
    \|\nabla h(\psi(P), \hat{\zeta})\|_4 \xrightarrow{n\rightarrow \infty}\|\nabla h(\psi(P), \zeta)\|_4 \leq \tau_0.
\end{equation*}
As a result, $\|\nabla h(\psi(P), \hat{\zeta})\|_4 $ is bounded by a constant not depending on $n$. So we conclude that there exists such a $\hat{\tau}$ that $\|\nabla h(\psi(\hat{P}_n), \hat{\zeta})\|_4 \leq \hat{\tau}$ for all $n> N_0$.

To determine $\hat{C}$, it suffices to find $\hat{\lambda}_1, \hat{\lambda}_2, \hat{l}$ and $\hat{L}$, since $\alpha$ and $\eta$ are not affected by the underlying data distribution $P$.

Consider $\hat{\lambda}_1$ and $\hat{\lambda}_2$ first. $S(x) \triangleq \mathbb E_{\zeta\sim P} [\nabla h(x, \zeta)(\nabla h(x, \zeta))^\top]$ and the corresponding empirical version $S_n(x) \triangleq \mathbb E_{\hat{\zeta}\sim \hat P_n} [\nabla h(x, \hat{\zeta})(\nabla h(x, \hat{\zeta}))^\top]$. We have:
\begin{align}\label{eqt:SnDecomposition}
    \| (S_n(\psi(\hat{P}_n)) - S(x^*)) \|
    \leq \|(S_n(\psi(\hat{P}_n)) - S_n(x^*)) \| + \|( S_n(x^*) - S(x^*)) \|.
\end{align}
Consider the first term on the right-hand side of \eqref{eqt:SnDecomposition}. We have:
\begin{align*}
    & \|(S_n(\psi(\hat{P}_n)) - S_n(x^*)) \| \\
    = & \|\mathbb{E}_{\hat{\zeta}\sim \hat{P}_n}[\nabla h(\psi(\hat{P}_n), \hat{\zeta})(\nabla h(\psi(\hat{P}_n), \hat{\zeta}))^\top - \nabla h(x^*, \hat{\zeta})(\nabla h(x^*, \hat{\zeta}))^\top]\| \\
    \leq & \|\mathbb{E}_{\hat{\zeta}\sim \hat{P}_n}[\nabla h(\psi(\hat{P}_n), \hat{\zeta})(\nabla h(\psi(\hat{P}_n), \hat{\zeta}))^\top - \nabla h(x^*, \hat{\zeta})(\nabla h(\psi(\hat{P}_n), \hat{\zeta}))^\top]\| \\
    &\ + \|\mathbb{E}_{\hat{\zeta}\sim \hat{P}_n}[\nabla h(\psi(\hat{P}_n), \hat{\zeta})(\nabla h(x^*, \hat{\zeta}))^\top - \nabla h(\psi(P), \hat{\zeta})(\nabla h(x^*, \hat{\zeta}))^\top]\|\\
    \leq & \mathbb{E}_{\hat{\zeta}\sim \hat{P}_n}[\|\nabla h(\psi(\hat{P}_n), \hat{\zeta}) - \nabla h(x^*, \hat{\zeta}) \| \| \nabla h(\psi(\hat{P}_n), \hat{\zeta})\|] \\
    &\ + \mathbb{E}_{\hat{\zeta}\sim \hat{P}_n}[\|\nabla h(\psi(\hat{P}_n), \hat{\zeta}) - \nabla h(x^*, \hat{\zeta}) \| \| \nabla h(x^*, \hat{\zeta})\|] \\
    \leq & \mathbb{E} [L \|\psi(\hat{P}_n) - x^*\| \| \nabla h(\psi(\hat{P}_n), \hat{\zeta})\|] + \mathbb{E} [L \|\psi(\hat{P}_n) - x^*\| \| \nabla h(\psi(P), \hat{\zeta})\|] \xrightarrow{P} 0.
\end{align*}
In the above derivations, the first and second inequalities follow standard inequalities. The third inequality follows from \Cref{assumption: strongly convex objective with bounded hessian}. The last convergence holds since $\|\psi(\hat{P}_n) - x^*\|$ converges to 0 in probability and the remaining factors are bounded.

The second term of \eqref{eqt:SnDecomposition}, $\|(S_n(x^*) - S(x^*))\|$, converges to 0 almost surely by the law of large numbers. As a result, for any $\epsilon_1, \delta_1 > 0$, there is $N=N(\epsilon_1, \delta_1, \epsilon_0, \delta_0) > N_0$ such that,  $\forall n> N$, eigenvalues of $S_n(\hat \psi(\hat P_n))$ lies in $[\frac{1}{1+\epsilon}\lambda_1, (1+\epsilon)\lambda_2]$ with probability at least $1-\delta_1$. So we can choose $\hat{\lambda}_1 = \frac{1}{1+\epsilon}\lambda_1$ and $\hat{\lambda}_2 = (1+\epsilon)\lambda_2$.

In a similar manner, we can obtain desired constants $\hat{l}$ and $\hat{L}$ and conclude the proof for \eqref{pf: thm1 claim2}. 

Now, we consider $\alpha = 1$. Similar to the case when $\alpha \in (\frac{1}{2}, 1)$, it suffices to show that there exists a 4-tuple $(\hat{\tau}_0, \hat{\tau}, \hat{C}, N)$ such that for any $\epsilon > 0$:
\begin{equation}
    \label{pf: thm1 claim2 alpha1}
    \sup_{n > N} |\mathbb{P}^* (\sqrt{n} (\psi_n (\hat{P}_n) - \psi(\hat{P}_n)) \in D) - \mathbb{P}^*(\hat{Z}_n \in D)| \leq \hat{C} (d^{\frac{3}{2}} + \hat{\tau}^3 + \hat{\tau}_0^3) n^{-\frac{1}{2}+\epsilon}d^{\frac{1}{2}}.
\end{equation}
Theorem 3.4 in \cite{shao2022berry} gives that when $\alpha = 1$, for all $\epsilon>0$,
\begin{align}\label{eqt: A3 shao 1}
\begin{split}
    |\mathbb{P}(\sqrt{n}(\psi_n(P) - \psi(P))\in D) - \mathbb{P}(Z_0\in D)|
    \leq C(d^{3/2} + \tau^3 + \tau_0^3) n^{-\frac{1}{2}+\epsilon} d^{\frac{1}{2}} ,
\end{split}
\end{align}
where $C = C(\eta, \lambda_1, \lambda_2, \alpha, l, L)>0$ is a constant independent of $D$, $d$, $\tau$, and $\tau_0$. By the same arguments, we can derive \eqref{pf: thm1 claim2 alpha1} from \eqref{eqt: A3 shao 1}.
\end{proof}

\subsection{Proof of Theorem \ref{thm:nonasymptotic bound} in the SGD Case}\label{sup: SGD}
Before proving \Cref{thm:nonasymptotic bound} in the SGD case, we state Corollary 2.3 from \cite{shao2022berry} that plays a critical role in our proof. 
\begin{lemma}\label{lem: shao lemma 4}
\citep{shao2022berry} Let $T$ be a d-dimensional statistic. Suppose that $T = T(\zeta_1, \dots, \zeta_n)$ admits the decomposition $T=W + D$, where $W = \sum_{i=1}^n f_i(\zeta_i)$ satisfies:
\[
\mathbb{E}[f_i(\zeta_i)] = 0, \forall i,\quad \text{and } \sum_{i=1}^n \mathbb{E}[f_i(\zeta_i)(f_i(\zeta_i))^\top] = \Sigma_n,
\]
where $0<\sigma_n = \lambda_{min} (\Sigma_n)$, and $f$ is some mapping to $\mathbb{R}^d$. Define $\gamma_n = \sum_{i=1}^n \mathbb{E}\|f_i(\zeta_i)\|^3$. Let $\Delta$ and $(\Delta^{(i)}){1\leq i \leq n}$ be random variables such that $\Delta \geq \|D\|$ and $\Delta^{(i)}$ is independent of $\zeta_i$. Let $Z_0\sim \mathcal{N}(0, I_d)$.
Then, for any Borel set $D$,
\begin{align*}
& | \mathbb{P}(\Sigma_n^{-1/2} T\in D) - \mathbb{P}(Z_0\in D) | \\
\leq & \ 259 \sigma_n^{-3/2} d^{1/2} \gamma_n + 2\sigma_n^{-1} \mathbb{E}[\|W\|\Delta]  + 2\sigma_n^{-1} \sum_{i=1}^n \mathbb{E} [\|f_i(\zeta_i)\| |\Delta - \Delta^{(i)}|].
\end{align*}
\end{lemma}
The above lemma gives the Berry-Esseen bound for any statistic $T$ that has a specific decomposition. Now we are ready to prove \Cref{thm:nonasymptotic bound} in the SGD case.
\begin{proof}
We start with the following recursive formula derived from \eqref{SA iteration}:
\begin{equation}\label{eqt: sgd recursive relation}
x_{k+1} - x^* = x_k - x^* - \eta_k G(x^*) (x_k - x^*) - \eta_k \delta_k - \eta_k E_k,
\end{equation}
where $\delta_k \triangleq \delta(x_k) = \nabla H(x_k) - G(x^*)(x_k-x^*)$, and $E_{k-1} = \nabla h(x_{k-1}, \zeta_k) - \nabla H(x_{k-1})$. 

% We simplify the problem first. 
Let $G(x^*) = U \Lambda U^\top $ be the singular value decomposition of $G(x^*)$, where $\Lambda$ is the diagonal matrix consisting of eigenvalues of $G(x^*)$. Then, \eqref{eqt: sgd recursive relation} becomes $d$ parallel updates with respect to $x'_n = U^\top x_n$, $\nabla h' = U^\top \nabla h$. As a result, we can assume $d=1$ without loss of generality. In this case, $G(x^*)$ becomes a scalar, and we let $\alpha_1 = G(x^*)$, using the notation $\alpha_1$ later on so that it is not recognized as a matrix.

The goal is to show that
\[
\sup_t|\mathbb{P}^*(\sqrt{n}(\psi_n (\hat{P}_n) - {\psi}(\hat{P}_n))\leq t) - \mathbb{P}^* (\hat{Z}_n \leq t)| \rightarrow 0,\ \text{in probability as } n\rightarrow \infty,
\]
where $\hat{Z}_n \sim \mathcal{N}(0, \tilde{\sigma}^2 (\hat{P}_n))$, and $\tilde{\sigma}^2(P) = \eta^2 (2\eta \alpha_1 - 1)^{-1}\text{Var}_{\zeta\sim P}(\nabla h(\psi(P), \zeta))$ for generic distribution $P$.

Define $\beta_{mn} = \prod_{j=m+1}^n (1-\eta_j \alpha_1) \in \mathbb{R}$ and $h_n = (\sum_{m=1}^n \alpha_1^2 \eta_m^2  \beta_{mn}^2)^{-\frac{1}{2}}\in\mathbb{R}$. We have the following closed-form expression of $h_n x_{n+1}$,
\[
h_n (x_{n+1}-x^*) = h_n  \beta_{0n}(x_1-x^*) - h_n \sum_{m=1}^n \eta_m \beta_{mn}\delta_m - h_n \sum_{m=1}^n \eta_m \beta_{mn} E_m.
\]
To simplify the notation later on, let $I^{(1)}_n$, $I^{(2)}_n$, and $I^{(3)}_n$ represent the three terms on the right-hand side respectively. Namely, $I^{(1)}_n = h_n \beta_{0,n} (x_1-x^*)$, $I^{(2)}_n = h_n \sum_{m=1}^n \eta_m \beta_{mn} \delta_m$ and $I^{(3)}_n = h_n \sum_{m=1}^n \eta_m \beta_{mn} E_m$.

Let $Y_n(P) = \sqrt{n}(\psi_n (P) - \psi(P))$, and define $I^{(i)}_n(P)$ in a similar manner when addressing the underlying data distribution. Note that $I^{(1)}_n$ is deterministic and does not depend on $P$. Consider the following decomposition:
\begin{align}
\begin{split}
\label{eqt: SGD split 1}
    & |\mathbb{P}^*(\sqrt{n}(\psi_n (\hat{P}_n) - \psi(\hat{P}_n))\leq t) - \mathbb{P}^* (Z_n \leq t)|\\
    \leq & |\mathbb{P}^*(Y_n(\hat{P}_n)  \leq t) - \mathbb{P}^*(Y_n(\hat{P}_n)  - \frac{\sqrt{n}}{h_n} (I^{(1)}_n  -I^{(2)}_n(\hat{P}_n) ) \leq t)|\\
    &\ +|\mathbb{P}^*(Y_n(\hat{P}_n)  - \frac{\sqrt{n}}{h_n} (I^{(1)}_n  -I^{(2)}_n(\hat{P}_n) ) \leq t) - \mathbb{P}^*(Z_n \leq t)|.
\end{split}
\end{align}
The remainder of this proof will show that both terms on the right-hand side converge to 0 in probability.

To see that $|\mathbb{P}^*(Y_n(\hat{P}_n)  \leq t) - \mathbb{P}^*(Y_n(\hat{P}_n)  - \frac{\sqrt{n}}{h_n} (I^{(1)}_n(\hat{P}_n)  -I^{(2)}_n(\hat{P}_n) ) \leq t)|\xrightarrow{P} 0$,
it suffices to verify that $I^{(1)}_n \rightarrow 0$ and $I^{(2)}_n(\hat{P}_n) \rightarrow 0 $ conditional on $\zeta_1, \zeta_2, \dots$, in probability. This follows from the $(3.9a)$ and $(3.9b)$ in \cite{sacks1958asymptotic}.

It remains to show that $|\mathbb{P}^*(Y_n(\hat{P}_n)  - \frac{\sqrt{n}}{h_n} (I^{(1)}_n(\hat{P}_n)  -I^{(2)}_n(\hat{P}_n) ) \leq t) - \mathbb{P}^*(Z_n \leq t)| \xrightarrow{P} 0$.
We will make use of \Cref{lem: shao lemma 4} to prove this claim.
Following the framework for proving the ASGD case, it suffices to establish a bound similar to \eqref{eqt: A3 shao}. To be specific, we need to show that, for any Borel set $D$,
\[
|P(-\frac{\sqrt{n}}{h_n}I^{(3)}_n(P)\in D) - P(\hat{Z}_n\in D)| \leq C_n,
\]
where $C_n$ converges to 0 as $n\rightarrow \infty$, and $C_n$ (to be specified) depends on $\eta, \lambda_1, \lambda_2, l, L, \tau$, and $\tau_0$.
To simplify the notation, let 
\[
    A_n = \nabla h (x^*, \zeta_n),\quad B_n  = E_n - A_n = \nabla h(x_{n-1},\zeta_n) - \nabla h(x^*, \zeta_n) - \nabla H (x_{n-1}).
\]
We have
\[
I^{(3)}_n = h_n \sum_{m=1}^n \eta_m \beta_{mn} A_m + h_n \sum_{m=1}^n \eta_m \beta_{mn} B_m.
\]
Let $Q_{mn} = \eta_m \beta_{mn}$, and $\bar{\Sigma}_n = h_n^2 \sum_{m=1}^n Q_{mn}^2 \Tilde{\sigma}(P) $. Define 
\[
T_n \triangleq \bar{\Sigma}_n^{-\frac{1}{2}} I_n^{(3)} = W_n + D_n,
\]
where $W_n = \sum_{m=1}^n Y_{mn}$, $Y_{mn} = h_n \bar{\Sigma}_n^{-\frac{1}{2}} Q_{mn}A_m$, and $D_n = h_n \sum_{m=1}^n Q_{mn} B_m \bar{\Sigma}_n^{-\frac{1}{2}}$. Notice that $\{Y_{mn}\}_{m=1}^n$ are independent and that
\[
\begin{cases}
    \mathbb{E}[W_n] = 0\\
    Var (W_n) = 1\\
    T_n = W_n + D_n
\end{cases}
\]
Now, we construct $\Delta$ and $\{\Delta^{(i)}\}$ that will later be useful when applying \Cref{lem: shao lemma 4}. The idea is borrowed from \cite{shao2022berry}.
Let $(\zeta_1', \dots,\zeta_n')$ be an independent copy of $(\zeta_1, \dots, \zeta_n)$ and define $(A_1', \dots, A_n')$ according to the relationship
$
A_i' = \nabla h (x^*, \zeta_i')$, $ i=1,\dots,n
$.
For each $i$, construct
$x_1^{i},\dots,x_n^{i}$ as follows:
\begin{itemize}
    \item If $j<i$, $x_j^{(i)} = x_j$
    \item If $j=i$, $x_j^{(i)} = x_{j-1}^{(i)} - \eta_j (\nabla H(x_j^{(i)}) + A'_j + B(x_j^{(i)}, A'_j))$
    \item If $j>i$, $x_j^{(i)} = x_{j-1}^{(i)} - \eta_j (\nabla H(x_j^{(i)}) + A_j + B(x_j^{(i)}, A_j))$
\end{itemize}
That is, $(x_1^{i},\dots,x_n^{i})$ is obtained by running SGD with only the $i$-th data replaced by $\zeta_i'$. It is worth noting that this construction is only for the sake of establishing the Berry-Esseen type bound, and it is not required in real applications.
For each $i=1,\dots,n$, we define $T_n^{(i)}$, $W_n^{(i)}$ and $D_n^{(i)}$ following the same procedure described above. And notice that $\forall 1\leq i \leq n,\ D_n^{(i)} \indep \zeta_i$. Let $\Delta \triangleq \|D_n\|$ and $\Delta^{(i)}\triangleq \|D_n^{(i)}\|$. Then $\Delta^{(i)} \indep \zeta_i,\ \forall i=1,\dots,n$.

Applying \Cref{lem: shao lemma 4} to $T_n = W_n + D_n$, $\Delta$, and $\{\Delta^{(i)}\}$ defined above, we obtain the following inequality:
\begin{equation}
\label{eqt: berry-esseen ineq}
| \mathbb{P}(T_n\in D) - \mathbb{P}(Z_0\in D) | \leq 259 \gamma_n + 2 \mathbb{E}[\|W_n\|\Delta] + 2 \sum_{i=1}^n \mathbb{E} [\|Y_{in}\| |\Delta - \Delta^{(i)}|],
\end{equation}
where $\gamma_n = \sum_{i=1}^n \mathbb{E}|Y_{in}|^3 = h_n \bar{\Sigma}_n^{-\frac{3}{2}} Q_{in}^3 |A_m|^3$ and $Z_0$ follows a standard normal distribution. To establish the desired conclusion, it remains to prove that the following three terms vanish as $n\rightarrow \infty$:
\begin{itemize}
    \item $\gamma_n$
    \item $\mathbb{E}[\|W_n\|\Delta]$ 
    \item $\sum_{i=1}^n \mathbb{E} [\|Y_{in}\| |\Delta - \Delta^{(i)}|]$
\end{itemize}
Consider $\gamma_n$ first. By the definition of $Y_{mn}$ and $\bar{\Sigma}_n$, we have
\begin{align*}
    \gamma_n &= \sum_{m=1}^n \mathbb{E}[|Y_{mn}|^3]
    = (\sum_{i=1}^n Q^2_{in}\bar\Sigma^2)^{-\frac{3}{2}} \sum_{m=1}^n Q^3_{mn}\mathbb{E}[|A_m|^3].
\end{align*}
Since $Q_{mn} = \eta_m \beta_{mn} = \eta m^{-1} \prod_{j=m+1}^n (1-\alpha_1 \eta j^{-1})$, it is not hard to derive the following sandwich inequality (see, e.g. (2.3) in \cite{sacks1958asymptotic})
\[
(1-\epsilon'_m) m^{\alpha_1 \eta -1 } n^{-\alpha_1 \eta} \leq Q_{mn} \leq (1+\epsilon'_m) m^{\alpha_1 \eta -1 } n^{-\alpha_1 \eta},
\]
for any $n\geq m$, where $\epsilon'_m \rightarrow 0 $ as $m\rightarrow \infty$.
By \Cref{assumption: gradient variability}, $\mathbb{E}[|A_m|^3] \leq \tau^3$, so there exists constant $C_\gamma>0$ such that
\[
\gamma_n \leq C_\gamma\sum_{j=1}^n j^{3\alpha_1\eta -3} (\sum_{m=1}^n m^{2\alpha_1\eta -2})^{-\frac{3}{2}}.
\]
The magnitude of the above expression depends on the value of $\alpha_1 \eta$. To be specific, we have
\[
\gamma_n \leq 
\begin{cases}
C_{\gamma} n^{-\frac{1}{2}} & \text{if } \alpha_1\eta > \frac{2}{3}\\
C_\gamma n^{-\frac{1}{2}}\log n& \text{if } \alpha_1\eta = \frac{2}{3}\\
C_\gamma n^{-3\alpha_1\eta + \frac{3}{2}} & \text{if } \frac{1}{2} < \alpha_1\eta < \frac{2}{3}\\
C_\gamma (\log n)^{-\frac{3}{2}} & \text{if } \alpha_1 \eta = \frac{1}{2}\\
C_\gamma & \text{if } \alpha_1\eta < \frac{1}{2}
\end{cases}
\]
for some $C_\gamma$ depending on $\eta, \lambda_1, \lambda_2, \alpha, l, L$.

Now, we consider $\mathbb{E}[\|W_n\|\Delta]$. There exists constant $c'>0$ such that
\begin{align*}
 \mathbb{E}[|W_n| \Delta]
    \leq &(\mathbb{E}[W_n^2] \mathbb{E}[\Delta^2])^{\frac{1}{2}} \\
    = & (\mathbb{E}[\Delta^2])^\frac{1}{2} \\
    = & (\mathbb{E}[(h_n \sum_{m=1}^n Q_{mn} B_m \bar{\Sigma}_n^{-\frac{1}{2}})^2])^\frac{1}{2} \\
    = & (\mathbb{E}[(\bar\Sigma^{-\frac{1}{2}} (\frac{(\sum_{i=1}^n Q_{in} B_i)^{2}}{\sum_{m} Q_{mn}^2}))^2])^\frac{1}{2} \\
    = &  (\mathbb{E}[\bar\Sigma^{-\frac{1}{2}} \frac{\sum_{m=1}^n Q_{mn}^2 B_m^2 }{\sum_{m=1}^n Q_{mn}^2}])^{\frac{1}{2}} \\
    = &  (\bar\Sigma^{-\frac{1}{2}} \frac{\sum_m Q_{mn}^2 \mathbb{E}[B_m^2]}{ \sum_m Q_{mn}^2})^{\frac{1}{2}} \\
    \leq &c' (\frac{\sum_{i=1}^n i^{2\alpha_1 \eta -2}\mathbb{E}[B_m^2]}{\sum_{m=1}^n m^{2\alpha_1\eta -2}})^{\frac{1}{2}}.
\end{align*}
For any $m$, by \Cref{assumption: strongly convex objective with bounded hessian},
\begin{align*}
|B_m| =& |\nabla h(x_{m-1}, \zeta_m) - \nabla h(x^*, \zeta_m) - \nabla H(x_{m-1}) + \nabla H(x^*)|\\
\leq & |\nabla h(x_{m-1}, \zeta_m) - \nabla h(x^*, \zeta_m)|+|\nabla H(x_{m-1}) - \nabla H(x^*)|\\
\leq & 2L |x_{m-1} - x^*|.
\end{align*}
Combining Lemma 5.12 in \cite{shao2022berry} with the above inequality, we get:
\[
\mathbb{E} [B_m^2]\leq 2L\mathbb{E}[|x_{m-1} - x^*|^2]\leq 
\begin{cases}
c n^{-1} & \text{if }  \alpha_1 \eta > 1,\\
c n^{-1} \log n & \text{if }  \alpha_1 \eta = 1,\\
c n^{-\alpha_1 \eta} & \text{if } \alpha_1 \eta < 1,
\end{cases}
\]
for some constant $c=c(\tau,\tau_0,L)>0$. Thus, there exists constant $\hat{c}=\hat{c}(\tau,\tau_0,L)>0$ such that:
\begin{equation*}
    \mathbb{E}[|W_n|\Delta] \leq\begin{cases}
    \hat{c}n^{-\frac{1}{2}} & \text{if } \alpha_1\eta > 1,\\
    \hat{c}n^{-\frac{1}{2}} (\log^2n) & \text{if } \alpha_1 \eta = 1,\\
    \hat{c}n^{-2\alpha_1 \eta + 1} & \text{if } \alpha_1 \eta \in (\frac{1}{2}, 1),\\
    \hat{c} (\log n)^{-1} & \text{if } \alpha_1 \eta=\frac{1}{2},\\
    \hat{c} & \text{if } \alpha_1 \eta < \frac{1}{2}.
    \end{cases}
\end{equation*}
For the third term, consider $\mathbb{E}[Y^2_{in}]$ first, we have:
\begin{align*}
    \mathbb{E}[Y_{in}^2] & = \mathbb{E}[\bar\Sigma^{-2} (\sum_{m=1}^n Q_{mn}^2)^{-1} Q_{in}^2 A_i^2]\\
    & = \bar\Sigma^{-2} (\sum_{m=1}^n Q_{mn}^2)^{-1} Q_{in}^2 \mathbb{E}[A_i^2]\\
    & \leq c (\sum_{m=1}^n m^{2\alpha_1\eta -2} n^{-2\alpha_1 \eta})^{-1} i^{2\alpha_1\eta - 2} n^{-2\alpha_1\eta}\\
    &= c (\sum_{m=1}^n m^{2\alpha_1\eta -2} )^{-1} i^{2\alpha_1\eta - 2} .
\end{align*}
Consider $\mathbb{E}[|\Delta - \Delta^{(i)}|^2]$. By definition of $\Delta$ and $\Delta^{(i)}$:
\begin{align*}
    \mathbb{E}[|\Delta - \Delta^{(i)}|^2] &= \mathbb{E}[ \frac{ ( \sum_{m=1}^n Q_{mn} (B_m - B_m^{(i)}) )^2 }{ \sum_{m=1}^n Q_{mn}^2 } ].
\end{align*}
Since ${(B_m - B_m^{(i)})}_m$ forms a martingale difference sequence for any $i$ with respect to its canonical filtration, the crossing terms in the above complete square are zero. By definition of $B_m$ and $B_m^{(i)}$, we have:
\[
B_m - B_m^{(i)} = 
\begin{cases}
0 & m < i,\\
B(x_{m-1}, A_m) - B(x_{m-1}, A'_m) & m=i,\\
B(x_{m-1}, A_m) - B(x_{m-1}^{(i)}, A_m) & m>i.
\end{cases}
\]
As a result,
\begin{align*}
     \mathbb{E}[|\Delta - \Delta^{(i)}|^2] &= \frac{Q_{in}^2}{\sum_{m=1}^n Q_{mn}^2} \mathbb{E}[(B_i - B_i^{(i)})^2] + \frac{1}{\sum_{m=1}^n Q_{mn}^2} \sum_{m>n}Q^2_{mn} \mathbb{E}[(B_m - B_m^{(i)})^2].
\end{align*}
Lemma 5.13 in \cite{shao2022berry} provides the following bound for the martingale difference term $(B_j - B_j^{(i)})$:
\begin{equation*}
    \mathbb{E}[(B_j - B_j^{(i)})^2] \leq c(\tau^2 + \tau_0^2) i^{-2} (\frac{i}{j})^{2\alpha_1 \eta}.
\end{equation*}
Combining the above two relationships, there is some constant $\hat{c}$ such that
\begin{equation*}
 \mathbb{E}[|\Delta - \Delta^{(i)}|^2] \leq \hat{c} (\tau^2 + \tau_0^2) \frac{1}{\sum_m m^{2\alpha_1 \eta - 2}} i^{2\alpha_1\eta-3}.
\end{equation*}
Now we are ready to bound $\sum_i \mathbb{E}[|Y_{in}| |\Delta - \Delta^{(i)}|]$. There exists constant $\hat{c}$, depending on $\eta$, $\lambda_1$, $\lambda_2$, $l, L, \tau, \tau_0$, such that:
\begin{align*}
\sum_i \mathbb{E}[|Y_{in}| |\Delta - \Delta^{(i)}|]
& \leq \sum_i (\mathbb{E}[Y_{in}^2] \mathbb{E}[(\Delta - \Delta^{(i)})^2])^\frac{1}{2}\\
& \leq \hat{c} (\sum_m m^{2\alpha_1 \eta -2})^{-1} \sum_i i^{2\alpha_1\eta - 5/2}.
\end{align*}
So we have the following vanishing rate for $\sum_i \mathbb{E}[|Y_{in}| |\Delta - \Delta^{(i)}|]$, depending on the choice of $\alpha_1 \eta$:
\[
\sum_i \mathbb{E}[|Y_{in}| |\Delta - \Delta^{(i)}|] \leq 
\begin{cases}
\hat{c} n^{-\frac{1}{2}} & \text{ if } \alpha_1\eta > \frac{3}{4},\\
\hat{c} n^{-\frac{1}{2}} \log n & \text{ if } \alpha_1\eta = \frac{3}{4},\\
\hat{c} n^{-2\alpha_1\eta + 1} & \text{ if } \frac{1}{2}<\alpha_1\eta < \frac{3}{4},\\
\hat{c} (\log n)^{-1} & \text{ if } \alpha_1\eta = \frac{1}{2},\\
\hat{c} & \text{ if } 0 < \alpha_1\eta < \frac{1}{2}.
\end{cases}
\]
Thus, we have established the vanishing property of the right-hand-side of \eqref{eqt: berry-esseen ineq}, which we now denote by $C_n$. Specifically,
\[
|P(\Bar{\Sigma}_n^{-\frac{1}{2}} I^{(3)}_n\in D) - P(Z_0\in D)| \leq C_n.
\]
Therefore, for any measurable set $D$,
\[
|P(-\frac{\sqrt{n}}{h_n}I^{(3)}_n\in D) - P(-\frac{\sqrt{n}}{h_n} \Bar{\Sigma}_n^\frac{1}{2}Z_0\in D)| \leq C_n.
\]
By \cite{sacks1958asymptotic}, $\frac{\sqrt{n}}{h_n} \Bar{\Sigma}_n^\frac{1}{2}\rightarrow \eta (2\eta \alpha_1 - 1)^{-\frac{1}{2}}\Tilde{\sigma}(P)^\frac{1}{2}$ as $n\rightarrow\infty$. This concludes the proof.
\end{proof}
\subsection{Proof of Theorem \ref{thm: two-stage guarantee}}
\label{subsection: two-stage proof}
\Cref{thm: two-stage guarantee} is an immediate consequence of the following result regarding the consistency of Lasso.
\begin{lemma}\citep{zhao2006model}
    \label{lem: lasso consistency}
    Under \Cref{assumption: lasso correctness,assumption: strong irrepresentable condition}, we have
    \[
    \lim_{n\rightarrow\infty}\mathbb{P}(\hat{x}^{(n)}(\lambda_n) =_s x^{(n)}) = 1,
    \]
    where $=_s$ denotes equality in sign, which holds true if and only if the values on both sides are either both positive, both negative, or both zero.
\end{lemma}
Let $i\notin T^*$, and notice that $\mathbb{P}(\hat{x}^{(n)}(\lambda_n) \leq \mathbb{P}(x^{(n)}_i = 0)$.
\Cref{lem: lasso consistency} directly implies that $\mathbb{P}(x^{(n)}_i \in \mathcal{I}_{i,n}) \rightarrow 1$ for $i\notin T^*$. For $i\in T^*$, $\mathbb{P}(x^{(n)}_i \in \mathcal{I}_{i,n}) \rightarrow 1-\gamma$ follows by combining \Cref{lem: lasso consistency,thm: main theorem}.

\section{Additional Numerical Results}\label{app: additional numerical results}
Please refer to \Cref{SMtable: table 1} for full results for linear regression, \Cref{SMtable: logistic regression table 3} for logistic regression, and \Cref{SMtable: Sparse Linear Regression} for sparse linear regression.

\section{Additional Discussion on Assumptions and Further Sensitivity Analyses}
\label{app: additional discussion on assumptions}
% able, , which is stronger than what is strictly required. Conceptually, the essential requirement for our analysis is that
\Cref{assumption: strongly convex objective with bounded hessian} imposes uniform strong convexity of the sample loss $h(\cdot, \zeta)$. This assumption can potentially be relaxed. Essentially, we need, with high probability, that the empirical Hessians for both the original and bootstrap samples to be well-conditioned, i.e., 
\[
\lambda_{min} (G_n(x^*)) \geq c, \lambda_{min} (G^{*b}(x^*)) \geq c,
\]
for some $c>0$ for all bootstrap replicates $b$, together with local smoothness of $H$ and stability of the (A)SGD iterates near $x^*$. Under these conditions, the nonlinear remainder term involving $\delta_m$ in \eqref{eqt: sgd closed form} in the error analysis is summable along the (A)SGD trajectory and asymptotically negligible at the $\sqrt{n}$ scale with probability tending to one sufficiently fast, so that the uniform control over all bootstrap replicates required later in the proof remains valid.
% (THIS DEPENDS ON THE DECAY OF THE HIGH PROBABILITY, CORRECT?). . Moreover, one needs the probability of the complement event to vanish sufficiently fast
Establishing our results using these relaxed conditions would be delegated to future work.
% We adopt the stronger sample-wise curvature assumption in the main body to guarantee these properties uniformly and to simplify the presentation. Establishing high-probability conditioning on $G_n$ and bootstrap version $G^*_n$ under weaker assumptions requires more delicate arguments.

In the following subsections, we present additional numerical experiments to illustrate the behavior of the proposed method when key assumptions are violated, mainly to investigate the robustness of our proposed methods.
\subsection{Ill-conditioned Curvature}
We first examine the effect of ill-conditioned curvature by considering ridge regression with varying levels of conditioning in the feature covariance matrix $\Sigma$. Throughout, we quantify conditioning by the condition number 
\[
\kappa(\Sigma) = \lambda_{\max}(\Sigma) / \lambda_{\min}(\Sigma),
\]
where $\lambda_{\min}(\Sigma)$ and $\lambda_{\max}(\Sigma)$ denote the smallest and largest eigenvalues of $\Sigma$ respectively.
Specifically, we consider the model
\[
b = a^\top x^* + \varepsilon, \quad a \sim \mathcal{N}(0, \Sigma),\quad \varepsilon \sim \mathcal{N}(0, \sigma^2),
\]
and minimize the ridge regression objective
\[
H_\lambda(x) = \mathbb{E}\big[\frac{1}{2}\mathbb{E}[(a^\top x - b)^2]\big] + \frac{\lambda}{2} \|x\|^2.
\]
Under this model, the population Hessian of $H_\lambda$ equals $\Sigma + \lambda I$. $\kappa(\Sigma)$ governs the anisotropic curvature induced by the data, while $\lambda$ adds isotropic curvature that improves conditioning by increasing small eigenvalues. In particular, increasing $\kappa(\Sigma)$ makes the problem more ill-conditioned, whereas increasing $\lambda$ mitigates ill-conditionedness by shrinking $\kappa(\Sigma + \lambda I)$. 

We construct $\Sigma$ as an equicorrelation matrix with unit diagonal to isolate the effect of curvature rather than feature scaling. 
% We control the well-posedness of the problem with the condition number $\kappa(\Sigma)$. 
Coverage is evaluated with respect to $x^*_\lambda$ that minimizes the ridge regression objective. \Cref{tab:COfB_ridge,tab:COnB_ridge} report empirical coverages with nominal level $95\%$ and average interval widths of COfB and COnB, respectively. The upper and right parts of each table reflect the well-conditioned regime, and the lower-left portion corresponds to the most ill-conditioned setting.

The results illustrate some robustness of both methods to curvature degradation, but also caution the sensitivity to extreme ill-conditionedness. When the problem is well-conditioned ($\kappa(\Sigma) = 10$), both COfB and COnB achieve near nominal coverages for all choices of $\lambda$, with short confidence intervals. As $\kappa(\Sigma)$ increases, coverages deteriorate and the interval lengths grow, indicating that ill-conditioned curvature negatively impacts the performance of both methods. Nonetheless, the coverages in this setting are still moderately close to the nominal, with the interval lengths of COfB relatively small but those of COnB larger. In the most ill-conditioned case when $\kappa(\Sigma) = 10^3$ and the ridge regularization parameter $\lambda$ is small, both methods substantially undercover and COnB in particular produces very wide intervals, although its coverages are better than COfB. These observations put caution on using our methods in extreme ill-conditioned problems. 

\begin{table}[t]
\centering
\caption{COfB under curvatures with different conditioning (ridge regression).
Empirical coverages (nominal $0.95$) of componentwise confidence intervals for the ridge minimizer $x_\lambda^*$, with average interval lengths in parentheses.
Features have unit variance and equicorrelation structure with condition number $\kappa(\Sigma)$.
}
\label{tab:COfB_ridge}

\begin{tabular}{c|cccc}
\toprule
$\kappa(\Sigma)$ & $\lambda=0$ & $10^{-4}$ & $10^{-3}$ & $10^{-2}$ \\
\midrule
$10$ & 0.975 (0.02) & 0.975 (0.02) & 0.975 (0.02) & 0.975 (0.02) \\
$10^2$ & 0.915 (0.08) & 0.915 (0.08) & 0.915 (0.08) & 0.915 (0.07) \\
$10^3$ & 0.735 (0.67) & 0.745 (0.66) & 0.765 (0.64) & 0.845 (0.47) \\
\hline
\bottomrule
\end{tabular}
\end{table}

\begin{table}[t]
\centering
\caption{COnB under curvatures with different conditioning (ridge regression).
Empirical coverages (nominal $0.95$) of componentwise confidence intervals for the ridge minimizer $x_\lambda^*$, with average interval lengths in parentheses.
All settings match \Cref{tab:COfB_ridge}.
% (MAKE SURE THIS IS UNDERSTANDABLE).
}
\label{tab:COnB_ridge}

\begin{tabular}{c|cccc}
\toprule
$\kappa(\Sigma)$ & $\lambda=0$ & $10^{-4}$ & $10^{-3}$ & $10^{-2}$ \\
\midrule
$10$ & 0.965 (0.03) & 0.965 (0.03) & 0.965 (0.03) & 0.965 (0.03) \\
$10^2$ & 0.995 (0.53) & 0.995 (0.53) & 0.995 (0.52) & 0.995 (0.47) \\
$10^3$ & 0.825 (10.34) & 0.825 (10.31) & 0.835 (9.98) & 0.915 (7.29) \\
\bottomrule
\end{tabular}
\end{table}

\subsection{Learning Rate Schemes}
In this subsection, we examine the stability of our methods to the step size assumptions by varying the learning rate schedule while keeping the objective well-conditioned. 

We consider linear regression with a well-conditioned design, fixing $d=20$, $n=10^5$, $\kappa(\Sigma)=10$. We evaluate a range of learning rate schemes that may not be covered by our theory, including power-law decay with $\alpha=0.501$ (covered by our theory), and $\alpha=0.25$ and a constant step size. In addition, we also consider piecewise constant schedule and cosine annealing. Empirical coverages and average confidence interval lengths of the solutions are reported for both COfB and COnB. \Cref{tab:COfB_lr} and \Cref{tab:COnB_lr} report empirical coverages with nominal level $95\%$ and average interval widths of COfB and COnB, respectively.

Across all learning rate schemes considered, both COfB and COnB achieve near nominal empirical coverages, with only modest variation in confidence interval lengths. These results indicate that while the step size conditions imposed in the theory are sufficient for establishing asymptotic validity, the proposed methods exhibit substantial empirical robustness to learning rate scheme in well-conditioned, large-sample settings.

\begin{table}[t]
\centering
\caption{COfB under different learning rate schemes.
Empirical coverages (nominal $0.95$) of componentwise confidence intervals for the minimizer $x^*$ under different learning rate schemes, with average confidence interval lengths in parentheses.
All settings use $n=10^5$, $d=20$, equicorrelated $\Sigma$ with $\kappa(\Sigma)=10$, noise variance $\sigma^2=1$, and $B=5$ bootstrap replicates.
}
\label{tab:COfB_lr}

\begin{tabular}{lcc}
\toprule
Learning-rate scheme & Coverage (Avg. length) \\
\midrule
Power-law ($\alpha=0.501$)   & 0.985 (0.0188) \\
Power-law ($\alpha=0.25$)    & 0.955 (0.0187) \\
Constant step size           & 0.940 (0.0187) \\
Piecewise constant           & 0.980 (0.0216) \\
Cosine annealing             & 0.965 (0.0192) \\
\bottomrule
\end{tabular}
\end{table}

\begin{table}[t]
\centering
\caption{COnB under different learning rate schemes.
Empirical coverages (nominal $0.95$) of componentwise confidence intervals for the minimizer $x^*$ under different learning rate schemes, with average confidence interval lengths in parentheses.
All settings match \Cref{tab:COfB_lr}.
}
\label{tab:COnB_lr}

\begin{tabular}{lcc}
\toprule
Learning-rate scheme & Coverage (Avg. length) \\
\midrule
Power-law ($\alpha=0.501$)   & 0.960 (0.0218) \\
Power-law ($\alpha=0.25$)    & 0.960 (0.0169) \\
Constant step size           & 0.955 (0.0190) \\
Piecewise constant           & 0.960 (0.0217) \\
Cosine annealing             & 0.955 (0.0182) \\
\bottomrule
\end{tabular}
\end{table}

\begin{table*}
\centering
\resizebox{\linewidth}{!}{
\begin{tabular}{cc c c c cc} 
\hline
\multicolumn{7}{c}{Identity $\Sigma$, $n=10^5$} \\
 \hline
 & \multicolumn{2}{c}{$d=5$} & \multicolumn{2}{c}{$d=20$} & \multicolumn{2}{c}{$d=200$}\\ 
 & Cov (\%) & Len ($\times 10^{-2}$)& Cov (\%) & Len ($\times 10^{-2}$) & Cov (\%) & Len ($\times 10^{-2}$) \\ \hline
 delta & $\textbf{94.84}$ ($0.07$) & $1.24$ ($0.00$)& $\textbf{94.38}$ ($0.07$) & $1.24$ ($0.00$)& $\textit{71.37}$ ($0.14$) & $1.24$ ($0.00$)\\ \hline
BM & $\textbf{92.72}$ ($0.08$) & $1.23$ ($0.00$)& $\textbf{92.92}$ ($0.08$) & $1.29$ ($0.00$)& $98.34$ ($0.04$) & $3.65$ ($0.00$)\\ \hline
RS & $\textbf{94.50}$ ($0.07$) & $1.61$ ($0.01$)& $\textbf{93.92}$ ($0.08$) & $1.62$ ($0.01$)& $\textbf{96.94}$ ($0.05$) & $3.37$ ($0.02$)\\ \hline
OB $B=10$ & $\textbf{92.36}$ ($0.08$) & $1.26$ ($0.00$)& $\textbf{92.42}$ ($0.08$) & $1.26$ ($0.00$) & $99.78$ ($0.05$) &  $6.43$ ($0.01$)\\ \hline
OB $B=100$ & $\textbf{94.88}$ ($0.07$) & $1.29$ ($0.00$)& $\textbf{95.16}$ ($0.07$) & $1.29$ ($0.00$) & $100.00$ ($0.00$) & $7.30$ ($0.01$) \\ \hline
$\text{HiGrad}_{(2,2)}$ & $\textbf{95.67}$ ($0.06$) & $2.56$ ($0.01$)& $\textbf{95.29}$ ($0.07$) & $2.66$ ($0.01$)& $\textbf{96.03}$ ($0.06$) & $2.67$ ($0.01$)\\ \hline
COfB ASGD $B=3$  & $\textbf{95.44}$ ($0.07$) & $2.44$ ($0.00$)& $\textbf{94.49}$ ($0.07$) & $2.46$ ($0.00$)& $\textbf{92.31}$ ($0.08$) & $20.71$ ($16.16$)\\ \hline
COfB ASGD $B=5$  & $\textbf{95.24}$ ($0.07$) & $1.68$ ($0.00$)& $\textbf{94.55}$ ($0.07$) & $1.69$ ($0.00$)& $\textbf{94.59}$ ($0.07$) & $3.06$ ($1.18$)\\ \hline
COfB ASGD $B=10$ & $\textbf{95.04}$ ($0.07$) & ${1.40}$ ($0.00$)& $\textbf{94.82}$ ($0.07$) & ${1.42}$ ($0.00$)& $\textbf{94.30}$ ($0.07$) & $2.56$ ($0.00$)\\ \hline
COfB SGD $B=3$  & $\textbf{94.96}$ ($0.07$) & $3.16$ ($0.00$)& $\textbf{94.53}$ ($0.07$) & $3.18$ ($0.00$)& $\textbf{94.59}$ ($0.07$) & $3.20$ ($0.00$)\\ \hline
COfB SGD $B=5$  & $\textbf{94.44}$ ($0.07$) & $2.17$ ($0.00$)& $\textbf{94.30}$ ($0.07$) & $2.17$ ($0.00$)& $\textbf{94.42}$ ($0.07$) & $2.19$ ($0.00$)\\ \hline
COfB SGD $B=10$ & $\textbf{94.68}$ ($0.07$) & $1.82$ ($0.00$)& $\textbf{93.97}$ ($0.08$) & $1.83$ ($0.00$)& $\textbf{94.33}$ ($0.07$) & ${1.84}$ ($0.00$)\\ \hline
COnB $B=3$ & $\textbf{95.32}$ ($0.07$) & $1.94$ ($0.00$)& $\textbf{94.63}$ ($0.07$) & $1.97$ ($0.00$)& $99.17$ ($0.03$) & $9.15$ ($0.02$) \\ \hline  
COnB $B=5$ & $\textbf{95.04}$ ($0.07$) & $1.61$ ($0.00$)& $\textbf{95.06}$ ($0.07$) & $1.61$ ($0.00$)& $99.62$ ($0.02$) & $8.02$ ($0.01$) \\ \hline  
COnB $B=10$ & $\textbf{95.16}$ ($0.07$) & $1.40$ ($0.00$)& $\textbf{95.52}$ ($0.07$) & $1.43$ ($0.00$)& $99.88$ ($0.01$) & $7.31$ ($0.01$) \\ \hline 

\multicolumn{7}{c}{Toeplitz $\Sigma$, $n=10^5$ } \\ \hline
delta & $\textbf{94.20}$ ($0.07$) & $1.53$ ($0.00$)& $\textbf{93.23}$ ($0.08$) & $1.58$ ($0.00$)& $\textit{37.16}$ ($0.15$) & $1.60$ ($0.00$)\\ \hline
BM & $91.80$ ($0.09$) & $1.53$ ($0.00$)& $88.51$ ($0.10$) & $1.51$ ($0.00$)& $\textbf{97.37}$ ($0.05$) & $6.54$ ($0.01$)\\ \hline
RS & $\textbf{92.83}$ ($0.08$) & $1.94$ ($0.01$)& $\textbf{93.38}$ ($0.08$) & $2.15$ ($0.01$)& $\textbf{97.02}$ ($0.05$) & $8.60$ ($0.06$)\\ \hline
OB $B=10$ & $\textbf{92.48}$ ($0.08$) & $1.66$ ($0.00$)& $\textbf{93.68}$ ($0.08$) & $1.81$ ($0.00$) & $\textbf{93.33}$ ($0.25$) & $13.73$ ($0.01$)\\ \hline
OB $B=100$ & $\textbf{95.48}$ ($0.07$) & $1.72$ ($0.00$)& $\textbf{95.84}$ ($0.06$) & $1.97$ ($0.00$) & $\textbf{96.58}$ ($0.18$) & $14.26$ ($0.01$) \\ \hline
$\text{HiGrad}_{(2,2)}$ & $\textbf{94.33}$ ($0.07$) & $2.69$ ($0.01$)& $\textbf{94.88}$ ($0.07$) & $2.82$ ($0.01$)& $\textbf{94.09}$ ($0.07$) & $2.83$ ($0.01$)\\ \hline
COfB ASGD $B=3$  & $\textbf{94.72}$ ($0.07$) & $3.06$ ($0.00$)& $\textbf{94.95}$ ($0.07$) & $3.30$ ($0.00$)& $\textbf{94.41}$ ($0.07$) & $14.44$ ($0.02$)\\ \hline
COfB ASGD $B=5$  & $\textbf{95.24}$ ($0.07$) & $2.21$ ($0.00$)& $\textbf{94.89}$ ($0.07$) & $2.26$ ($0.00$)& $\textbf{93.91}$ ($0.08$) & $9.96$ ($0.01$)\\ \hline
COfB ASGD $B=10$ & ${\textbf{95.28}}$ ($0.07$) & ${1.74}$ ($0.00$)& ${\textbf{94.95}}$ ($0.07$) & ${2.12}$ ($0.00$)& $\textbf{94.01}$ ($0.08$) & $8.44$ ($0.01$)\\ \hline
COfB SGD $B=3$  & $\textbf{95.16}$ ($0.07$) & $6.93$ ($0.01$)& $\textbf{94.76}$ ($0.07$) & $4.34$ ($0.00$)& $\textbf{95.16}$ ($0.07$) & $6.90$ ($0.01$)\\ \hline
COfB SGD $B=5$  & $\textbf{95.84}$ ($0.06$) & $4.70$ ($0.00$)& $\textbf{94.20}$ ($0.07$) & $2.97$ ($0.00$)& $\textbf{95.04}$ ($0.07$) & $4.72$ ($0.00$)\\ \hline
COfB SGD $B=10$ & $\textbf{95.36}$ ($0.07$) & $3.95$ ($0.00$)& $\textbf{94.20}$ ($0.07$) & $2.50$ ($0.00$)& ${\textbf{95.00}}$ ($0.07$) & ${3.98}$ ($0.00$)\\ \hline
COnB $B=3$ & $\textbf{95.00}$ ($0.07$) & $2.49$ ($0.00$)& $\textbf{95.36}$ ($0.07$) & $2.94$ ($0.00$)& $\textbf{95.41}$ ($0.07$) & $21.04$ ($0.02$) \\ \hline 
COnB $B=5$ & $\textbf{94.60}$ ($0.07$) & $1.96$ ($0.00$)& $\textbf{95.43}$ ($0.07$) & $2.44$ ($0.00$)& $\textbf{95.48}$ ($0.07$) & $17.42$ ($0.02$) \\ \hline 
COnB $B=10$ & $\textbf{94.92}$ ($0.07$) & $1.77$ ($0.00$)& $\textbf{95.42}$ ($0.07$) & $2.18$ ($0.00$)& $\textbf{95.78}$ ($0.06$) & $15.61$ ($0.01$) \\ \hline

\multicolumn{7}{c}{equicorrelation $\Sigma$, $n=10^5$} \\ \hline
 delta & $\textbf{94.76}$ ($0.07$) & $1.31$ ($0.00$)& $\textbf{93.94}$ ($0.08$) & $1.36$ ($0.00$)& $\textit{25.95}$ ($0.14$) & $1.38$ ($0.00$)\\ \hline
 BM & $\textbf{93.20}$ ($0.08$) & $1.30$ ($0.00$)& $\textbf{92.10}$ ($0.09$) & $1.44$ ($0.00$)& $99.74$ ($0.02$) & $34.65$ ($0.22$)\\ \hline
 RS & $\textbf{94.50}$ ($0.07$) & $1.69$ ($0.01$)& $\textbf{94.42}$ ($0.07$) & $1.82$ ($0.01$)& $98.61$ ($0.04$) & $13.90$ ($0.13$)\\ \hline
OB $B=10$ & $\textbf{92.88}$ ($0.08$) & $1.35$ ($0.00$)& $\textbf{92.78}$ ($0.08$) & $1.45$ ($0.00$) & $89.23$ ($0.31$) & $19.42$ ($0.04$) \\ \hline
OB $B=100$ & $\textbf{95.28}$ ($0.07$) & $1.39$ ($0.00$)& $\textbf{94.73}$ ($0.07$) & $1.52$ ($0.00$) & $\textbf{94.29}$ ($0.23$) & $20.78$ ($0.03$)\\ \hline
$\text{HiGrad}_{(2,2)}$ & $\textbf{95.00}$ ($0.07$) & $2.59$ ($0.01$)& $\textbf{95.12}$ ($0.07$) & $2.78$ ($0.01$)& $\textbf{95.93}$ ($0.06$) & $2.83$ ($0.01$)\\ \hline
COfB ASGD $B=3$  & $\textbf{94.76}$ ($0.07$) & $2.58$ ($0.00$)& $\textbf{94.66}$ ($0.07$) & $2.74$ ($0.00$)& $\textbf{93.23}$ ($0.08$) & $32.56$ ($0.04$)\\ \hline
COfB ASGD $B=5$  & $\textbf{95.52}$ ($0.07$) & $1.77$ ($0.00$)& $\textbf{95.02}$ ($0.07$) & $1.88$ ($0.00$)& $90.80$ ($0.09$) & $52.72$ ($0.29$)\\ \hline
COfB ASGD $B=10$ & ${\textbf{95.44}}$ ($0.07$) & ${1.48}$ ($0.00$)& ${\textbf{94.95}}$ ($0.07$) & ${1.58}$ ($0.00$)& $91.87$ ($0.09$) & $49.84$ ($0.22$)\\ \hline
COfB SGD $B=3$  & $\textbf{93.28}$ ($0.08$) & $3.19$ ($0.00$)& $\textbf{93.85}$ ($0.08$) & $3.26$ ($0.00$)& ${\textbf{94.53}}$ ($0.07$) & ${4.43}$ ($0.00$)\\ \hline
COfB SGD $B=5$  & $\textbf{93.16}$ ($0.08$) & $2.18$ ($0.00$)& $\textbf{93.45}$ ($0.08$) & $2.23$ ($0.00$)& $\textbf{94.20}$ ($0.07$) & $3.06$ ($0.00$)\\ \hline
COfB SGD $B=10$ & $\textbf{95.20}$ ($0.07$) & $3.94$ ($0.00$)& $\textbf{94.78}$ ($0.07$) & $3.94$ ($0.00$)& $\textbf{94.08}$ ($0.07$) & $2.57$ ($0.00$)\\ \hline
COnB $B=3$ & $\textbf{95.16}$ ($0.07$) & $2.05$ ($0.00$)& $\textbf{95.00}$ ($0.07$) & $2.28$ ($0.00$)& $\textbf{92.55}$ ($0.08$) & $31.28$ ($0.16$) \\ \hline 
COnB $B=5$ & $\textbf{95.16}$ ($0.07$) & $1.65$ ($0.00$)& $\textbf{94.60}$ ($0.07$) & $1.89$ ($0.00$)& $\textbf{92.26}$ ($0.08$) & $23.83$ ($0.04$) \\ \hline 
COnB $B=10$ & $\textbf{95.60}$ ($0.06$) & $1.48$ ($0.00$)& $\textbf{95.07}$ ($0.07$) & $1.68$ ($0.00$)& $\textbf{92.54}$ ($0.08$) & $22.08$ ($0.04$) \\ \hline
    \end{tabular}}
 \caption{Full results for the linear regression experiment.}
\label{SMtable: table 1}
\end{table*}

\begin{table*}
\centering
\resizebox{\linewidth}{!}
{\begin{tabular}{cc c c c cc} 
\hline
\multicolumn{7}{c}{Identity $\Sigma$, $n=10^5$} \\
\hline
& \multicolumn{2}{c}{$d=5$} & \multicolumn{2}{c}{$d=20$} & \multicolumn{2}{c}{$d=200$}\\ 
\hline
& Cov (\%) & Len ($\times 10^{-2}$)& Cov (\%) & Len ($\times 10^{-2}$) & Cov (\%) & Len ($\times 10^{-2}$) \\ \hline
delta & $\textbf{95.00}$ ($0.07$) & $3.10$ ($0.00$)& $\textbf{94.12}$ ($0.07$) & $3.68$ ($0.00$)& $\textit{61.92}$ ($0.15$) & $5.85$ ($0.00$)\\ \hline
BM & $89.33$ ($0.10$) & $2.64$ ($0.00$)& $87.29$ ($0.11$) & $3.11$ ($0.01$)& $\textit{57.47}$ ($0.16$) & $5.56$ ($0.02$)\\ \hline
RS & $\textbf{94.17}$ ($0.07$) & $3.84$ ($0.01$)& $\textbf{94.04}$ ($0.07$) & $5.41$ ($0.02$)& $76.55$ ($0.13$) & $9.74$ ($0.04$)\\ \hline
OB ($B=100$) & $\textbf{95.00}$ ($0.07$) & $3.21$ ($0.00$)& $\textbf{96.71}$ ($0.06$) & $4.34$ ($0.01$)& $99.88$ ($0.01$) & $50.95$ ($0.23$) \\ \hline
OB ($B=200$) & $\textbf{94.83}$ ($0.07$) & $3.18$ ($0.00$)& $\textbf{96.96}$ ($0.05$) & $4.41$ ($0.01$)& $99.91$ ($0.01$) & $50.71$ ($0.23$) \\ \hline
$\text{HiGrad}_{(2,2)}$ & $\textbf{94.33}$ ($0.07$) & $5.77$ ($0.02$)& $\textbf{95.46}$ ($0.07$) & $7.10$ ($0.02$)& $80.61$ ($0.13$) & $10.26$ ($0.04$)\\ \hline
COfB ASGD $B=3$  & $\textbf{94.12}$ ($0.07$) & $4.21$ ($0.00$)& $\textbf{95.03}$ ($0.07$) & $7.32$ ($0.01$)& $\textbf{92.34}$ ($0.08$) & $19.00$ ($0.02$)\\ \hline
COfB ASGD $B=5$  & $\textbf{95.08}$ ($0.07$) & $4.14$ ($0.00$)& $\textbf{94.77}$ ($0.07$) & $5.01$ ($0.00$)& $90.64$ ($0.09$) & $13.03$ ($0.01$)\\ \hline
COfB ASGD $B=10$ & $\textbf{94.88}$ ($0.07$) & $3.52$ ($0.00$)& $\textbf{94.51}$ ($0.07$) & $4.22$ ($0.00$)& $89.37$ ($0.10$) & $11.06$ ($0.01$)\\ \hline
COfB SGD $B=3$  & $\textbf{95.40}$ ($0.07$) & $9.38$ ($0.01$)& $\textbf{94.99}$ ($0.07$) & $9.42$ ($0.01$)& $\textbf{94.72}$ ($0.07$) & $25.61$ ($0.03$)\\ \hline
COfB SGD $B=5$  & $\textbf{95.44}$ ($0.07$) & $7.93$ ($0.00$)& $\textbf{94.70}$ ($0.07$) & $7.93$ ($0.00$)& $\textbf{94.66}$ ($0.07$) & $17.57$ ($0.01$)\\ \hline
COfB SGD $B=10$ & $\textbf{95.44}$ ($0.07$) & $7.93$ ($0.00$)& $\textbf{94.70}$ ($0.07$) & $7.93$ ($0.00$)& $\textbf{94.54}$ ($0.07$) & $14.84$ ($0.01$)\\ \hline
COnB ($B=3$) & $\textbf{94.33}$ ($0.07$) & $4.71$ ($0.02$)& $\textbf{95.62}$ ($0.06$) & $6.56$ ($0.03$)& $99.42$ ($0.02$) & $75.25$ ($0.43$) \\ \hline
COnB ($B=5$) & $\textbf{95.33}$ ($0.07$) & $3.93$ ($0.01$)& $\textbf{96.79}$ ($0.06$) & $5.35$ ($0.02$)& $99.50$ ($0.02$) & $64.17$ ($0.35$) \\ \hline
COnB ($B=10$) & $\textbf{94.83}$ ($0.07$) & $3.55$ ($0.01$)& $\textbf{97.00}$ ($0.05$) & $4.85$ ($0.01$)& $99.72$ ($0.02$) & $57.67$ ($0.30$) \\ \hline
% \hline
% \multicolumn{7}{c}{Identity $\Sigma$, $B\times n=10^5$, $B\times n=10^5$}\\
% \hline

%     \end{tabular}
%  \caption{Logistic Regression experiment; Identity $\Sigma$}
% \label{table: logistic regression table 1}
% \end{table*}

% \begin{table*}
% \centering
% \begin{tabular}{cc c c c cc} 
% \hline
\multicolumn{7}{c}{Toeplitz $\Sigma$, $n=10^5$ } \\
\hline
% & \multicolumn{2}{c}{$d=5$} & \multicolumn{2}{c}{$d=20$} & \multicolumn{2}{c}{$d=200$}\\ \hline
%  & Cov (\%) & Len ($\times 10^{-2}$)& Cov (\%) & Len ($\times 10^{-2}$) & Cov (\%) & Len ($\times 10^{-2}$) \\ \hline
delta & $\textbf{94.83}$ ($0.07$) & $4.05$ ($0.00$)& $\textbf{93.29}$ ($0.08$) & $5.59$ ($0.00$)& $\textit{53.69}$ ($0.16$) & $9.56$ ($0.00$)\\ \hline
BM & $84.00$ ($0.12$) & $3.16$ ($0.01$)& $\textit{75.25}$ ($0.14$) & $3.75$ ($0.01$)& $\textit{34.93}$ ($0.15$) & $7.30$ ($0.03$)\\ \hline
RS & $\textbf{92.67}$ ($0.08$) & $5.14$ ($0.02$)& $90.88$ ($0.09$) & $7.26$ ($0.03$)& $76.38$ ($0.13$) & $17.45$ ($0.10$)\\ \hline
OB ($B=100$) & $\textbf{95.00}$ ($0.07$) & $4.22$ ($0.00$)& $\textbf{94.04}$ ($0.07$) & $6.65$ ($0.01$)& $99.78$ ($0.01$) & $69.55$ ($0.26$) \\ \hline
OB ($B=200$) & $\textbf{95.00}$ ($0.07$) & $4.24$ ($0.00$)& $\textbf{94.67}$ ($0.07$) & $6.70$ ($0.01$)& $99.78$ ($0.01$) & $69.28$ ($0.26$) \\ \hline
$\text{HiGrad}_{(2,2)}$ & $\textbf{95.33}$ ($0.07$) & $7.18$ ($0.03$)& $\textbf{93.38}$ ($0.08$) & $8.92$ ($0.03$)& $\textit{57.02}$ ($0.16$) & $10.27$ ($0.04$)\\ \hline
COfB ASGD $B=3$  & $\textbf{94.12}$ ($0.07$) & $5.70$ ($0.00$)& $\textbf{94.77}$ ($0.07$) & $11.49$ ($0.01$)& $\textbf{93.79}$ ($0.08$) & $42.27$ ($0.05$)\\ \hline
COfB ASGD $B=5$  & $\textbf{94.32}$ ($0.07$) & $4.82$ ($0.00$)& $\textbf{94.81}$ ($0.07$) & $7.97$ ($0.01$)& $\textbf{93.23}$ ($0.08$) & $28.81$ ($0.02$)\\ \hline
COfB ASGD $B=10$ & $\textbf{94.88}$ ($0.07$) & $4.61$ ($0.00$)& $\textbf{94.65}$ ($0.07$) & $6.69$ ($0.00$)& $\textbf{93.09}$ ($0.08$) & $24.43$ ($0.01$)\\ \hline
COfB SGD $B=3$  & $\textbf{95.36}$ ($0.07$) & $9.48$ ($0.01$)& $\textbf{94.80}$ ($0.07$) & $9.65$ ($0.01$)& $\textbf{94.41}$ ($0.07$) & $30.53$ ($0.03$)\\ \hline
COfB SGD $B=5$  & $\textbf{95.40}$ ($0.07$) & $7.98$ ($0.00$)& $\textbf{94.37}$ ($0.07$) & $8.16$ ($0.00$)& $\textbf{93.79}$ ($0.08$) & $20.76$ ($0.02$)\\ \hline
COfB SGD $B=10$ & $\textbf{95.40}$ ($0.07$) & $7.98$ ($0.00$)& $\textbf{94.37}$ ($0.07$) & $8.16$ ($0.00$)& $\textbf{93.62}$ ($0.08$) & $17.56$ ($0.01$)\\ \hline
COnB ($B=3$) & $\textbf{94.00}$ ($0.08$) & $6.22$ ($0.02$)& $\textbf{94.71}$ ($0.07$) & $10.27$ ($0.05$)& $\textbf{97.82}$ ($0.05$) & $99.61$ ($0.60$) \\ \hline
COnB ($B=5$) & $\textbf{95.00}$ ($0.07$) & $5.25$ ($0.02$)& $\textbf{94.71}$ ($0.07$) & $8.20$ ($0.03$)& $98.75$ ($0.04$) & $85.15$ ($0.45$) \\ \hline
COnB ($B=10$) & $\textbf{94.83}$ ($0.07$) & $4.84$ ($0.01$)& $\textbf{94.29}$ ($0.07$) & $7.25$ ($0.02$)& $99.31$ ($0.03$) & $78.02$ ($0.37$) \\ \hline
% \multicolumn{7}{c}{Toeplitz $\Sigma$, $B\times n=10^5$ , $B\times n=10^5$} \\
%  \hline
%     \end{tabular}
%  \caption{Logistic Regression experiment; Toeplitz $\Sigma$}
% \label{table: logistic regression table 2}
% \end{table*}

% \begin{table*}
% \centering
% \begin{tabular}{cc c c c cc} 
% \hline
\multicolumn{7}{c}{equicorrelation $\Sigma$, $n=10^5$} \\
\hline
%  & \multicolumn{2}{c}{$d=5$} & \multicolumn{2}{c}{$d=20$} & \multicolumn{2}{c}{$d=200$}\\ \hline
%  & Cov (\%) & Len ($\times 10^{-2}$)& Cov (\%) & Len ($\times 10^{-2}$) & Cov (\%) & Len ($\times 10^{-2}$) \\ \hline
delta & $\textbf{94.83}$ ($0.07$) & $3.39$ ($0.00$)& $\textbf{94.12}$ ($0.07$) & $5.19$ ($0.00$)& $\textit{31.55}$ ($0.15$) & $13.50$ ($0.01$)\\ \hline
BM & $90.00$ ($0.09$) & $2.92$ ($0.00$)& $81.25$ ($0.12$) & $3.84$ ($0.01$)& $\textit{14.67}$ ($0.11$) & $7.17$ ($0.03$)\\ \hline
RS & $\textbf{94.50}$ ($0.07$) & $4.41$ ($0.02$)& $91.08$ ($0.09$) & $6.31$ ($0.02$)& $74.79$ ($0.14$) & $32.13$ ($0.20$)\\ \hline
OB ($B=100$) & $\textbf{95.00}$ ($0.07$) & $3.50$ ($0.00$)& $\textbf{95.46}$ ($0.07$) & $6.37$ ($0.02$)& $98.12$ ($0.04$) & $110.74$ ($0.42$) \\ \hline
OB ($B=200$) & $\textbf{94.67}$ ($0.07$) & $3.50$ ($0.00$)& $\textbf{95.92}$ ($0.06$) & $6.42$ ($0.02$)& $98.38$ ($0.04$) & $109.89$ ($0.40$) \\ \hline
$\text{HiGrad}_{(2,2)}$ & $\textbf{96.17}$ ($0.06$) & $5.97$ ($0.02$)& $\textbf{95.00}$ ($0.07$) & $9.11$ ($0.03$)& $\textit{26.94}$ ($0.14$) & $11.30$ ($0.04$)\\ \hline
COfB ASGD $B=3$  & $\textbf{94.12}$ ($0.07$) & $4.68$ ($0.00$)& $\textbf{94.87}$ ($0.07$) & $10.42$ ($0.01$)& $83.65$ ($0.12$) & $64.61$ ($0.08$)\\ \hline
COfB ASGD $B=5$  & $\textbf{93.96}$ ($0.08$) & $3.92$ ($0.00$)& $\textbf{94.96}$ ($0.07$) & $7.11$ ($0.01$)& $\textit{76.52}$ ($0.13$) & $44.29$ ($0.04$)\\ \hline
COfB ASGD $B=10$ & $\textbf{95.12}$ ($0.07$) & $3.85$ ($0.00$)& $\textbf{94.78}$ ($0.07$) & $6.04$ ($0.00$)& $\textit{71.44}$ ($0.14$) & $37.42$ ($0.02$)\\ \hline
COfB SGD $B=3$  & $\textbf{94.96}$ ($0.07$) & $9.39$ ($0.01$)& $\textbf{93.85}$ ($0.08$) & $9.50$ ($0.01$)& $\textit{77.71}$ ($0.13$) & $39.47$ ($0.05$)\\ \hline
COfB SGD $B=5$  & $\textbf{94.76}$ ($0.07$) & $7.92$ ($0.00$)& $\textbf{93.59}$ ($0.08$) & $8.06$ ($0.00$)& $\textit{66.67}$ ($0.15$) & $26.98$ ($0.03$)\\ \hline
COfB SGD $B=10$ & $\textbf{94.76}$ ($0.07$) & $7.92$ ($0.00$)& $\textbf{93.59}$ ($0.08$) & $8.06$ ($0.00$)& $\textit{58.94}$ ($0.16$) & $22.78$ ($0.02$)\\ \hline
COnB ($B=3$) & $\textbf{94.17}$ ($0.07$) & $5.18$ ($0.02$)& $\textbf{95.96}$ ($0.06$) & $9.51$ ($0.06$)& $\textbf{93.07}$ ($0.08$) & $138.99$ ($1.13$) \\ \hline
COnB ($B=5$) & $\textbf{94.67}$ ($0.07$) & $4.38$ ($0.01$)& $\textbf{95.12}$ ($0.07$) & $7.66$ ($0.03$)& $\textbf{94.14}$ ($0.07$) & $115.73$ ($0.73$) \\ \hline
COnB ($B=10$) & $\textbf{95.00}$ ($0.07$) & $3.94$ ($0.01$)& $\textbf{95.83}$ ($0.06$) & $6.92$ ($0.02$)& $\textbf{95.90}$ ($0.06$) & $111.29$ ($0.60$) \\ \hline
% \multicolumn{7}{c}{equicorrelation $\Sigma$, $B\times n=10^5$, $B\times n=10^5$ } \\
% \hline
\end{tabular}}
 \caption{Full results for the logistic regression experiment.}
\label{SMtable: logistic regression table 3}
\end{table*}

\begin{table*}
\centering
\resizebox{\linewidth}{!}
{\begin{tabular}{cc c c c c} 
\hline
\multicolumn{6}{c}{Identity $\Sigma$, $n=100$} \\
\hline
& &\multicolumn{2}{c}{$p=3$}  & \multicolumn{2}{c}{$p=15$}\\ 
& & Cov (\%) & Len ($\times 10^{-2}$)& Cov (\%) & Len($\times 10^{-2}$) \\
% \multirow{2}{*}{COfB ASGD $B=3$}
% & $\in T^*$ &  & & & \\
% & $\notin T^{*}$ & & & & \\
% \multirow{2}{*}{COfB ASGD $B=10$}
% & $\in T^*$ & & & & \\
% & $\notin T^{*}$ & & & & \\
% \multirow{2}{*}{COnB ASGD $B=3$}
% & $\in T^*$ & & & & \\
% & $\notin T^{*}$ & & & & \\
% \multirow{2}{*}{COnB ASGD $B=10$}
% & $\in T^*$ & & & & \\
% & $\notin T^{*}$ & & & & \\
\hline
\multirow{2}{*}{COfB ASGD $B=3$}
& $\in T^*$ & $\textbf{94.87}$ ($22.07$) &  $1.52 $ ($0.00$) & $\textbf{94.92}$ ($21.96$) & $5.57 $ ($0.01$)\\
& $\notin T^*$ & $\textbf{99.87}$ ($3.62$)& $0.03$ ($0.00$) &  $\textbf{99.76}$ ($4.92$)& $0.38$ ($0.00$) \\
\hline
\multirow{2}{*}{COfB ASGD $B=10$}
& $\in T^*$ & $\textbf{95.13}$ ($21.52$) &  $0.82 $ ($0.00$) & $\textbf{95.44}$ ($20.86$) & $3.74 $ ($0.01$)\\
& $\notin T^*$ & $\textbf{99.85}$ ($3.91$)& $0.02$ ($0.00$) &  $\textbf{99.77}$ ($4.81$)& $0.26$ ($0.00$) \\
\hline
\multirow{2}{*}{COnB ASGD $B=3$}
& $\in T^*$ & $\textbf{95.40}$ ($20.95$) &  $1.15 $ ($0.00$) & $\textbf{96.07}$ ($19.44$) & $5.13 $ ($0.01$)\\
& $\notin T^*$ & $\textbf{99.92}$ ($2.83$)& $0.02$ ($0.00$) &  $\textbf{99.93}$ ($2.69$)& $0.33$ ($0.00$) \\
\hline
\multirow{2}{*}{COnB ASGD $B=10$}
& $\in T^*$ & $\textbf{95.40}$ ($20.95$) &  $0.92 $ ($0.00$) & $\textbf{96.96}$ ($17.17$) & $3.99 $ ($0.01$)\\
& $\notin T^*$ & $\textbf{99.93}$ ($2.56$)& $0.02$ ($0.00$) &  $\textbf{99.97}$ ($1.71$)& $0.27$ ($0.00$) \\

\hline
\multicolumn{6}{c}{Toeplitz $\Sigma$, $n=100$} \\
\hline
\multirow{2}{*}{COfB ASGD $B=3$}
& $\in T^*$ & $\textbf{95.20}$ ($21.38$) &  $2.75 $ ($0.01$) & $\textbf{95.12}$ ($21.54$) & $34.87 $ ($0.11$)\\
& $\notin T^*$ & $\textbf{99.88}$ ($3.42$)& $0.05$ ($0.00$) &  $\textbf{99.09}$ ($9.52$)& $8.62$ ($0.07$) \\
\hline
\multirow{2}{*}{COfB ASGD $B=10$}
& $\in T^*$ & $\textbf{95.07}$ ($21.66$) &  $0.86 $ ($0.00$) & $\textbf{93.13}$ ($25.29$) & $4.33 $ ($0.01$)\\
& $\notin T^*$ & $\textbf{99.86}$ ($3.74$)& $0.02$ ($0.00$) &  $\textbf{99.12}$ ($9.32$)& $1.00$ ($0.01$) \\
\hline
\multirow{2}{*}{COnB ASGD $B=3$}
& $\in T^*$ & $\textbf{94.73}$ ($22.34$) &  $1.28 $ ($0.01$) & $\textbf{97.19}$ ($16.54$) & $10.68 $ ($0.03$)\\
& $\notin T^*$ & $\textbf{99.93}$ ($2.73$)& $0.03$ ($0.00$) &  $\textbf{99.80}$ ($4.47$)& $2.47$ ($0.02$) \\
\hline
\multirow{2}{*}{COnB ASGD $B=10$}
& $\in T^*$ & $\textbf{95.40}$ ($20.95$) &  $1.07 $ ($0.00$) & $98.85$ ($10.65$) & $10.27 $ ($0.04$)\\
& $\notin T^*$ & $\textbf{99.96}$ ($2.00$)& $0.02$ ($0.00$) &  $\textbf{99.98}$ ($1.51$)& $2.41$ ($0.02$) \\
\hline
\multicolumn{6}{c}{equicorrelation $\Sigma$, $n=100$} \\
\hline
\multirow{2}{*}{COfB ASGD $B=3$}
& $\in T^*$ & $\textbf{94.53}$ ($22.73$) &  $3.36 $ ($0.02$) & $\textbf{95.08}$ ($21.63$) & $1.94 $ ($21.00$)\\
& $\notin T^*$ & $\textbf{99.89}$ ($3.32$)& $0.08$ ($0.00$) &  $98.69$ ($11.37$)& $9.35$ ($20.00$) \\
\hline
\multirow{2}{*}{COfB ASGD $B=10$}
& $\in T^*$ & $\textbf{94.87}$ ($22.07$) &  $1.23 $ ($0.01$) & $\textbf{94.81}$ ($22.18$) & $33.65 $ ($0.30$)\\
& $\notin T^*$ & $\textbf{99.90}$ ($3.23$)& $0.03$ ($0.00$) &  $98.74$ ($11.16$)& $10.39$ ($0.17$) \\
\hline
\multirow{2}{*}{COnB ASGD $B=3$}
& $\in T^*$ & $\textbf{93.60}$ ($24.48$) &  $1.47 $ ($0.01$) & $\textbf{95.95}$ ($19.72$) & $20.26 $ ($0.12$)\\
& $\notin T^*$ & $\textbf{99.93}$ ($2.55$)& $0.03$ ($0.00$) &  $\textbf{99.80}$ ($4.42$)& $5.52$ ($0.07$) \\
\hline
\multirow{2}{*}{COnB ASGD $B=10$}
& $\in T^*$ & $\textbf{93.93}$ ($23.87$) &  $1.15 $ ($0.00$) & $\textbf{95.55}$ ($20.63$) & $15.59 $ ($0.17$)\\
& $\notin T^*$ & $\textbf{99.97}$ ($1.82$)& $0.03$ ($0.00$) &  $\textbf{99.96}$ ($2.00$)& $4.73$ ($0.10$) \\

\hline
\end{tabular}}
 \caption{Results for sparse linear regression. }
\label{SMtable: Sparse Linear Regression}
\end{table*}

\vskip 0.2in
\newpage
\bibliography{reference}

\end{document}